\documentclass{article}

\usepackage{microtype}
\usepackage{graphicx}
\usepackage{booktabs} % for professional tables

\usepackage{hyperref}
\usepackage{url}

\usepackage[utf8]{inputenc} % allow utf-8 input
\usepackage[T1]{fontenc}    % use 8-bit T1 fonts
\usepackage{hyperref}       % hyperlinks
\usepackage{url}            % simple URL typesetting
\usepackage{booktabs}       % professional-quality tables
\usepackage{amsfonts}       % blackboard math symbols
\usepackage{amsmath}
\usepackage{amssymb}
\usepackage{nicefrac}       % compact symbols for 1/2, etc.
\usepackage{microtype}      % microtypography
\usepackage{xcolor}         % colors
\usepackage{graphicx}
\usepackage{subcaption}
\usepackage{caption} 
\usepackage{algorithm}
 \usepackage[acronym,nohypertypes={acronym,notation}]{glossaries}
\usepackage{multirow}
\usepackage{doi}

\usepackage{xcolor}
\hypersetup{
    colorlinks,
    linkcolor={red!50!black},
    citecolor={blue!50!black},
    urlcolor={blue!80!black}
}

\usepackage[preprint]{icml2025}

\usepackage{amsmath}
\usepackage{amssymb}
\usepackage{mathtools}
\usepackage{amsthm}
\usepackage[subtle]{savetrees}
\usepackage[capitalize,noabbrev]{cleveref}

\theoremstyle{plain}

\theoremstyle{definition}

\theoremstyle{remark}

\usepackage[textsize=tiny]{todonotes}

\makeatletter
\DeclareRobustCommand\onedot{\futurelet\@let@token\@onedot}
\def\@onedot{\ifx\@let@token.\else.\null\fi\xspace}

\usepackage{xcolor}
\hypersetup{
    colorlinks,
    linkcolor={blue!80!black},
    citecolor={green!80!black},
    urlcolor={blue!50!black}
}

\newcommand{\defn}{\emph}

\glsdisablehyper

\icmltitlerunning{Sparse Training from Random Initialization:
Aligning Lottery Ticket Masks using Weight Symmetry}

\begin{document}
\newacronym{dnn}{DNN}{Deep Neural Network}
\newacronym{cnn}{CNN}{Convolutional Neural Network}
\newacronym{nn}{NN}{Neural Network}
\newacronym{ann}{ANN}{Artificial Neural Network}
\newacronym{snn}{SNN}{Sparse Neural Network}
\newacronym{lt}{LT}{Lottery Ticket}
\newacronym{ntk}{NTK}{the Neural Tangent Kernel}
\newacronym{lth}{LTH}{Lottery Ticket Hypothesis}
\newacronym{nlp}{NLP}{Natural Language Processing}
\newacronym{dst}{DST}{Dynamic Sparse Training}
\newacronym{gpu}{GPU}{Graphics Processing Unit}
\newacronym{pi}{PI}{Primary Investigator}
\newacronym{ai}{AI}{Artificial Intelligence}
\newacronym{ml}{ML}{Machine Learning}
\newacronym{cv}{CV}{Computer Vision}
\newacronym{neurips}{NeurIPS}{Neural Information Processing Systems}
\newacronym{iclr}{ICLR}{the International Conference on Learning Representations}
\newacronym{cvpr}{CVPR}{the IEEE/CVF Conference on Computer Vision and Pattern Recognition}
\newacronym{flops}{FLOPs}{Floating Point Operations}
\newacronym{relu}{ReLU}{ReLU}
\newacronym{srigl}{SRigL}{Structured RigL}
\newacronym{rigl}{RigL}{Rigging the Lottery Ticket}
\newacronym{set}{SET}{Sparse Evolutionary Training}
\newacronym{erk}{ERK}{Erd\H{o}s-R\'enyi-Kernel}
\newacronym{ste}{STE}{Straight-Through Estimator}
\newacronym{srste}{SR-STE}{Sparse-Refined Straight-Through Estimator}
\newacronym{ilsvrc}{ILSVRC-12}{2012 ImageNet Large Scale Visual Recognition Challenge}
\newacronym{ast}{AST}{Alternating Sparse Training}
\newacronym{deepr}{DeepR}{Deep Rewiring}
\newacronym{snfs}{SNFS}{Sparse Networks from Scratch}
\newacronym{dsr}{DSR}{Dynamic Sparse Reparameterization}
\newacronym{topkast}{Top-KAST}{Top-K Always Sparse Training}
\newacronym{mest}{MEST}{Memory-Economic Sparse Training}
\newacronym{dsb}{DSB}{Dynamic Shuffled Block}
\newacronym{dynsparse}{DynSparse}{Dynamic Sparsity}
\newacronym{vit}{ViT-B/16}{Vision Transformer}
\newacronym{cpu}{CPU}{Central Processing Unit}
\newacronym{csr}{CSR}{Compressed Sparse Row}
\newacronym{itop}{ITOP}{In Time Overparameterization Rate}
\newacronym{pai}{PaI}{Pruning at Initialization}
\newacronym{lmc}{LMC}{Linear-Mode Connectivity}
\newacronym{imp}{IMP}{Iterative Magnitude Pruning}
\newacronym{impft}{IMP-FT}{Iterative Magnitude Pruning - Fine Tuning}
\newacronym{sgd}{SGD}{Stochastic Gradient Descent}
\twocolumn[
\icmltitle{\texorpdfstring{Sparse Training from Random Initialization:\\
Aligning Lottery Ticket Masks using Weight Symmetry}{Sparse Training from Random Initialization:
Aligning Lottery Ticket Masks using Weight Symmetry}}



\icmlsetsymbol{equal}{*}

\begin{icmlauthorlist}
\icmlauthor{Mohammed Adnan}{equal,uofc,vector}
\icmlauthor{Rohan Jain}{equal,uofc}
\icmlauthor{Ekansh Sharma}{uoft,vector}
\icmlauthor{Rahul G.\ Krishnan}{uoft,vector}
\icmlauthor{Yani Ioannou}{uofc}
\end{icmlauthorlist}

\icmlaffiliation{uofc}{
% Dept.\ of Electrical and Software Engineering, 
Schulich School of Engineering,
University of Calgary}
\icmlaffiliation{vector}{Vector Institute for AI}
\icmlaffiliation{uoft}{Dept.\ of Computer Science, University of Toronto}

\icmlcorrespondingauthor{Mohammed Adnan}{adnan.ahmad@ucalgary.ca}
\icmlcorrespondingauthor{Yani Ioannou}{yani.ioannou@ucalgary.ca}

\icmlkeywords{Machine Learning, ICML}

\vskip 0.3in
]



\printAffiliationsAndNotice{\icmlEqualContribution} % otherwise use the standard text.

\begin{abstract}
  The \gls{lth} suggests there exists a sparse \gls{lth} mask and weights that achieve the same generalization performance as the dense model while using significantly fewer parameters. 
  % \gls{lth} achieves this by iteratively sparsifying and re-training within the pruned solution basin. 
  However, finding a \gls{lth} solution is computationally expensive, and a \gls{lth}'s sparsity mask does not generalize to other random weight initializations. 
  %any other random initialization sparsified using the winning ticket mask fails to achieve good generalization performance. 
  Recent work has suggested that neural networks trained from random initialization find solutions within the same basin modulo permutation, and proposes a method to align trained models within the same loss basin. We hypothesize that misalignment of basins is the reason why \gls{lth} masks do not generalize to new random initializations and propose permuting the \gls{lth} mask to align with the new optimization basin when performing sparse training from a different random init.
  %than the one used to obtained the pruned mask with \gls{lth}.
  %We propose to permute the winning ticket mask to align with the new optimization basin when sparse training from a different random initialization than the pruned mask was derived from. 
  We empirically show a significant increase in generalization when sparse training from random initialization with the permuted mask as compared to using the non-permuted \gls{lth} mask, on multiple datasets~(CIFAR-10/100 \& ImageNet) and models (VGG11 \& ResNet20/50). Our codebase for reproducing the results is publicly available at \href{https://github.com/calgaryml/sparse-rebasin}{here}.
  %Our experiments also provide new insights into the loss-landscapes of \gls{dnn} and \gls{lmc} of \gls{imp}.
  % of varying widths.
  % We believe better methods of matching permutation basins will improve the results significantly, demonstrated by the effect of increasing model width, and correspondingly the ease of activation/weight-matching, on our results.
\end{abstract}
\glsresetall

\section{Introduction}
In recent years, foundation models have achieved state-of-the-art results for different tasks. However, the exponential increase in the size of state-of-the-art models requires a similarly exponential increase in the memory and computational costs 
%hardware/compute 
required to train, store and use these models --- decreasing the accessibility of these models for researchers and practitioners alike. 
To overcome this issue, different model compression methods, such as pruning, quantization and knowledge distillation, have been proposed to reduce the model size at different phases of training or inference. Post-training model pruning~\citep{han2016deepcompression} has been shown to be effective in compressing the model size, and seminal works have demonstrated that large models can be pruned after training with minimal loss in accuracy~\citep{gale2019statesparsitydeepneural, han2015learning}.  While model pruning makes inference more efficient, it does not reduce the computational cost of training the model. 
% In contrast to post-training pruning, \gls{pai} methods aim to find a mask at the random initialization, which can yield matching accuracy. However, \gls{pai} methods~\citep{frankle2020pai} have not been able to match dense performance even at moderate sparsity level; recent work has highlighted the limitations of \gls{pai} from an information-theoretic perspective, showing that finding extremely sparse subnetworks without fully training the network cannot achieve performance comparable to dense models~\citep{kumar2024no}.

Motivated by the goal of training a sparse model from a random initialization, \citet{frankle2019lottery} demonstrated that training with a highly sparse mask is possible and proposed the \gls{lth} to identify sparse subnetworks that, when trained, can match the performance of a dense model. The key caveat is that a dense model must first be trained to find the sparse mask, which can \emph{only} be used with the same random initialization that was used to train the dense model.
% the \gls{lth} proposes to solve the sparse training problem by reusing the same initialization as used to train the pruned models. On very small models, training from such an initialization maintains the generalization performance of the pruned model and demonstrates that training with a highly sparse mask is possible~\citep{frankle2019lottery}. 
% In practice, however, when training even modestly-sized models, \emph{weight rewinding}~\citep{frankle2020linear} is necessary --- requiring significantly more compute than dense training alone. 
Despite \gls{lth} seeing significant interest in the research community, \gls{lth} masks cannot be used to train from a new random initialization. Furthermore, it has been observed empirically that the \gls{lth} is impractical for finding a diverse set of solutions~\citep{evci2022gradient}.

This posits our main research questions: \emph{How can we train a \gls{lth} mask from a different random initialization while maintaining good generalization? Would doing so find a more diverse set of solutions than observed with the \gls{lth} itself?}  
% If the winning ticket mask can be used again with a different random initialization, then it might make \gls{lth} a more practical algorithm for training multiple sparse models (ensembles), offsetting the cost of training a dense model once, and may provide a deeper understanding of sparse training and the \gls{lth}.

% Figure Source: https://docs.google.com/presentation/d/1sS2zjXqrXhsBVisSRm-sMA9ae9HM6NfxB29H7gzwSvU/edit?usp=sharing
\begin{figure*}[tbp]
    \centering
    \begin{subfigure}{0.48\textwidth}
      \centering
      \includegraphics[page=1, width=0.9\linewidth]{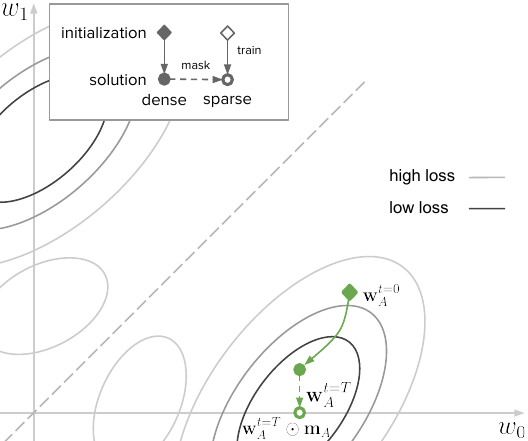}
      \caption{Dense training and pruning model $A$.}%
      \label{fig:firstsymmetry}%
    \end{subfigure}
    % \begin{subfigure}{0.33\textwidth}
    %   \centering
    %   \includegraphics[page=2, width=0.95\linewidth]{figures/sparse-rebasin-symmetry-illustration}
    %   \caption{\scriptsize{\gls{lth} training model $A$ with pruned mask.}}%
    %   \label{fig:firstsymmetrylth}%
    % \end{subfigure}
    % \qquad % space out the images a bit
    % \hfill
    \begin{subfigure}{0.48\textwidth}
      \centering
      \includegraphics[page=3, width=0.9\linewidth]{figures/sparse-rebasin-symmetry-illustration}
      \caption{Sparse training model $B$ with $A$ mask.}
      \label{fig:secondsymmetry}%
    \end{subfigure}
  \caption{\textbf{Weight Symmetry and the Sparse Training Problem}. A model with a single layer and only two parameters, $\mathbf{w}=({w_0, w_1})$, operating on a single input $x_0$ has weight symmetry in the 2D loss landscape as illustrated above. In (\subref{fig:firstsymmetry}) the original dense model, $\mathbf{w}_A$, is trained from a random dense initialization, $\mathbf{w}_A^{t=0}$ to a dense solution, $\mathbf{w}_A^{t=T}$, which is then pruned using weight magnitude resulting in the mask $\mathbf{m}_A = \left(1, 0\right)$. 
  % In (\subref{fig:firstsymmetrylth}) we re-use the init.\ $\mathbf{w}_A^{t=0}$, to train model $A$ with the pruned mask from (\subref{fig:firstsymmetry}), $\mathbf{m}_A$, as in \gls{lth}.
  In (\subref{fig:secondsymmetry}), naively using the same mask to train a model, B, from a different random initialization will likely result in the initialization being far from a good solution. Permuting the mask to match the (symmetric) basin in which the new initialization is in will enable sparse training.
  See \cref{fig:symmetry-illustration-full} in \cref{full_lth_figure} for the full figure also including the \acrfull{lth}.
  }% caption for whole figure
  \label{fig:symmetry-illustration}% label for whole figure
\end{figure*}

In this work, we try to understand why the \gls{lth} does not work for different random initializations from a weight-space symmetry perspective. 
Our hypothesis is that to reuse the \gls{lth} winning ticket mask with a different random initialization, the winning ticket mask obtained needs to be permuted such that it aligns with the optimization basin associated with the new random initialization. We illustrate our hypothesis in \cref{fig:symmetry-illustration}. 
 
To empirically validate our hypothesis, we obtain a sparse mask using \gls{imp}~\citep{renda2020,han2015learning} on model $A$ (from ~\cref{fig:symmetry-illustration}) and show that given a permutation that aligns the optimization basin of model $A$ and a new random initialization, the mask can be reused. The sparse model (with the permuted mask) can be trained to closer match the generalization performance of the LTH solution, and the permuted mask improves the generalization of the trained sparse model compared to the non-permuted mask. Furthermore, we observe drastically increased functional diversity when using our approach compared to \gls{lth} solutions.
% and show that once permutation symmetries are accounted for, it is indeed possible to reuse the \gls{lth} mask with different random initializations with improved generalization performance over naive sparse training and improved function diversity over \gls{lth}. 
Our contributions are as follows:
\begin{enumerate}

    \item 
    % Due to the permutation symmetry in the weight space, there are many functionally equivalent loss basins and the new random initialization lands in a different loss basin equivalent up to symmetry as shown in~\cref{fig:secondsymmetry}~\citep{ainsworth2023git}.
    We hypothesize that the \gls{lth}~\citep{frankle2019lottery} fails to generalize well with a new random initialization due to a mismatch between the optimization basin of the winning ticket mask and the new random initialization's solution basin. We propose a method based on permutation matching between two dense models, that permutes the winning ticket's sparse mask to align with the optimization basin of the new random initialization. We empirically demonstrate on CIFAR-10/100 and ImageNet datasets using VGG11 and ResNet models of varying widths that permuting the \gls{lth} sparse mask to align with the new random initialization improves the performance of the trained model (permuted), compared to the model trained without permuting the sparse mask (naive).
    \item We show that models trained from random initialization using the permuted \gls{lth} mask are much more functionally diverse in the solutions they learn than those found from training the \gls{lth} winning ticket mask and initialization alone~\citep{evci2022gradient}, across several existing functional diversity metrics and improved ensemble performance.
    \item Furthermore, our experiments provide novel insights about the \gls{lth} and the corresponding dense model: we show that for a fixed initialization, the dense solution and the corresponding LTH solution remain in the same loss basin once we take into account \defn{variance collapse}. 
    Notably, our conclusion differs from the conclusion drawn by \citet{paul2022unmasking}, where they did not consider the variance collapse issue when interpolating between the sparse and dense solutions. 
    
     % \item We show that by increasing the model width, the accuracy gap between the \gls{lth} and permuted solution decreases. Thus, given an algorithm, which can find an accurate permutation matching, it might be possible to reuse \gls{lth} on a new random initialization.
    % However, developing a better permutation-matching algorithm is out of the scope of this work. Our work tries to seek out a better understanding of \gls{lth} and why it does not work with a random initialization.
\end{enumerate}

\section{Background \& Related Work}
\paragraph{Linear Mode Connectivity.}

A pair of trained neural networks are said to be linearly connected if the loss along the linear path between the models remains small. The phenomenon of linear (mode) connectivity was first observed in the context of \gls{sgd} by \citet{nagarajan2019uniform}, where they showed that two neural networks trained from the same initialization but with different data orders exhibit linear connectivity. The term \gls{lmc} was introduced by \citet{frankle2020linear}, where they showed that independently trained neural networks can be linearly connected.

\paragraph{Linear Mode Connectivity \textit{modulo} Permutation.} 
% Typically, linearly interpolating between the weights of two independently trained networks usually results in a higher loss/0–1 error compared to the two endpoints. 
% Mode connectivity is a property of neural network loss landscapes, where two independently trained models can be connected by a non-linear path along which the loss remains low \cite{draxler2018essentially, garipov2018loss}.

% \citet{frankle2020linear} demonstrated the existence of a linear path between such models. 
% More formally, let $\theta_1, \theta_2$ be the parameters of two networks. The loss barrier is defined as:
% 
% \begin{equation}
%     \mathcal{B}(\theta_1, \theta_2) := \sup_{\alpha \in [0, 1]} \left[ \mathcal{L}\big((1 - \alpha)\theta_1 + \alpha\theta_2\big) - \big((1 - \alpha)\mathcal{L}(\theta_1) + \alpha\mathcal{L}(\theta_2)\big) \right] \geq 0 \ ,
% \end{equation}
% 
% where $\mathcal{L}$ is the desired loss function. If $\mathcal{B}(\theta_1, \theta_2) \approx 0$, it is said that $\theta_1$ and $\theta_2$ are linearly connected.
%\citet{entezari2022role} conjectured, that independently obtained SGD solutions have no loss barrier if one accounts for the permutation symmetries. 

\citet{entezari2022role} further observed that while a model and its randomly permuted counterpart are functionally equivalent, they are rarely linearly connected in the weight space. This misalignment suggests the presence of \textit{loss barriers} --- regions along a linear path between models where the loss is significantly higher than at the endpoints. They conjectured that independently obtained \gls{sgd} solutions exhibit no loss barrier when accounting for permutation symmetries, suggesting that all \gls{sgd}-trained networks converge to a single basin modulo permutations. Building on this conjecture, several algorithms have been developed to address permutation invariance by aligning trained networks to the same optimization basin \citep{ainsworth2023git, jordan2023repair, singh2023model, tatro2020optimizing}. \citet{ainsworth2023git} demonstrated that two models trained from different random initializations find solutions within the same basin modulo permutation symmetry. They proposed a permutation matching algorithm to permute the units of one model to align it with a reference model, enabling \gls{lmc}~\citep{frankle2020linear}. 
%\citet{benzing2022random} use a permutation found after training to exhibit LMC between networks at initialization. 
The use of activation matching for model alignment was originally introduced by \citet{liconvergent}, to ensure models learn similar representations when performing the same task. \citet{jordan2023repair} investigated the poor performance of interpolated networks, attributing it to a phenomenon they termed "\textit{variance collapse}". To address this, they proposed a method that rescales the hidden units, leading to significant improvements in the generalization performance of interpolated networks. 
A rigorous study from \citet{sharma2024simultaneous} introduced a notion of \textit{simultaneous weak linear connectivity} where a permutation, $\pi$, aligning two networks also simultaneously aligns two larger fully trained networks throughout the entire \gls{sgd} trajectory and the same $\pi$ also aligns successive iterations of independently sparsified networks found via weight rewinding. \citet{sharma2024simultaneous} also showed that for certain neural networks, sparse mask obtained via weight rewinding can be reused modulo permutations without hurting the test performance.

\paragraph{Lottery Ticket Hypothesis.}
%Neural network pruning is a highly effective method of reducing the parameters in a trained dense neural network, pruning as many as 85--95\% of weights while not significantly affecting generalization performance~\citep{han2015learning,han2016deepcompression}. %
%However, training with a sparse mask, even one obtained from pruning, from random initialization does not work well ---  motivating the \gls{lth}. 
The \gls{lth} proposes to solve the sparse training problem by re-using the same initialization as used to train the pruned models. For very small models, training from such an initialization maintains the generalization performance of the pruned model and demonstrates that training with a highly sparse mask is possible~\citep{frankle2019lottery}. 
However, subsequent work has shown that obtaining winning tickets for modestly-sized models requires using \emph{weight rewinding}~\citep{frankle2020linear} ---  requiring significantly more compute than dense training alone, especially considering that \gls{lth} also requires \gls{imp}, i.e.\ training of iteratively sparsified models. We include a detailed description of \gls{imp} in \cref{pruning}.
\citet{paul2022unmasking} analyzed the \gls{imp} algorithm and showed that sparse network obtained after $K^{\text{th}}$ IMP iteration is linearly connected to the sparse model obtained after $K+1^{\text{th}}$ \gls{imp} iteration. 
In this work, we show that once we take into account the \defn{variance collapse} studied in \citet{jordan2023repair}, we are able to show that the sparse solution obtained after the $K^{\text{th}}$ iteration is linearly connected to the dense solution. 
Furthermore, recent work has shown that the \gls{lth} effectively re-learns the original pruned solution it is derived from~\citep{evci2022gradient}. 
To make any practical use of sparse training, finding methods of sparse training from random initialization is necessary to realize any efficiency gains in training. 

\paragraph{Weight Symmetry.}
%The process of training \glspl{dnn} requires optimizing over a non-convex loss landscape consisting of numerous local minima, narrow ravines, plateaus, saddle points and {loss basins}~\citep{choromanska2015loss, dauphin2014identifying, goodfellow2015qualitativelycharacterizingneuralnetwork, liu2021loss, nguyen2017loss, simsek2021geometry}. 
%Despite non-convex optimization problems being NP-hard~\citep{shapiro2005complexity}, the nature of first-order stochastic optimizers such as \gls{sgd}~\citep{robbins1951stochastic} have been proven to be highly effective in optimizing \glspl{dnn} in practice ~\citep{hardt2016trainfastergeneralizebetter, jin2017escapesaddlepointsefficiently}. 
%Empirical evidence suggests that when training independent \gls{dnn} using \gls{sgd}, with different batch orders and initializations, the resulting training trajectories often exhibit remarkable similarities~\citep{ainsworth2023git,zhang2017understandingdeeplearningrequires}. 
%This phenomenon has been attributed to overparameterization, which creates numerous minima in the loss landscape, leading to multiple distinct functions that fit the data similarly~\citep{kawaguchi2016deep,neyshabur2017exploring}. 
\Citet{nielsen1990} demonstrated that neural networks are \textit{permutation invariant}, where swapping any two neurons within a hidden layer does not alter the underlying function being learned. The permuted network remains functionally equivalent to its original configuration, i.e.\ neural networks are symmetric functions. 
The existence of permutation symmetries in weight space creates copies of global minima at different points in weight space~\citep{entezari2022role, goodfellow2016deep, simsek2021geometry}. Weight sparsity, achieved through pruning, can reduce the number of weight symmetries in a neural network. Pruning neurons reduces the number of permutation symmetries in a layer. Unstructured pruning, in heterogeneously removing individual weights, can break the weight symmetry of individual neurons in layers.

% https://docs.google.com/presentation/d/179TFlmHkL1eTG6Y5PlXQ-y4nd7ZPWClvo43pHA6mTP8/edit?usp=sharing
\begin{figure}[tbp]
    \centering
    \includegraphics[width=\linewidth]{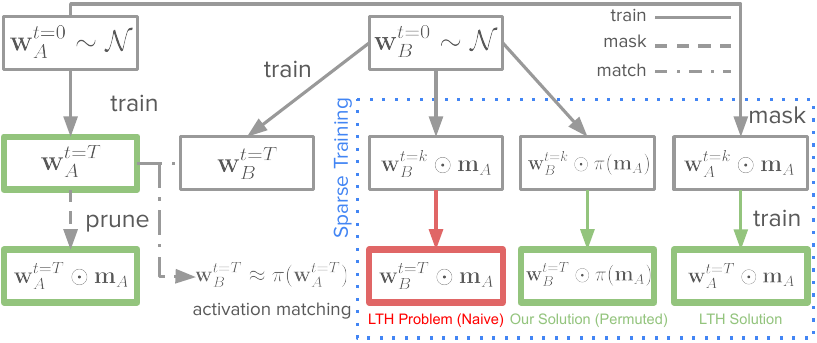}
    \caption{The overall framework of the training procedure, beginning with two distinct dense random weight initializations, $\textbf{w}_{A}^{t=0}$, $\textbf{w}_{B}^{t=0}$ sampled from a normal distribution, $\mathcal{N}$. The sparse training problem attempts to train the random initialization, $\textbf{w}_{B}^{t=0}$ using the naive mask $\mathbf{m}_{A}$, found by pruning a dense trained model, $\textbf{w}_{A}^{t=T}$. However, this results in poor generalization performance~\citep{frankle2020linear}. We propose to instead train $\textbf{w}_{B}^{t=k}$ at some rewound epoch $k$, equipped with a \textit{permuted} mask $\pi(\textbf{m}_A)$. We show that this achieves more comparable generalization to the pruned model/trained LTH solution, $\textbf{w}_{A}^{t=T} \odot \textbf{m}_{A}$.}
    \label{fig:workflow}
\end{figure}

\section{Method}

\paragraph{Motivation.}
% While \gls{lth} empirically demonstrated the existence of lottery tickets, which can be used to train a sparse network to achieve comparable accuracy to the original dense model, these lottery tickets fail to transfer to new random initializations.
In this work, we try to understand \emph{why \gls{lth} masks fail to transfer to a new random initialization}. Our hypothesis is that the loss basin corresponding to the \gls{lth} mask is not aligned with the new random initialization, as shown in~\cref{fig:symmetry-illustration}. Since the sparse mask is not aligned with the basin of the new random initialization, sparse training does not work well; therefore, aligning the \gls{lth} mask with the new random initialization may improve sparse training and enable the transfer of \gls{lth} masks to new random initializations. 

%Rewinding: 
\paragraph{Permutation Matching.}
\citet{ainsworth2023git} showed that the permutation symmetries in the weight space can be leveraged to align the basin of two models trained from different random initializations. 
% The authors proposed to match activations to align two models. 
The permutation mapping can be obtained by either matching activations or weights. In this work, we use activation matching to obtain the permutation mapping as it has been shown to be more stable in recent works~\citep{sharma2024simultaneous}. Activation matching tries to find a permutation mapping, $\pi \in S_d$ (where $S_d$ is the permutation group of order $d!$) such that by permuting the parameters of the second model, the correlation between the activations of the two models is maximized. For a model consisting of $L$ layers, each layer is sequentially matched and permuted starting from the input layer. Let $Z^{A}_l, Z^{B}_l \in \mathbb{R}^{d \times n}$ be the activations of layer $l$ of model $A$ and $B$ respectively obtained using the training data, where $d$ represents the dimensionality of the activations at layer $l$ and $n$ is the number of training data points. Then a permutation mapping for layer $l$, $\pi_l$, is obtained by solving:
% \begin{equation}
%     \pi_l = \underset{\pi}{\arg\min} ||Z^{B}_l - \pi Z^{A}_l|| = \underset{\pi}{\arg\max} \langle \pi, Z^{B}(Z^{A})^\top \rangle_F,
%     \label{eqn:matching}
% \end{equation}
\begin{equation}
\begin{split}
    \pi_l &= \underset{\pi}{\arg\min} ||Z^{B}_l - \pi Z^{A}_l|| \\
          &= \underset{\pi}{\arg\max} \langle \pi, Z^{B}(Z^{A})^\top \rangle_F
\end{split}
\label{eqn:matching}
\end{equation}
where $\langle . , .  \rangle_F$ denotes the Frobenius inner product. \cref{eqn:matching} can be formulated as a Linear Assignment Problem (LAP)~\citep{bertsekas1998network, ito2024analysis} solved via the Hungarian algorithm \citep{kuhn_hungarian}; however, the permutation found is not global optima but a greedy/approximate solution as permutation matching is a NP-hard problem. Once the permutation mapping is obtained for all the layers, the model $A$ can be permuted to match model $B$. To ensure that the permuted model does not change functionally when permuting the output dimension of layer $l$, the input dimension of the next layer is also permuted accordingly. Let $W_l$ and $b_l$ be the weights and bias of layer $l$ respectively, then the permuted weight matrix  $W_l^{p}$ and permuted bias $b_l^{p}$ for each layer can be mathematically represented as,
\begin{equation}
    W_l^{p} = \pi_l W_l (\pi_{l-1})^\top, \qquad b_l^{p} = \pi_l b_l.
    \label{eqn:permutation}
\end{equation}

\paragraph{Evaluating Permutation Matching.} Since LAP uses a greedy search to find an approximate solution, to ensure that the permuted model $A$ and model $B$ lie in the same basin, we evaluate the \gls{lmc} (loss barrier) between the two models. 
More formally, let $\theta_1, \theta_2$ be the parameters of two networks, then the loss barrier $\mathcal{B}$ is defined as:
% \begin{align}
%     \mathcal{B}(\theta_1, \theta_2) :=  \sup_{\alpha \in [0, 1]} \left[ \mathcal{L}\big((1 - \alpha)\theta_1  + \alpha\theta_2\big) -  \big((1 - \alpha)\mathcal{L}(\theta_1) + \alpha\mathcal{L}(\theta_2)\big) \right]  \geq 0 \ ,
%     \label{eqn:lmc}
% \end{align}
\begin{align}
\mathcal{B}(\theta_1, \theta_2) :=  
& \sup_{\alpha \in [0, 1]} \Big[ \mathcal{L}\big((1 - \alpha)\theta_1  + \alpha\theta_2\big) \nonumber \\ 
& \quad -  \big((1 - \alpha)\mathcal{L}(\theta_1) + \alpha\mathcal{L}(\theta_2)\big) \Big] \geq 0 \ , 
\label{eqn:lmc}
\end{align}

where $\mathcal{L}$ is the loss function evaluated on the training dataset. If $\mathcal{B}(\theta_1, \theta_2) \approx 0$, it is said that $\theta_1$ and $\theta_2$ are linearly mode connected. 

\begin{figure*}[tbp]
    \centering
    \begin{subfigure}{0.3\textwidth}
        \centering
        \includegraphics[width=\linewidth]{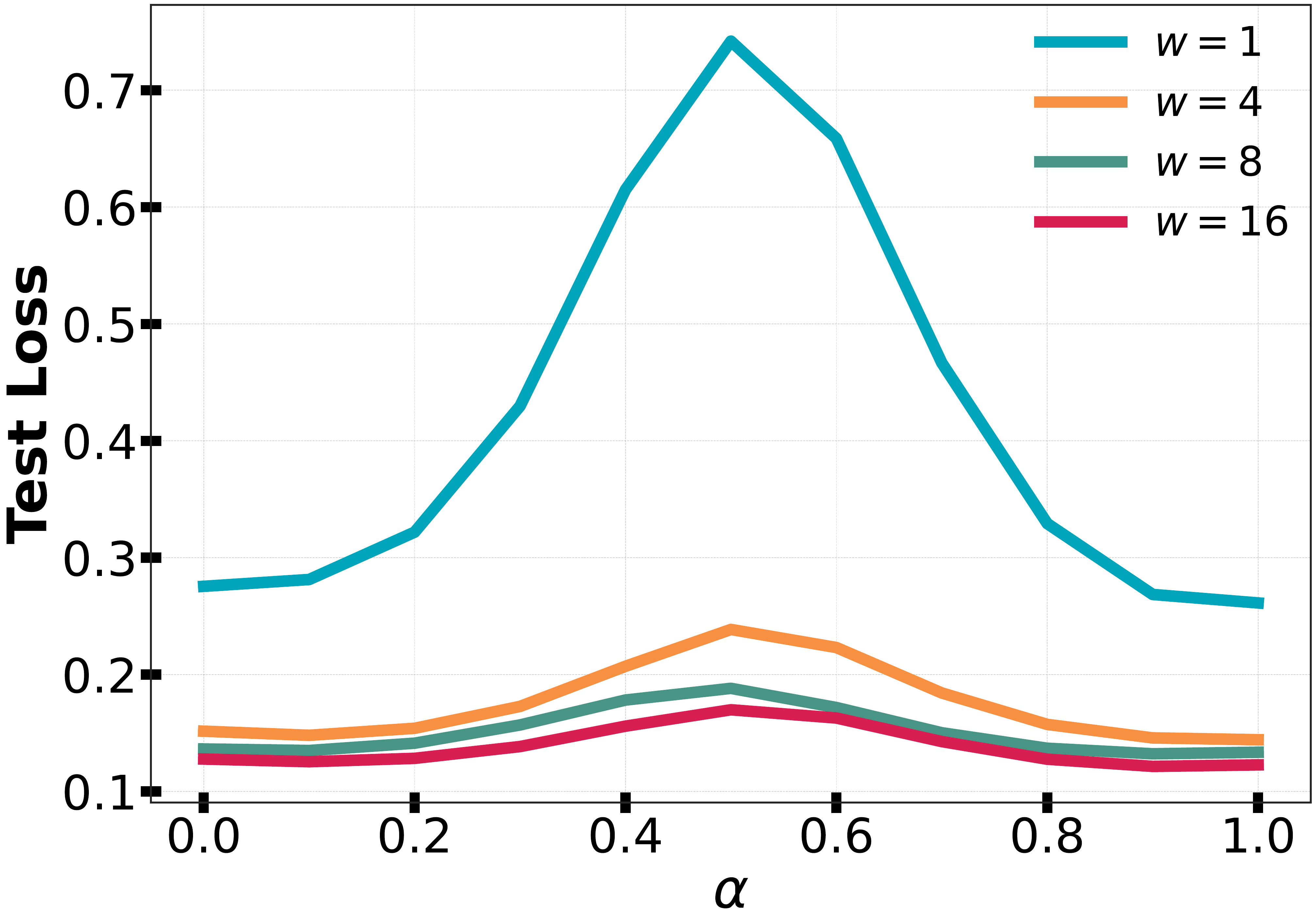}
        \caption{ResNet20$\times\{w\}$/CIFAR-10}
        \label{fig:interp_resnet20_c10}
    \end{subfigure}
    \quad
    \begin{subfigure}{0.3\textwidth}
        \centering
        \includegraphics[width=\linewidth]{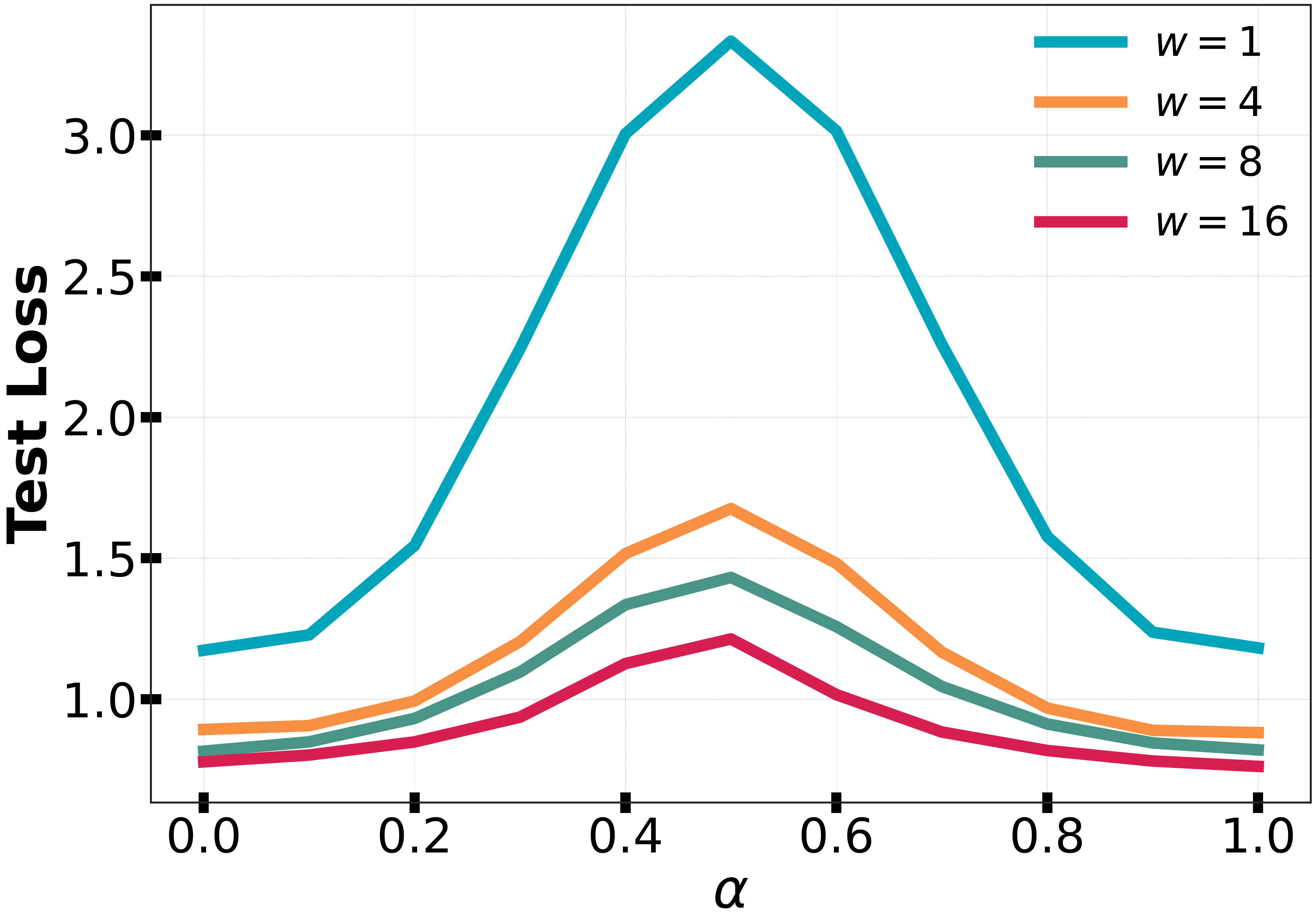}
        \caption{ResNet20$\times\{w\}$/CIFAR-100}
        \label{fig:interp_resnet20_c100}
    \end{subfigure}
    \quad
    \begin{subfigure}{0.3\textwidth}
        \centering
        \includegraphics[width=\linewidth]{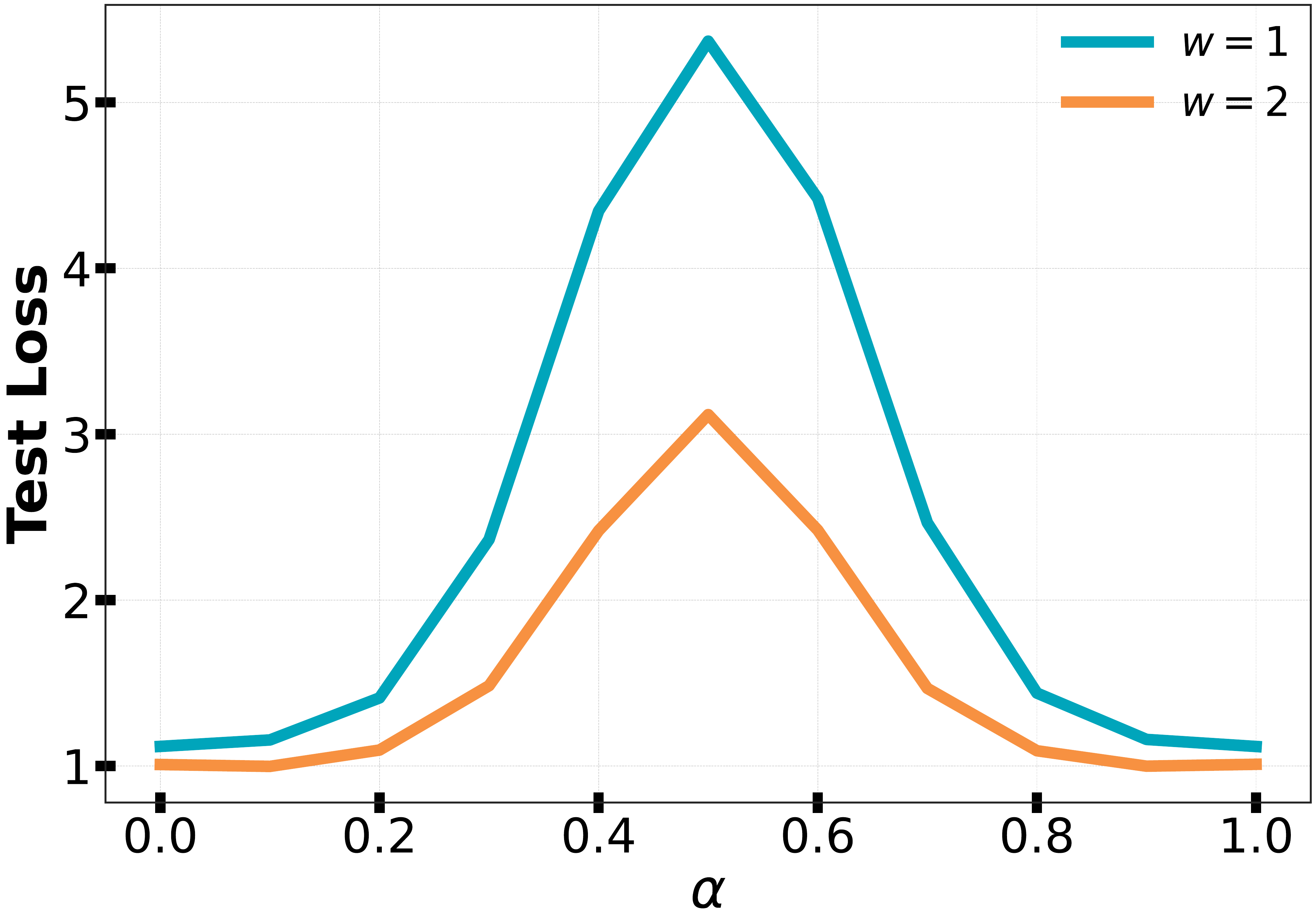}
        \caption{ResNet50$\times\{w\}$/ImageNet}
        \label{fig:interp_resnet50_imagenet}
    \end{subfigure}
    \caption{\textbf{Larger width exhibits better \gls{lmc}.} Plots showing linear interpolation between $\pi(\textbf{w}_A^{t=T})$ and $\textbf{w}_B^{t=T}$ where $\pi$ was obtained through activation matching between two dense models for varying widths, $w$. As the width of the model increases, the permutation matching algorithm gets more accurate, thereby reducing the loss barrier (i.e., better \gls{lmc}), which is evaluated on the test set. This shows that the permutation matching can find a better mapping, $\pi$, for wider models, explaining why the permuted mask works better in the case of wider models. }
    \label{fig:interp_plots}
\end{figure*}

To ensure that the permutation mapping, $\pi$, can closely match model $A$ and model $B$, we evaluate the loss barrier between the permuted model $A$ and model $B$. However, aligning neurons alone is not sufficient to establish a low loss barrier due to variance collapse~\citep{jordan2023repair}. To overcome the variance collapse issue, we used REPAIR~\citep{jordan2023repair} to correct the variance of the activations in the interpolated/merged model.
As shown in~\cref{fig:interp_plots}, the loss barrier after permutation matching and correcting the variance (REPAIR) is lower than the loss at random initialization, showing permutation mapping can match the models to bring them closer/in the loss basin.

\paragraph{Aligning Masks via Weight Symmetry.}
% \citet{ainsworth2023git} showed the permutation symmetries of the weight space can be leveraged to align the basin of two models trained from different random initializations. In their approach, the authors utilize activation matching to align the activations of two models. By permuting the parameters of the second model, they maximize the correlation between the activations of the first and second models. This method fits within the framework of solving a linear assignment problem (LAP), enabling efficient computation.
%The authors proposed to use activation matching to permute the second model parameters such that the correlation between the activation of the first and second models is maximized. 
In contrast to previous works~\citep{ainsworth2023git}, we are interested in permuting the mask obtained by \gls{lth} such that the optimization basin of the permuted sparse mask and the new random initialization is aligned.
To validate our hypothesis, we train two dense models, $\textbf{w}_A^{t=0}$ and $\textbf{w}_B^{t=0}$, where $t$ denotes the epoch, to convergence (trained for $T$ epochs) and then use activation matching~\citep{jordan2023repair} to find the permutation mapping $\pi$, such that the activations of $\pi(\textbf{w}_A^{t=T})$ and $\textbf{w}_B^{t=T}$ are aligned. Mask $\textbf{m}_A$, obtained using \gls{imp}, is also permuted with the same permutation map $\pi$. The intuition is that the permuted mask aligns with the loss basin of model $\textbf{w}_B^{t=T}$, which is necessary for sparse training and, therefore, the sparse model can be more easily optimized (see ~\cref{fig:workflow}). We denote training with the permuted mask, $\pi(\textbf{m}_A)$ as \emph{permuted} and with the non-permuted mask, $\textbf{m}_A$ as \emph{naive}.

\begin{figure}[tbp]
    \centering

    % \begin{subfigure}{0.49\columnwidth}
    %     \centering
    %     \raisebox{0.3em}{\includegraphics[width=\linewidth]{plots/ekansh_plots/lmc_sparse_dense.pdf}}
    %     \caption{ResNet20$\times\{4\}$/CIFAR-10}
    %     \label{fig:lmc-dense-lth}
    % \end{subfigure}
    % \begin{subfigure}{0.49\columnwidth}
    %     \centering
    %     \includegraphics[width=\linewidth]{plots/ekansh_plots/lmc_sparse_landscape.pdf}
    %     \caption{0--1 loss landscape}
    %     \label{fig:lmc-dense-lth-loss-landscape}
    % \end{subfigure}
    \begin{subfigure}{0.51\columnwidth}
        \centering
        \includegraphics[width=\linewidth]{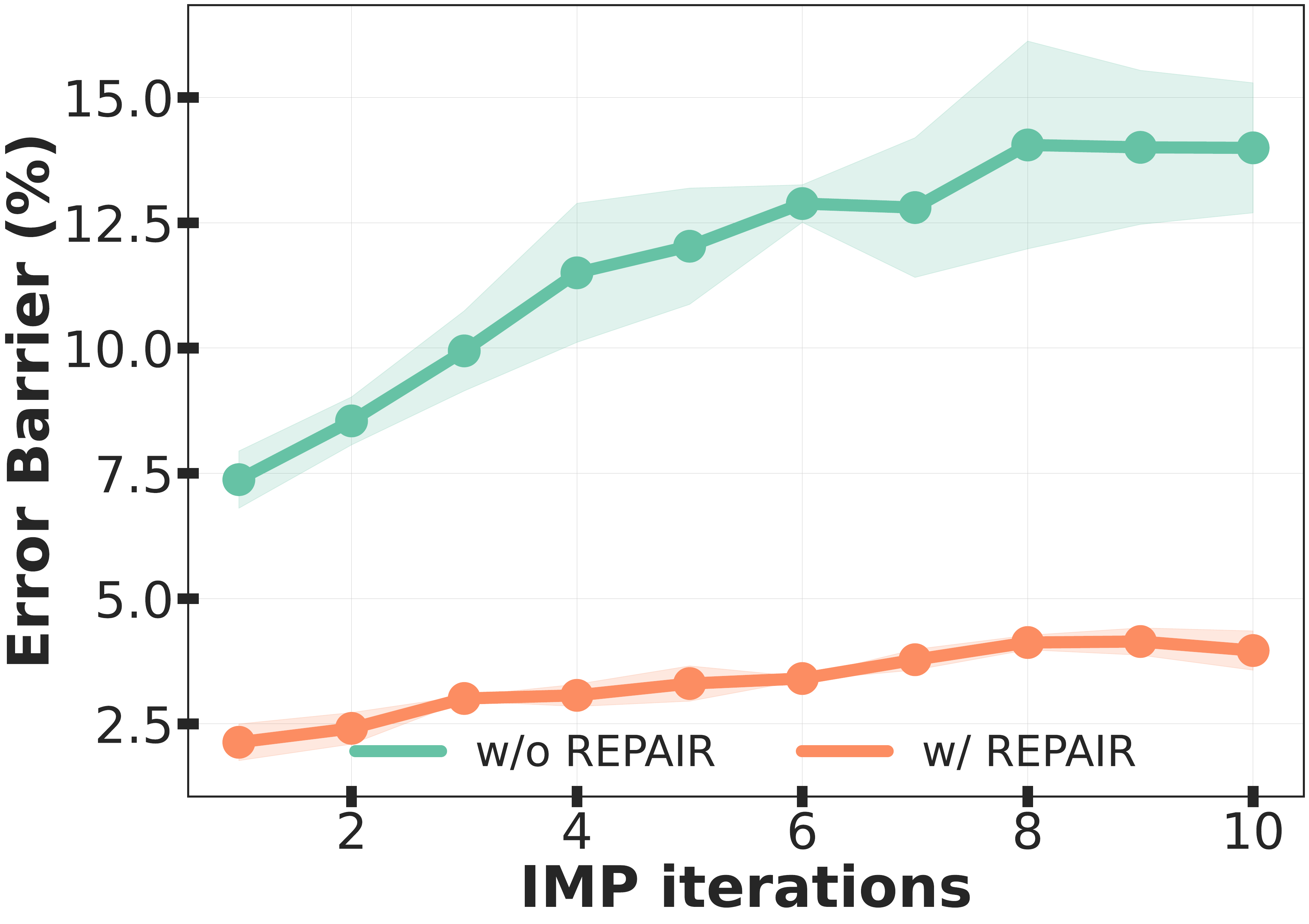}
        \caption{ResNet20$\times\{4\}$/CIFAR-10}
        \label{fig:lmc-dense-lth}
    \end{subfigure}
    \begin{subfigure}{0.48\columnwidth}
        \centering
        \includegraphics[width=\linewidth]{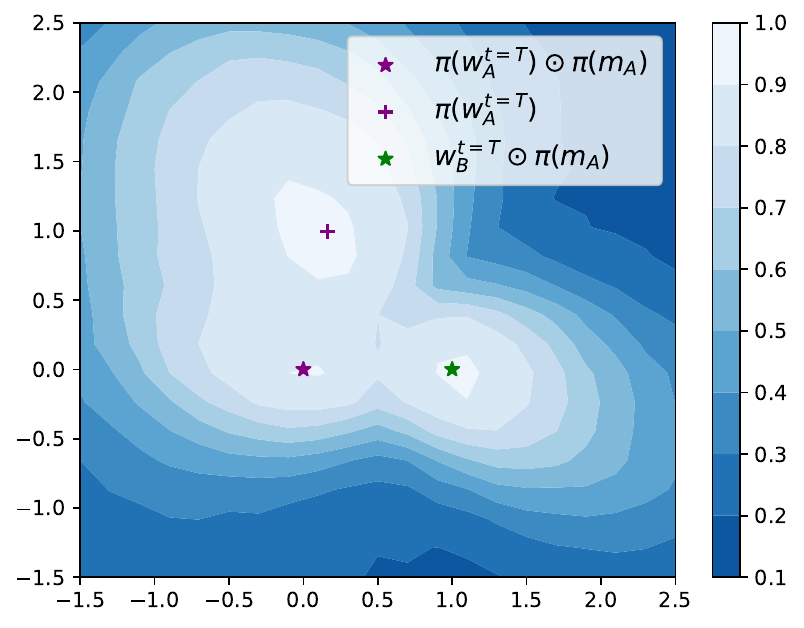}
        \caption{0--1 loss landscape}
        \label{fig:lmc-dense-lth-loss-landscape}
    \end{subfigure}

    \caption{\textbf{\gls{lth} solution remains in the same linearly connected mode as the dense solution.} In \cref{fig:lmc-dense-lth} we plot the error barrier between the dense solution and the sparse solution (y-axis) vs the IMP iteration corresponding to the sparse solution (x-axis), for $90\%$ sparsity. We observe that after fixing variance collapse via the REPAIR method, the error barrier between the dense and the sparse solutions remains small, thus showing that LTH solution remains in the same linearly connected mode as the dense solution.
    In \cref{fig:lmc-dense-lth-loss-landscape} we visualize the 0--1 loss landscape of ResNet20$\times\{4\}$/CIFAR-10. The figure is generated by evaluating the 0--1 loss spanned by three models in the figure. 
    % The sparse model in each of the figures is obtained by weight rewinding to achieve $\approx 90\%$ sparsity. 
    We show that, modulo permutations, reusing the permuted mask leads to convergence in the same mode as the original model, i.e.\ the LTH solution. Hence, there is a small loss barrier between the permuted and LTH solutions, demonstrating they are within the same linearly connected mode. 
    % In the visualizations, lighter and darker regions correspond to lower and higher loss, respectively.
    }
    \label{fig:loss_landscape}
\end{figure}

\paragraph{Sparse Training.}

We first show that the dense solution $\textbf{w}_{A}^{t=T}$ and the LTH solution obtained by training a model with sparse mask $\textbf{m}_A$ remain in the same linearly connected mode if one fixes the variance collapse identified by \citet{jordan2023repair} by updating the activation statistics via the REPAIR method. We show in \cref{fig:lmc-dense-lth}
 the error barrier after applying the REPAIR method remains considerably low as we increase sparsity by iteratively pruning (\gls{imp}). These results extend the findings of \citet{paul2022unmasking} to show that when variance collapse is taken into account, the LTH solution remains in the same linearly connected basin as the original dense solution. 

In sparse training, the model is trained with a mask $\textbf{m}$, masking some of the weights, during both forward and backward passes.
To evaluate the transferability of the permuted LTH mask we train, a different random initialization $\textbf{w}_{B}^{t=0}$, 
the \gls{lth} sparse mask $\textbf{m}_A$ and permuted \gls{lth} mask $\pi(\textbf{m}_A)$, which we denote the naive and permuted solution respectively. 
%Both forward and backward passes are masked by corresponding masks. 
% We use the same set of hyperparameters for a fair comparison. 
We also evaluate the \gls{lth} baseline, i.e., training model $\textbf{w}_A^{t=0}$ with mask $\textbf{m}_A$.
% Since LTH often works only with rewind checkpoints, we use a rewind checkpoint at epoch $t =k$ f where the sparse training starts for each of our solutions.
%Since LTH typically relies on rewind checkpoints, we use a rewind checkpoint from epoch $t = k$ for both the baselines and permuted solutions.
Since LTH requires weight rewinding to an earlier point in training, we also use a rewind checkpoint from epoch 
$t=k \ll T$ for both the baselines and permuted solution.
% \textcolor{red}{adnan todo: write sparse training equation.}
% For a model trained with loss $\mathcal{L}$, mask $m$, the gradients are obtained by masking the weights, both during forward and backward passes. 
\Cref{fig:lmc-dense-lth-loss-landscape} shows that \gls{lth}, dense model $A$ and sparse permuted solutions all lie in the same mode.
% \cref{fig:placeholder_loss2} shows that the permuted \gls{lth} solution is linearly mode connected with the model B trained with the permuted mask. \cref{fig:placeholder_loss3} shows that permuted \gls{lth} solution, permuted model $A$ and permuted sparse solution all lie in the same basin.

\section{Results}
To validate our hypothesis, we trained ResNet20~\citep{he2015deepresiduallearningimage} and VGG11~\citep{simonyan2015deepconvolutionalnetworkslargescale} models on the CIFAR-10/100 datasets~\citep{krizhevsky2009learning} (details in \cref{implementation}) across different levels of sparsity~($S=0.80,0.90,0.95,0.97$). We used ResNet20 with varying widths ($w=1,4,8,16$) to study the effect of increasing width on the permutation matching and, thereby, the performance of the permuted sparse model. We also demonstrate our hypothesis on the large-scale ImageNet dataset~\citep{imagenet} using ResNet50, showing the efficacy of our method across different models and datasets of varying sizes.
%We used the same set of hyper-parameters for training different sparse baselines and our permuted solution (details in ~\cref{optim}).

%Since activation matching is reported to work better with wider models \cite{ainsworth2023git, sharma2024simultaneous}, we evaluate our method of ResNets with varying width multipliers. 
%We train two models with different random initialization to convergence for activation matching.
% Results spreadsheet: https://docs.google.com/spreadsheets/d/1HK0bx0nshfAHbs4DCP8C0YKzzM1n4WoBvFIf2bu1BP8/edit#gid=0
%
% \vspace{-0.5em}
\begin{figure*}[tbp]
    \centering
    % width 1
    \begin{subfigure}{1.5em}
        \makebox[20pt]{\raisebox{50pt}{\rotatebox[origin=c]{90}{$w=1$}}}%
    \end{subfigure}
    \begin{subfigure}{0.235\textwidth}
        \centering
        \includegraphics[width=\linewidth]{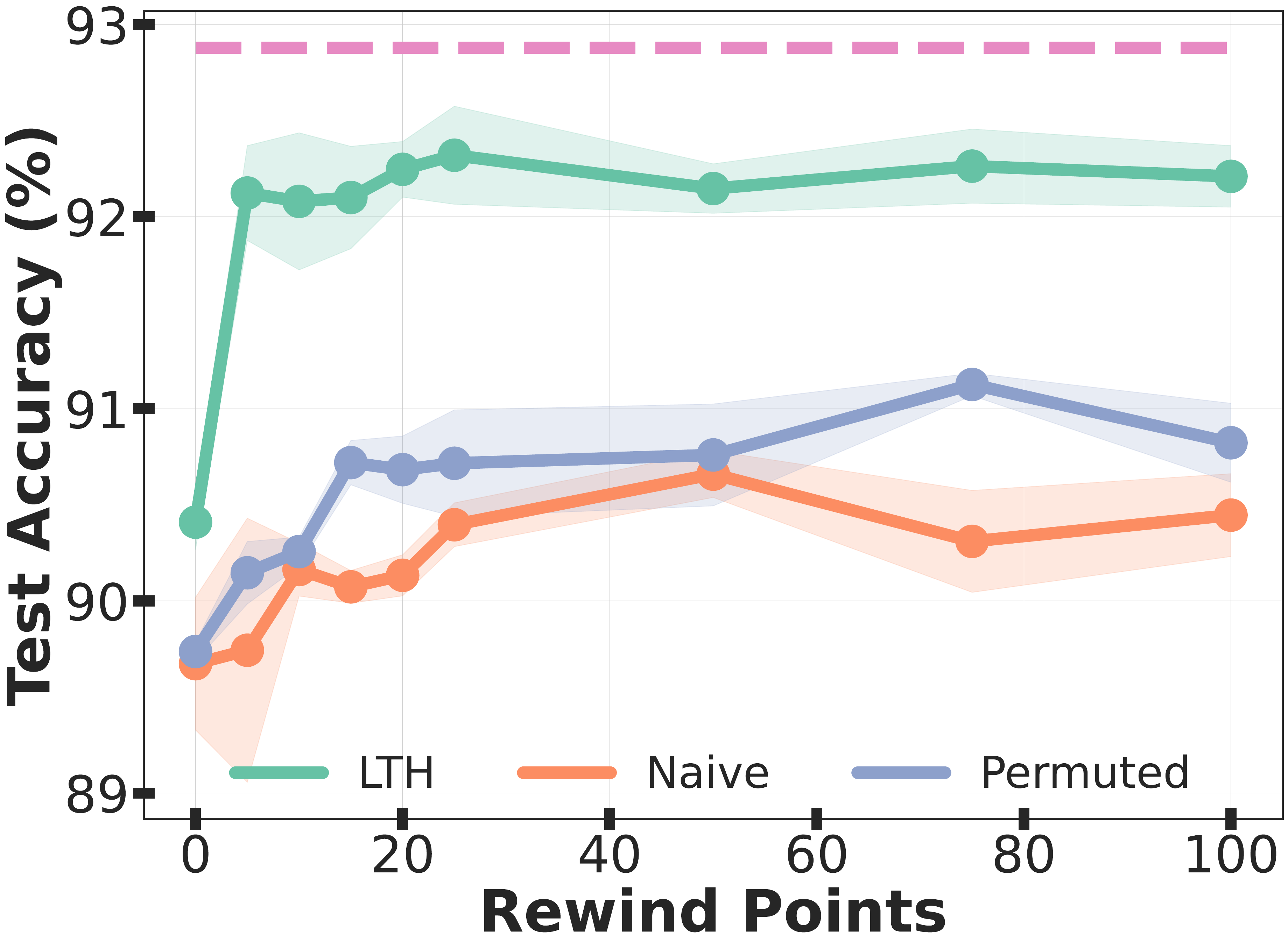}  
        % \caption{sparsity = 0.80}
        % \label{fig:resnet_c10_w1_s80:1}
    \end{subfigure}
    \begin{subfigure}{0.235\textwidth}
        \centering
        \includegraphics[width=\linewidth]{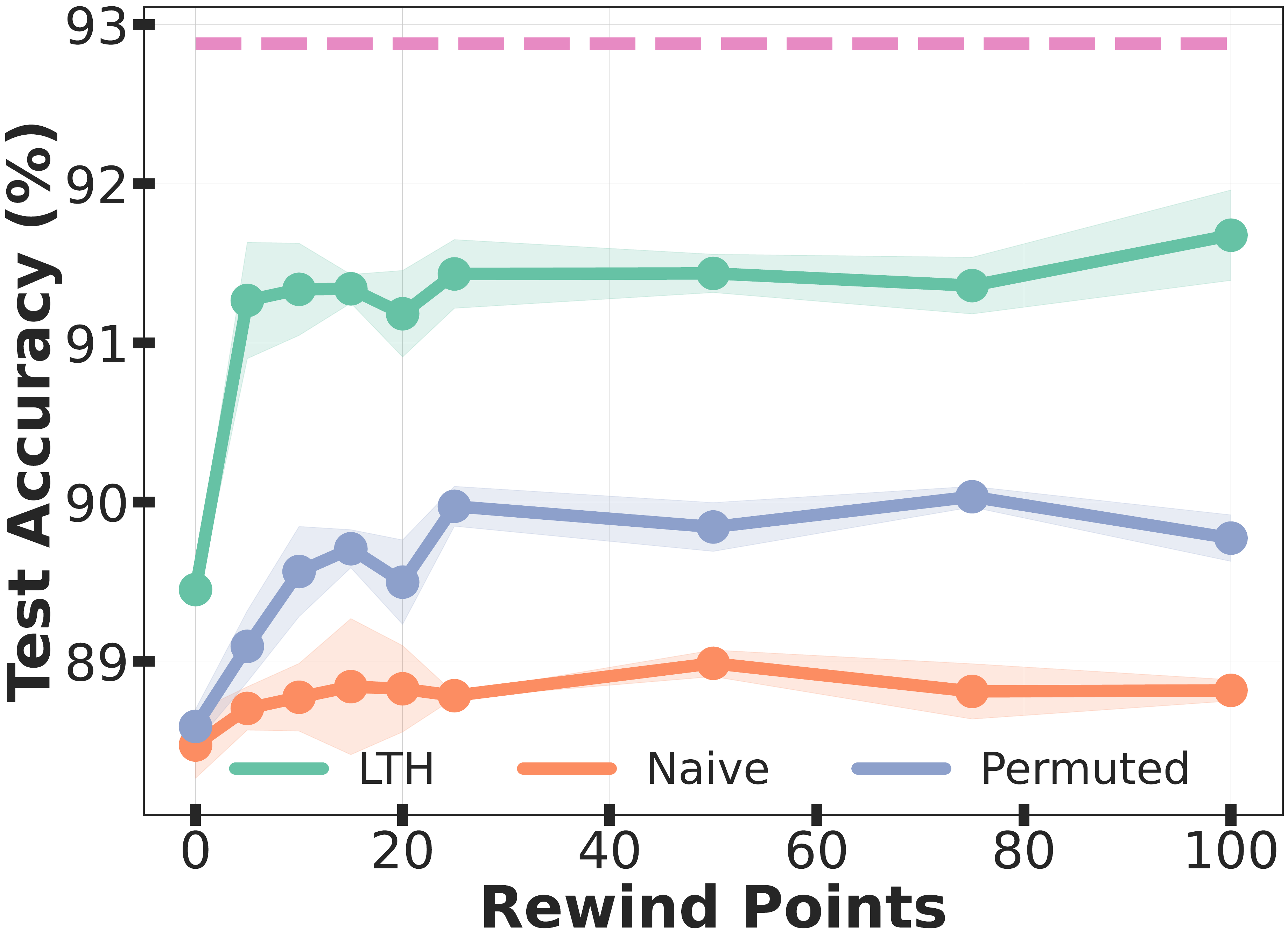}  
        % \caption{sparsity = 0.90}
        % \label{fig:resnet_c10_w1_s90:2}
    \end{subfigure}
    \begin{subfigure}{0.235\textwidth}
        \centering
        \includegraphics[width=\linewidth]{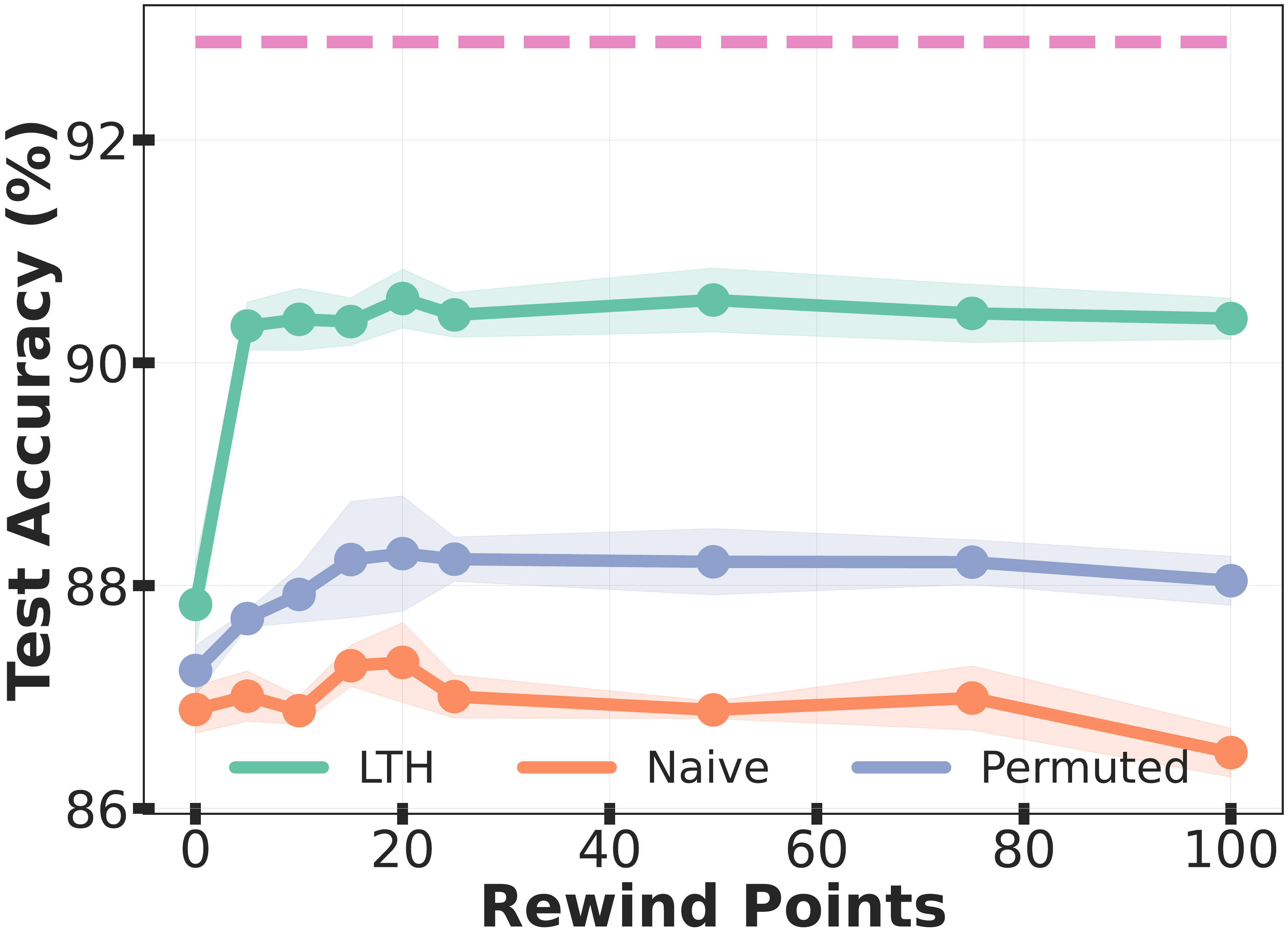}  
        % \caption{sparsity = 0.95}
        % \label{fig:resnet_c10_w1_s95:3}
    \end{subfigure}
    \begin{subfigure}{0.235\textwidth}
        \centering
        \includegraphics[width=\linewidth]{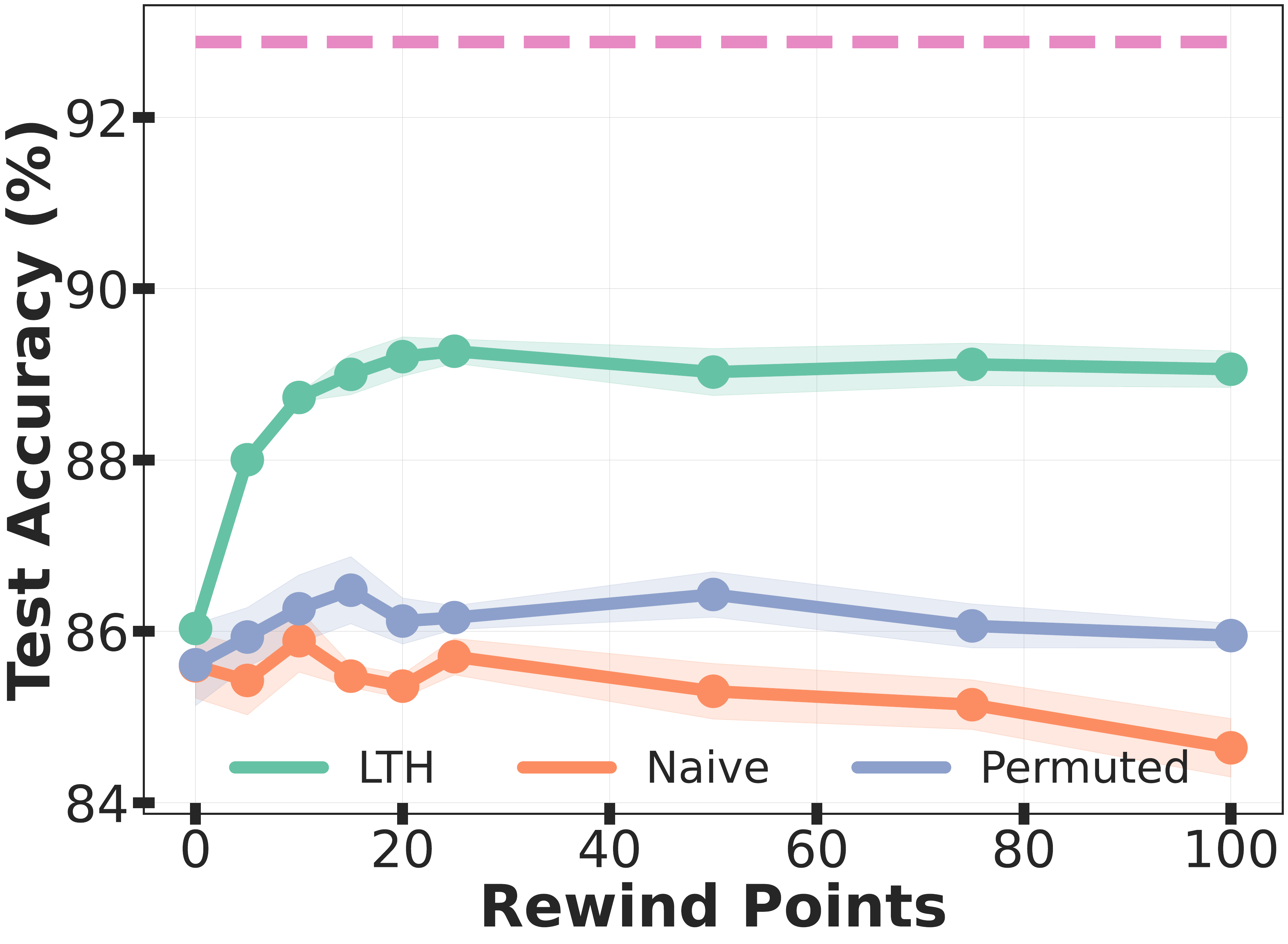}  
        % \caption{sparsity = 0.97}
        % \label{fig:resnet_c10_w1_s97:4}
    \end{subfigure}\\
%
    % width 4
    \begin{subfigure}{1.5em}
        \makebox[20pt]{\raisebox{50pt}{\rotatebox[origin=c]{90}{$w=4$}}}%
    \end{subfigure}
    \begin{subfigure}{0.235\textwidth}
        \centering
        \includegraphics[width=\linewidth]{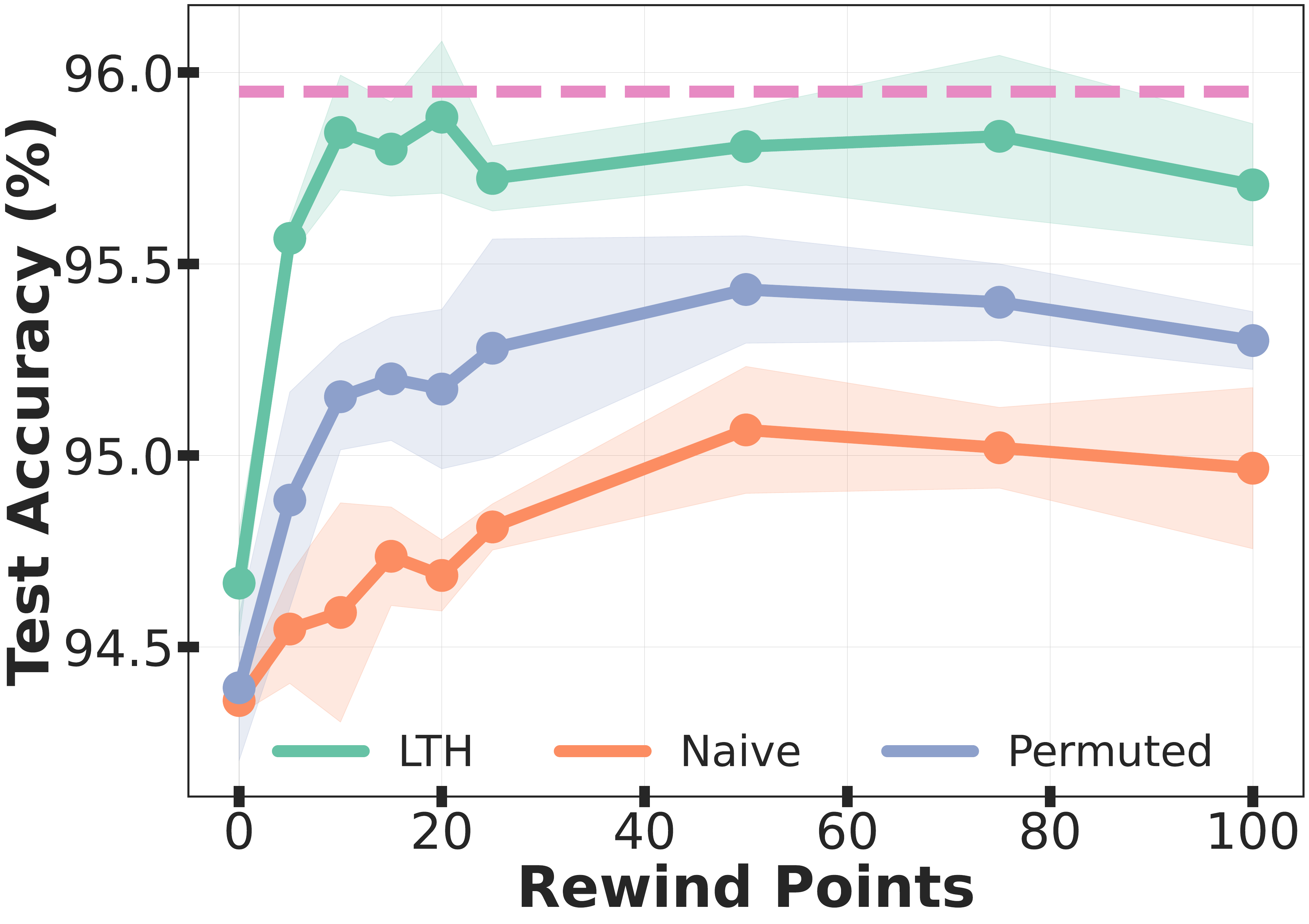}
        % \caption{sparsity = 0.80}
        % \label{fig:resnet_c10_w4_s80:1}
    \end{subfigure}
    \begin{subfigure}{0.235\textwidth}
        \centering
        \includegraphics[width=\linewidth]{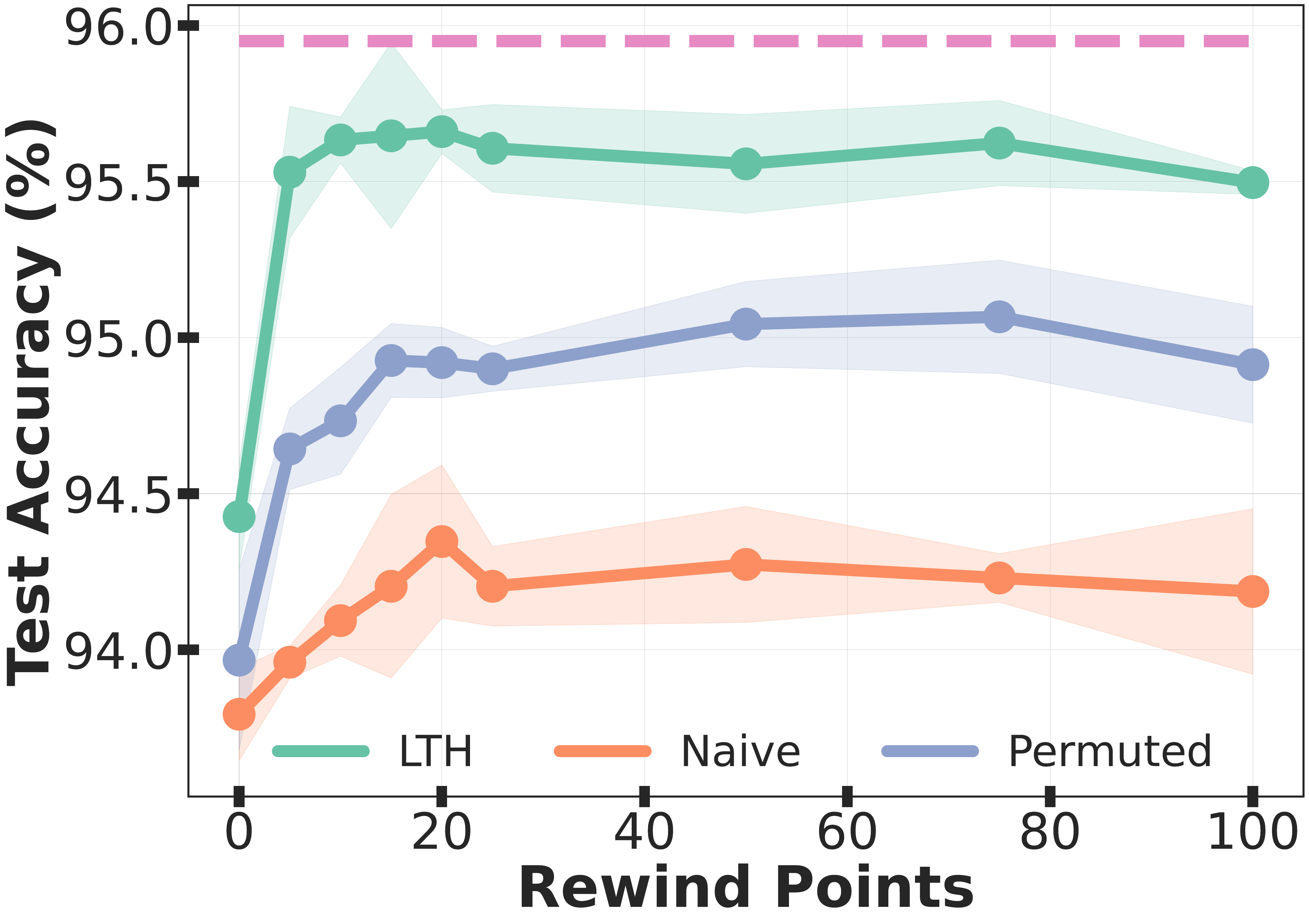}
        % \caption{sparsity = 0.90}
        % \label{fig:resnet_c10_w4_s90:2}
    \end{subfigure}
    \begin{subfigure}{0.235\textwidth}
        \centering
        \includegraphics[width=\linewidth]{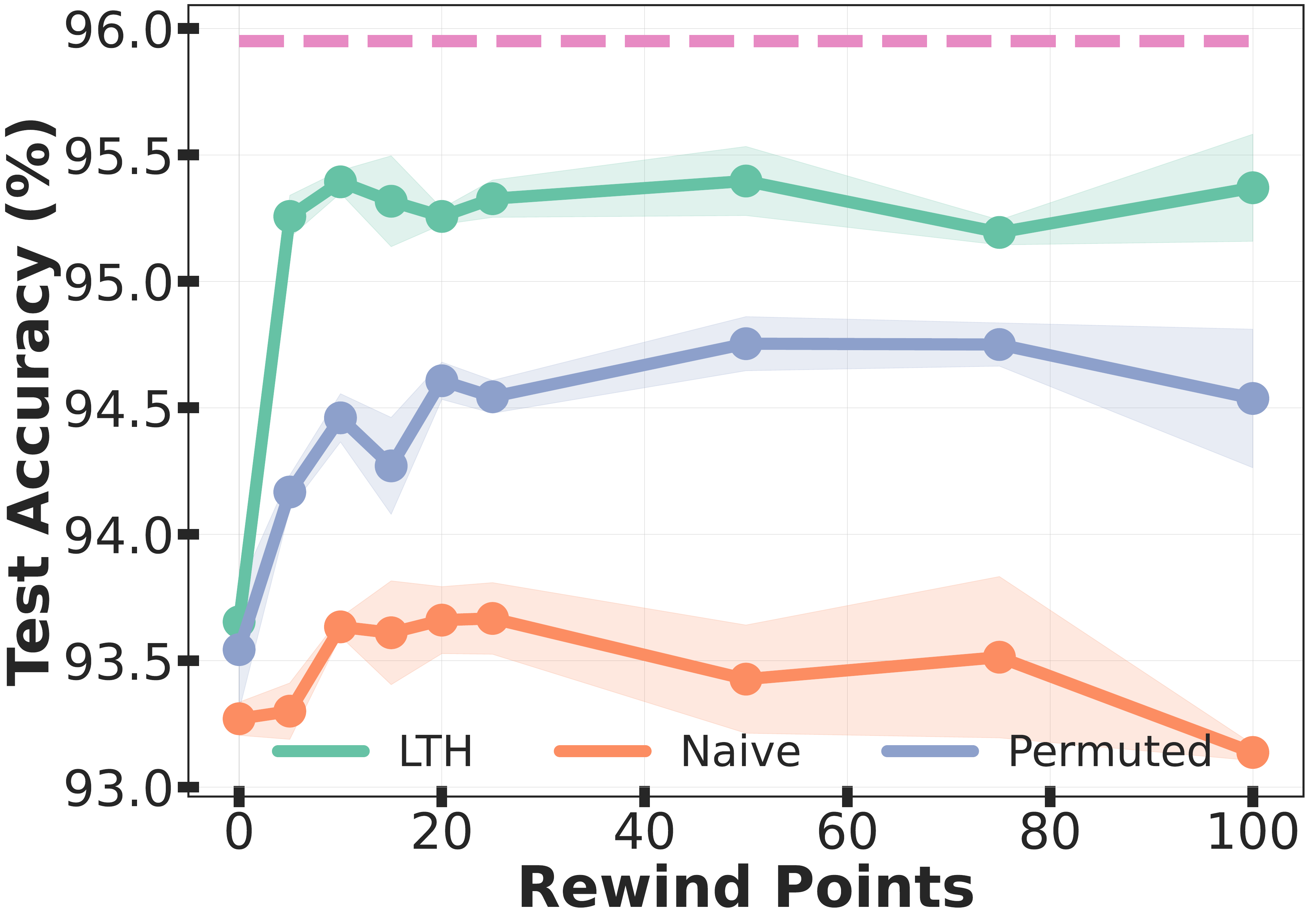}
        % \caption{sparsity = 0.95}
        % \label{fig:resnet_c10_w4_s95:3}
    \end{subfigure}
    \begin{subfigure}{0.235\textwidth}
        \centering
        \includegraphics[width=\linewidth]{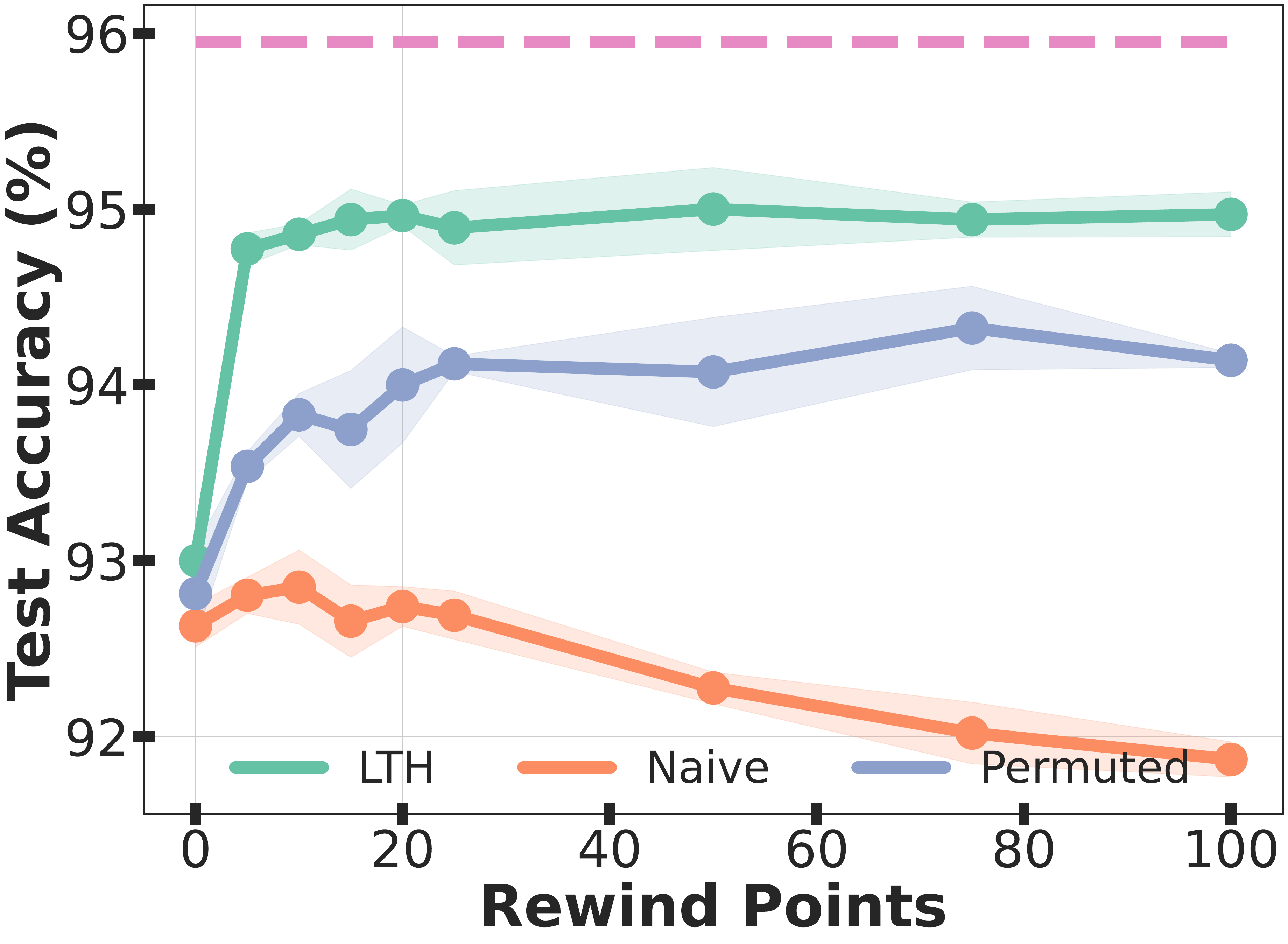}
        % \caption{sparsity = 0.97}
        % \label{fig:resnet_c10_w4_s97:4}
    \end{subfigure}\\
%    
    % width 8
    \begin{subfigure}{1.5em}
        \makebox[20pt]{\raisebox{50pt}{\rotatebox[origin=c]{90}{$w=8$}}}%
    \end{subfigure}
    \begin{subfigure}{0.235\textwidth}
        \centering
        \includegraphics[width=\linewidth]{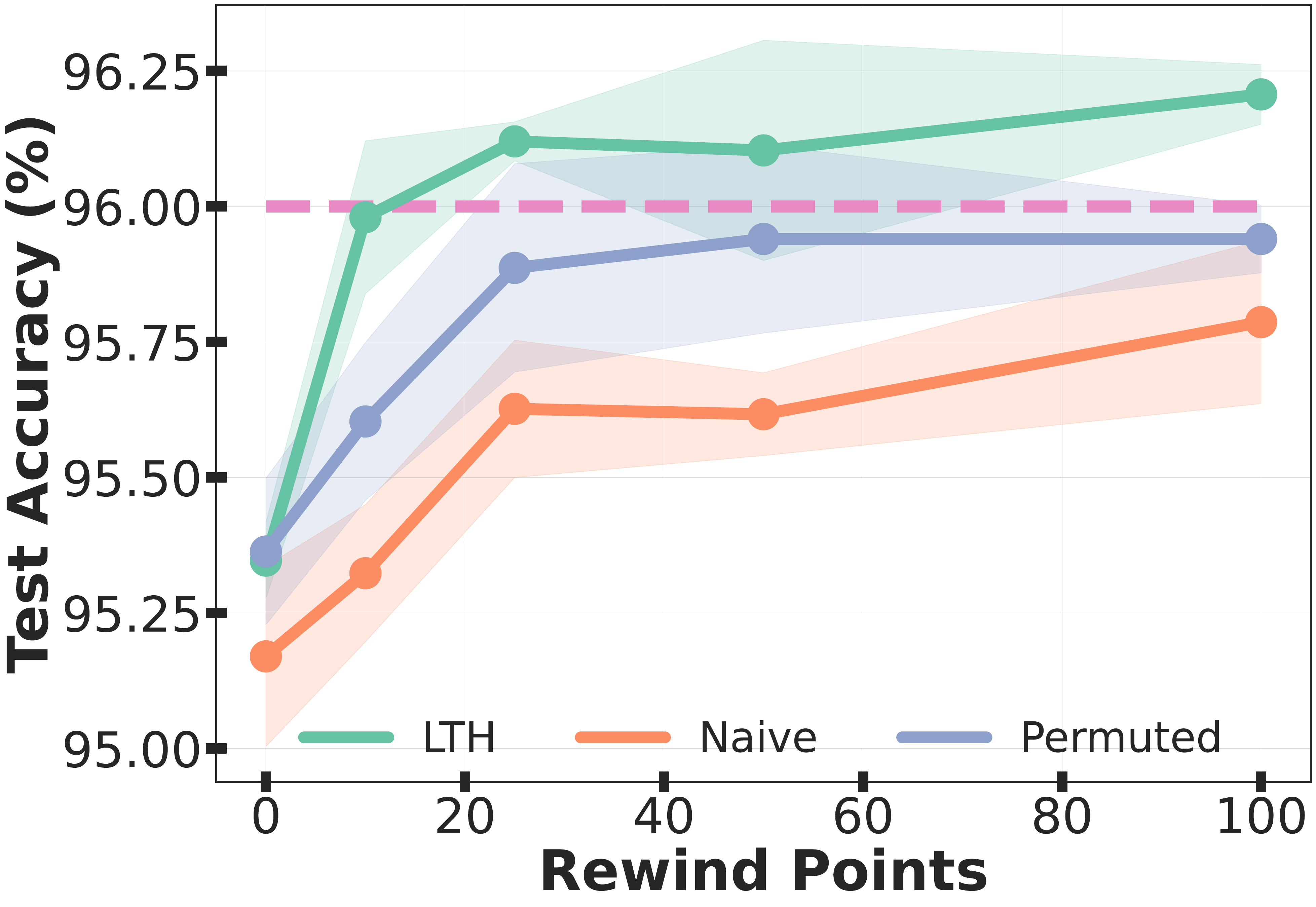}
        % \caption{sparsity = 0.80}
        % \label{fig:resnet_c10_w8_s80:1}
    \end{subfigure}
    \begin{subfigure}{0.235\textwidth}
        \centering
        \includegraphics[width=\linewidth]{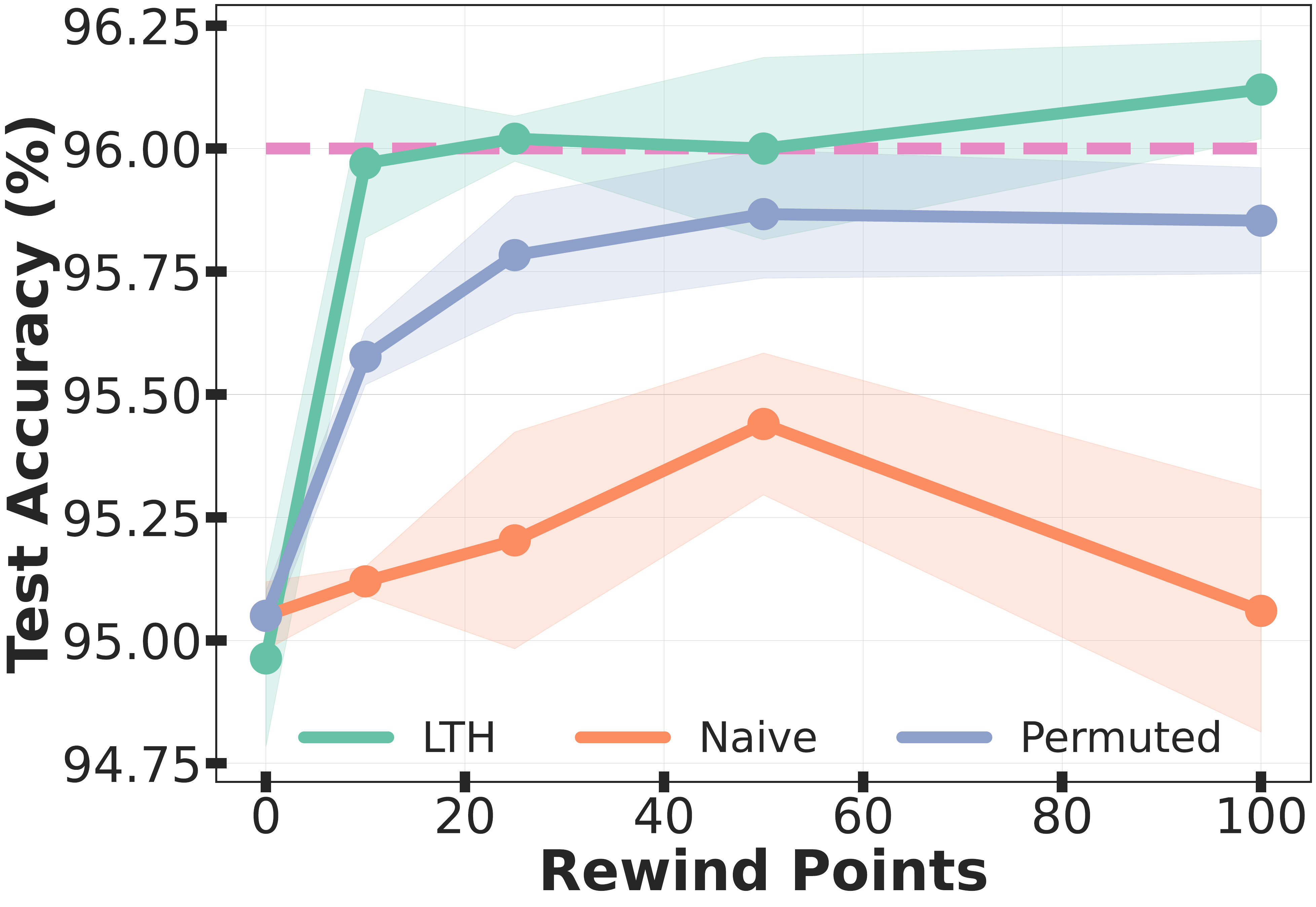}
        % \caption{sparsity = 0.90}
        % \label{fig:resnet_c10_w8_s90:2}
    \end{subfigure}
    \begin{subfigure}{0.235\textwidth}
        \centering
        \includegraphics[width=\linewidth]{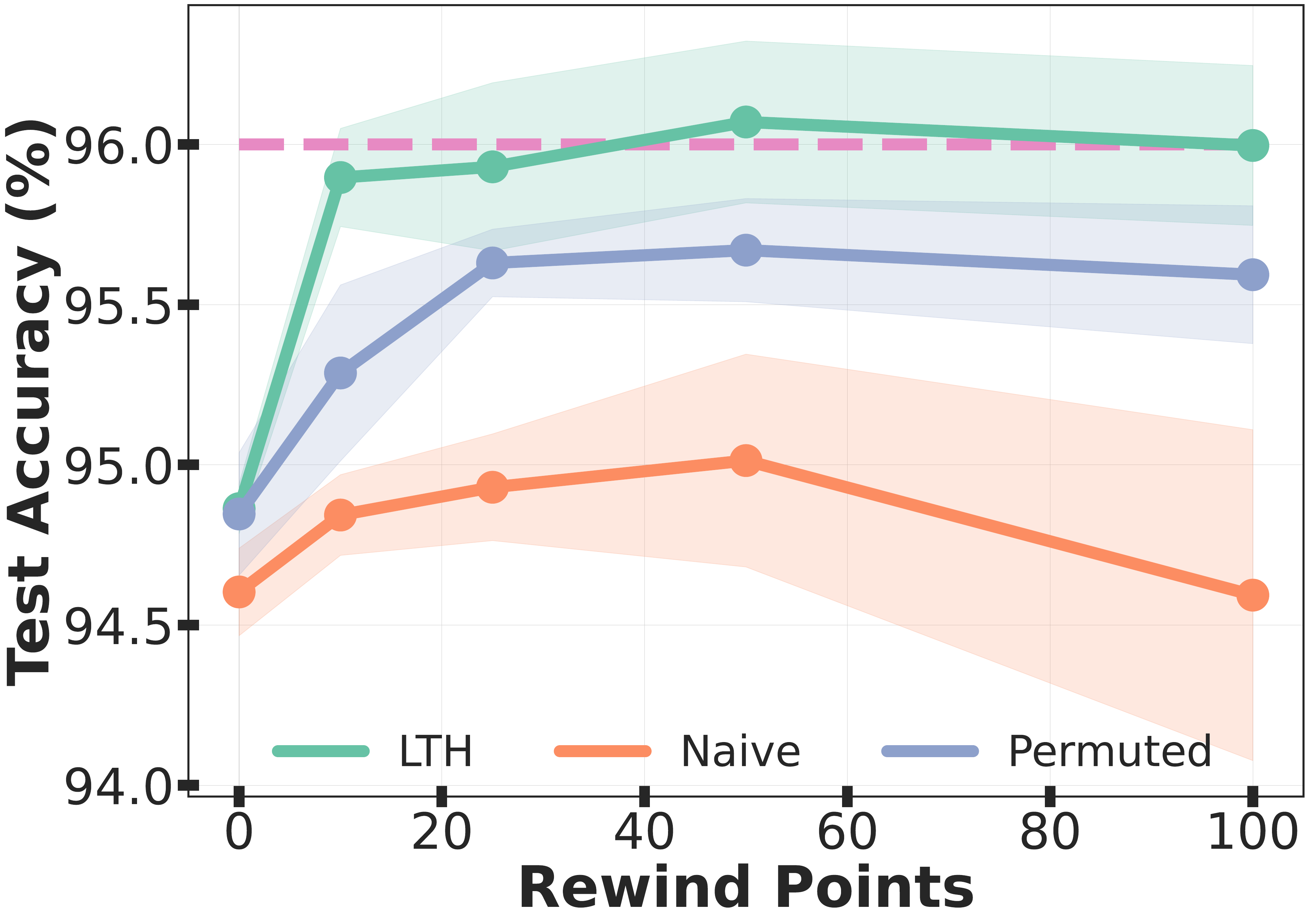}
        % \caption{sparsity = 0.95}
        % \label{fig:resnet_c10_w8_s95:3}
    \end{subfigure}
    \begin{subfigure}{0.235\textwidth}
        \centering
        \includegraphics[width=\linewidth]{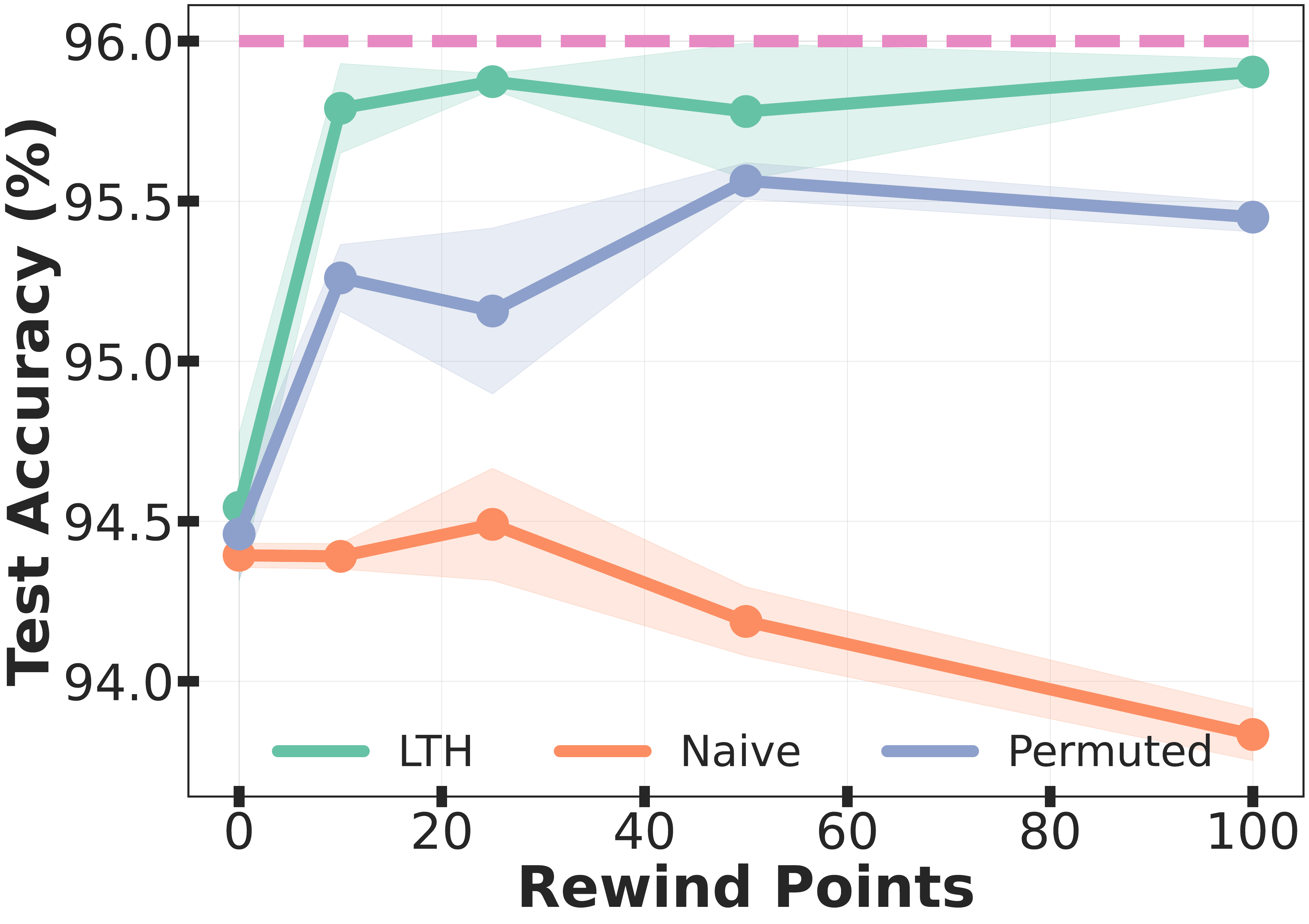}
        % \caption{sparsity = 0.97}
        % \label{fig:resnet_c10_w8_s97:4}
    \end{subfigure}

    % width 16
    \begin{subfigure}{1.5em}
        \makebox[20pt]{\raisebox{50pt}{\rotatebox[origin=c]{90}{$w=16$}}}%
    \end{subfigure}
    \begin{subfigure}{0.235\textwidth}
        \centering
        \includegraphics[width=\linewidth]{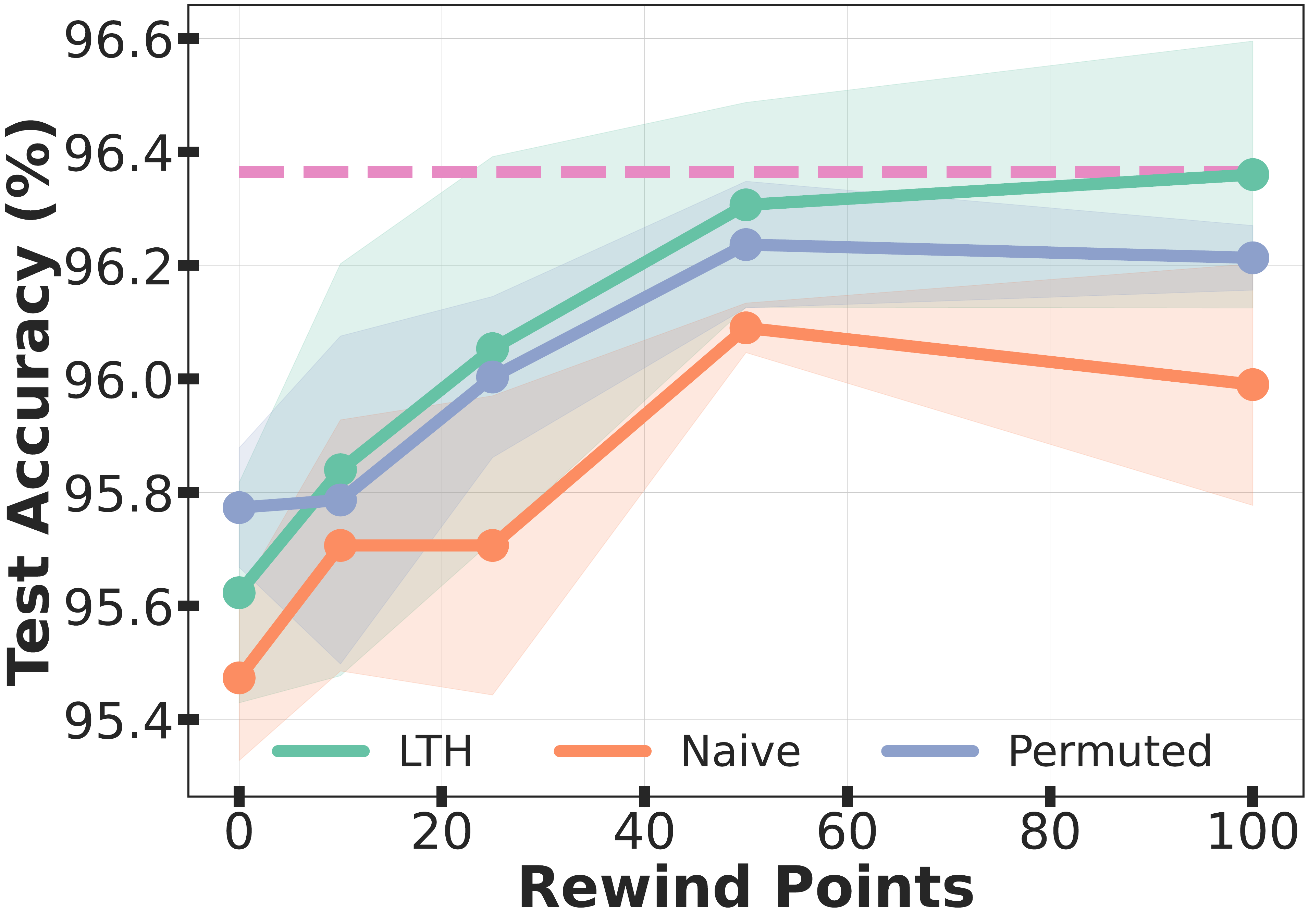}
        \caption{sparsity = 0.80}
        % \label{fig:resnet_c10_w16_s80:1}
        \label{fig:resnet_c10_s80}
    \end{subfigure}
    \begin{subfigure}{0.235\textwidth}
        \centering
        \includegraphics[width=\linewidth]{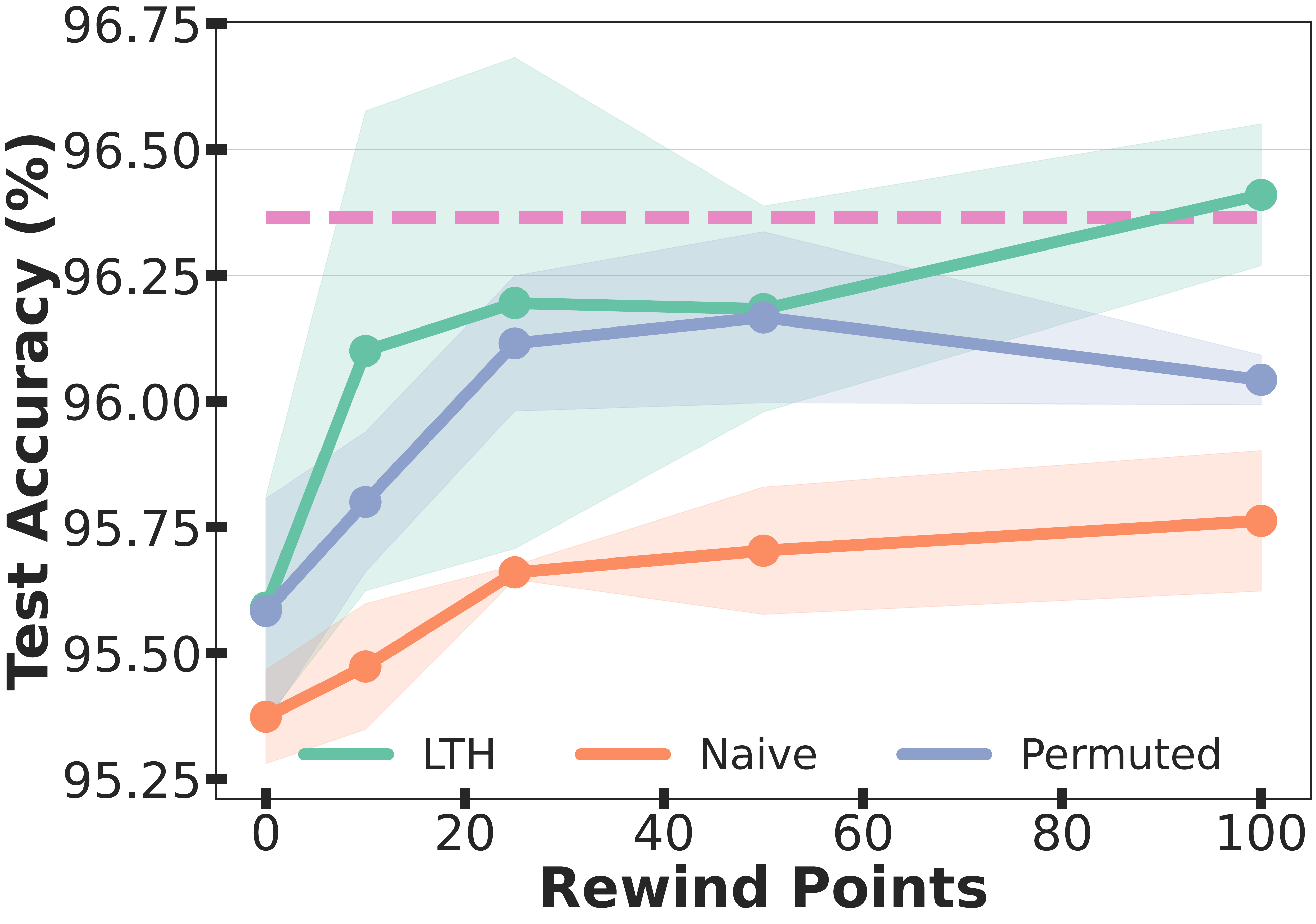}
        \caption{sparsity = 0.90}
        % \label{fig:resnet_c10_w16_s90:2}
        \label{fig:resnet_c10_s90}
    \end{subfigure}
    \begin{subfigure}{0.235\textwidth}
        \centering
        \includegraphics[width=\linewidth]{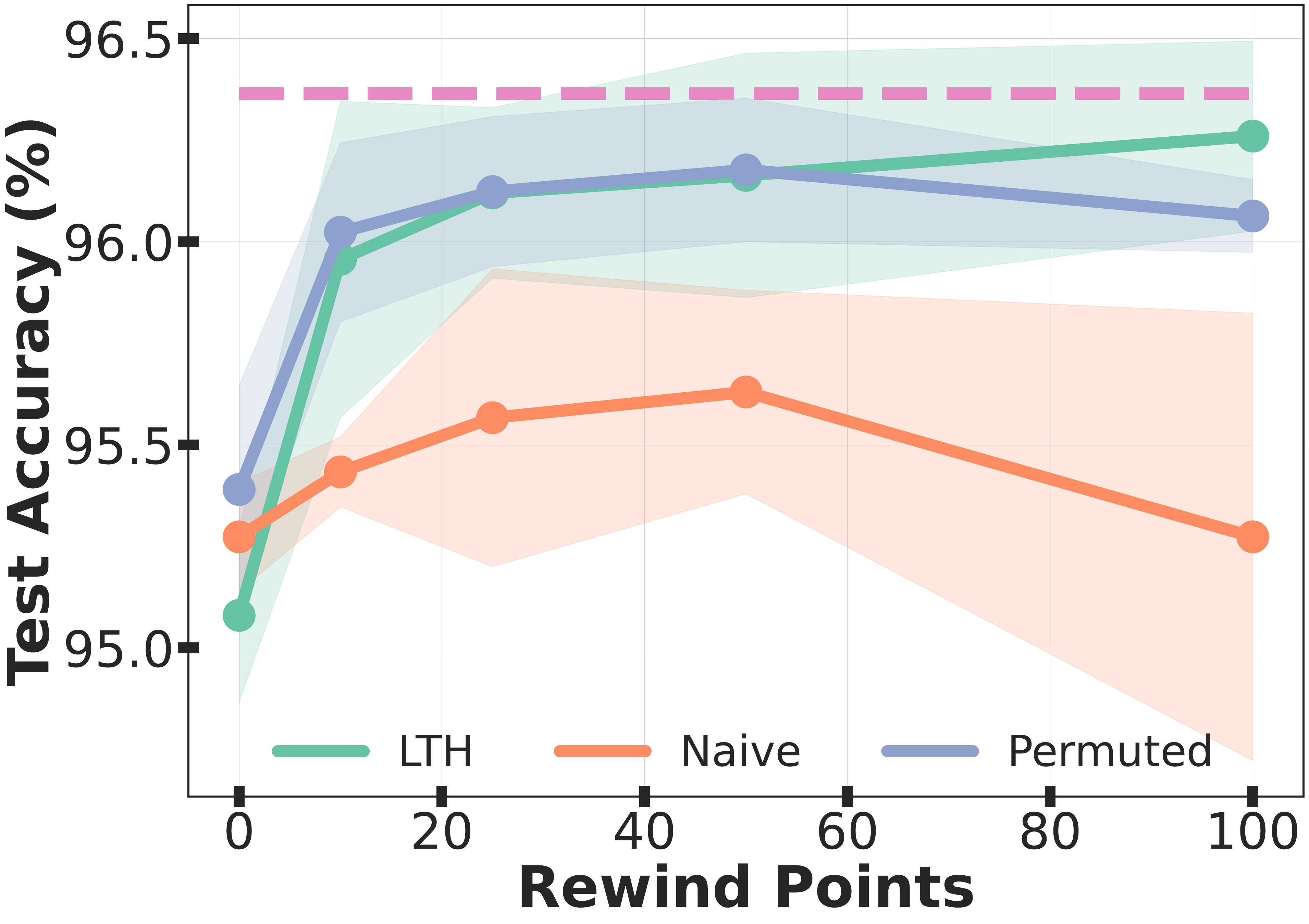}
        \caption{sparsity = 0.95}
        % \label{fig:resnet_c10_w16_s95:3}
        \label{fig:resnet_c10_s95}
    \end{subfigure}
    \begin{subfigure}{0.235\textwidth}
        \centering
        \includegraphics[width=\linewidth]{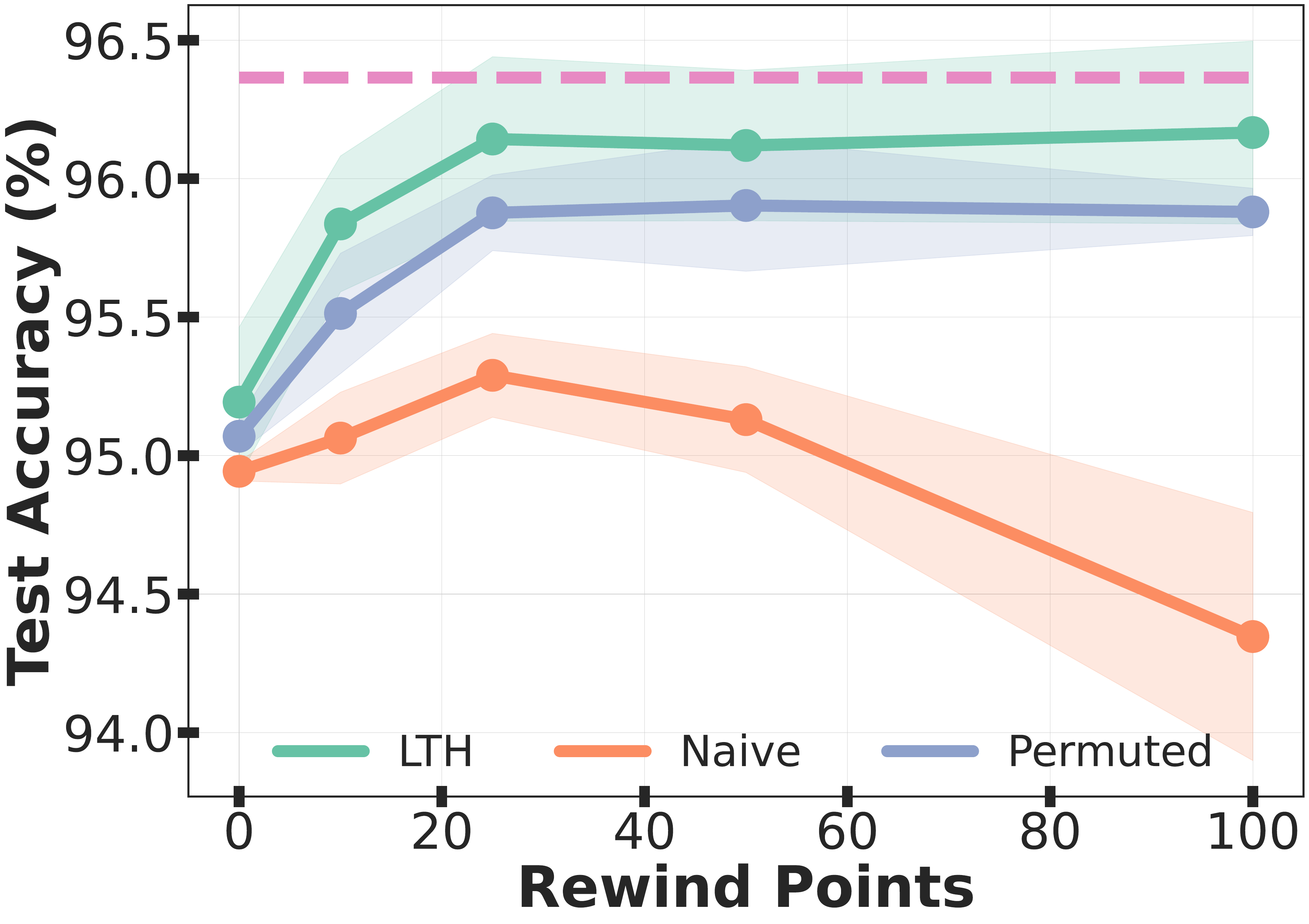}
        \caption{sparsity = 0.97}
        % \label{fig:resnet_c10_w16_s97:4}
        \label{fig:resnet_c10_s97}
    \end{subfigure}
% 
    % \begin{subfigure}{0.235\textwidth}
    %     \centering
    %     \hspace{2.5em} \small sparsity = 0.80
    % \end{subfigure}
    % \begin{subfigure}{0.235\textwidth}
    %     \centering
    %     \hspace{2.5em} \small  sparsity = 0.90
    % \end{subfigure}
    % \begin{subfigure}{0.235\textwidth}
    %     \centering
    %     \hspace{2.5em} \small  sparsity = 0.95
    % \end{subfigure}
    % \begin{subfigure}{0.235\textwidth}
    %     \centering
    %     \hspace{2.5em} \small  sparsity = 0.97
    % \end{subfigure}
    \caption{\textbf{ResNet20$\times\{w\}$/CIFAR-10}.Test accuracy of sparse network solutions vs. increasing rewind points for different sparsity levels and widths, $w$. The dashed ({\textbf{- -}}) line shows the dense model accuracy. The effect of the rewind point on the test accuracy for different sparsities is shown. As the width increases, the gap between training from a random initialization with the permuted mask and the LTH/dense baseline (dashed line) decreases, unlike training with the non-permuted mask (naive), showing a model trained with the permuted mask generalizes better than naive.}
    \label{fig:cifar10_allwidth_rewind}
\end{figure*}

\begin{figure*}[tbp]
    \centering
    % First row, first figure
    \begin{subfigure}{1.5em}
        \makebox[20pt]{\raisebox{50pt}{\rotatebox[origin=c]{90}{$w=1$}}}
    \end{subfigure}
    \begin{subfigure}{0.235\textwidth}
        \centering
        \includegraphics[width=\linewidth]{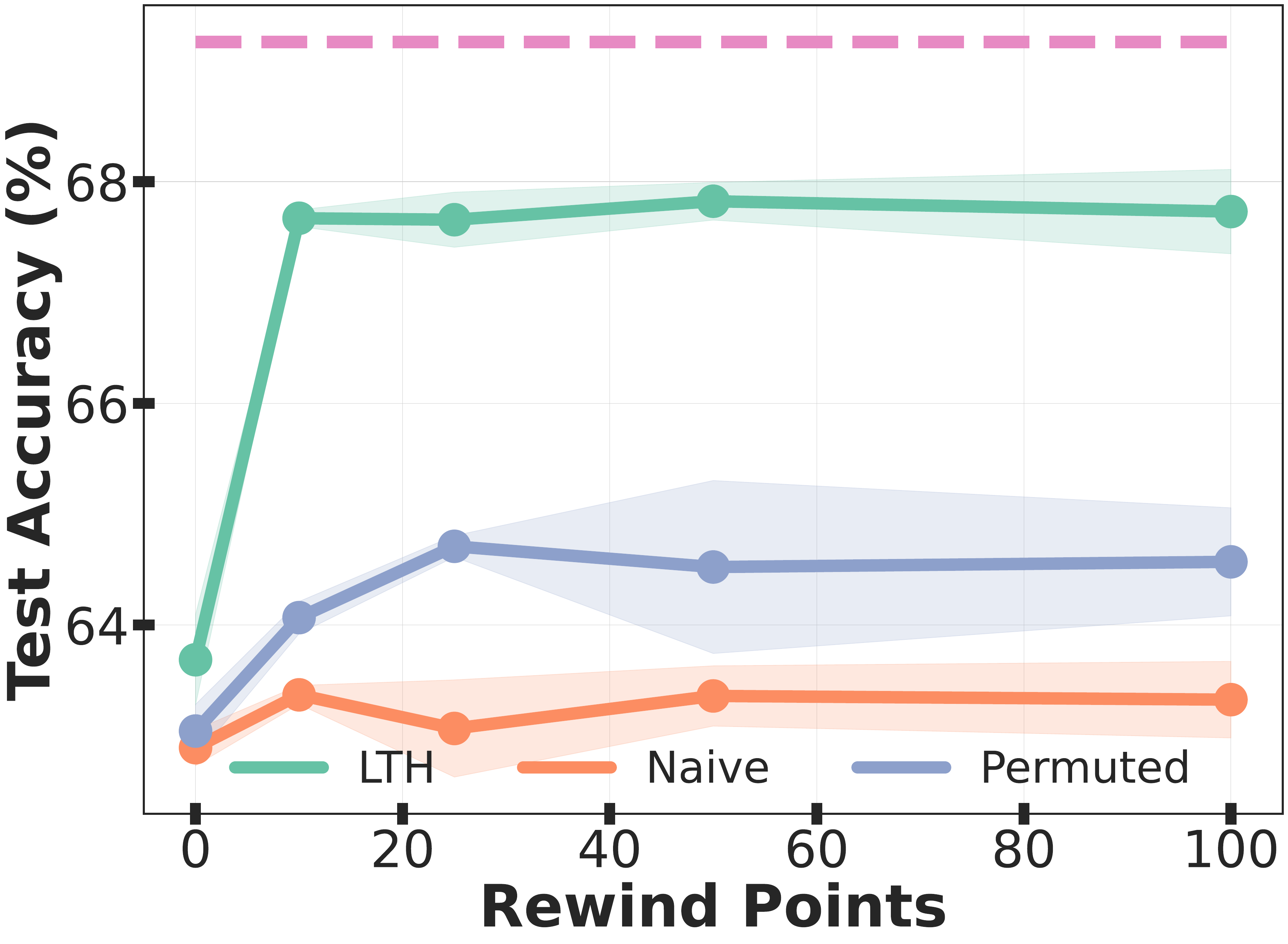}  % Replace with your image path
        % \caption{sparsity = 0.80}
        % \label{fig:resnet20_w1_sp_0.8_fig:1}
    \end{subfigure}
     % First row, second figure
    \begin{subfigure}{0.235\textwidth}
        \centering
        \includegraphics[width=\linewidth]{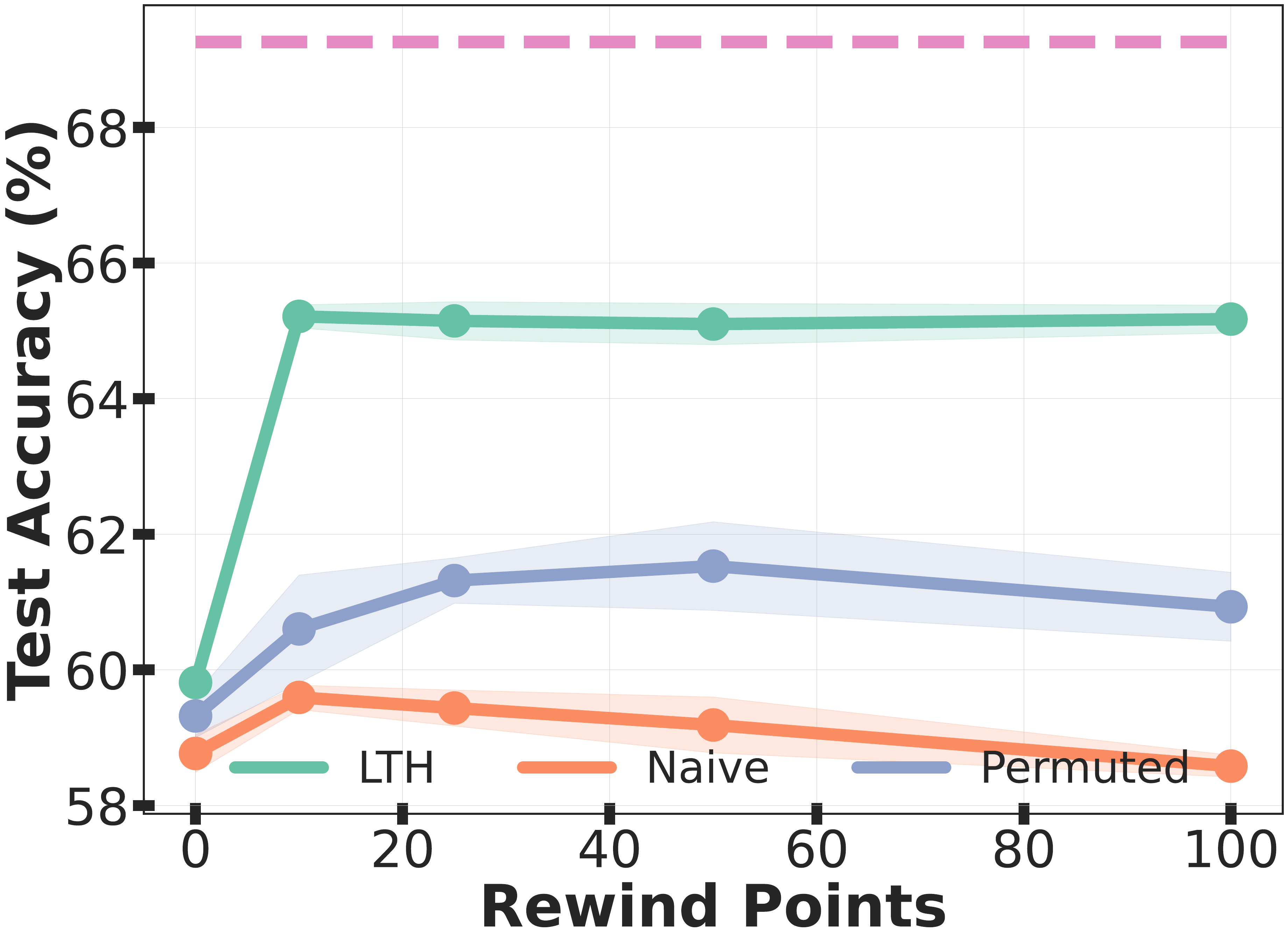}  % Replace with your image path
        % \caption{sparsity = 0.90}
        % \label{fig:resnet20_w1_sp_0.90_fig:1}
    \end{subfigure}
    \begin{subfigure}{0.235\textwidth}
        \centering
        \includegraphics[width=\linewidth]{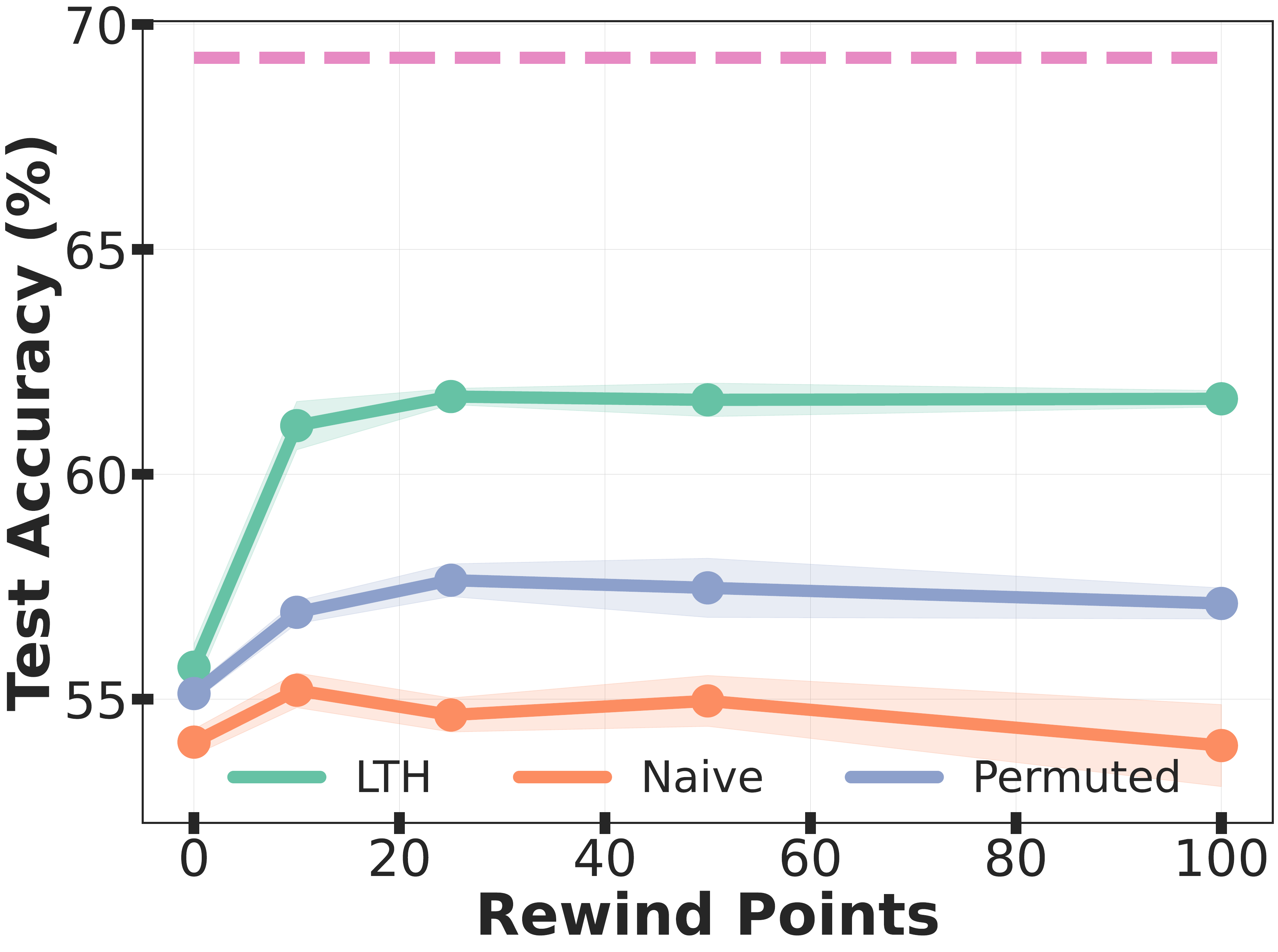}  % Replace with your image path
        % \caption{sparsity = 0.95}
        % \label{fig:resnet20_w1_sp_0.95_fig:1}
    \end{subfigure}
    % Second row, second figure
    \begin{subfigure}{0.235\textwidth}
        \centering
        \includegraphics[width=\linewidth]{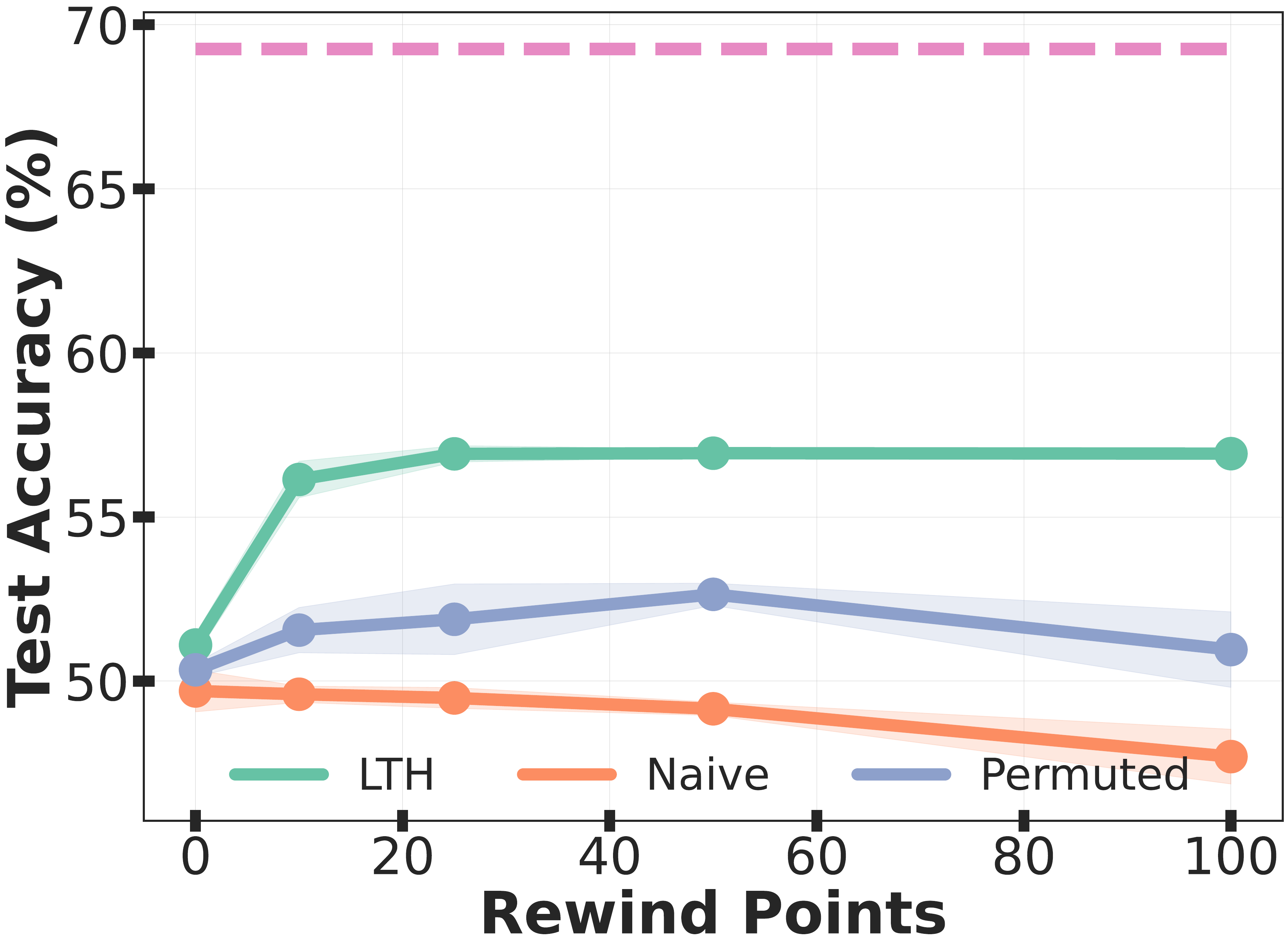}  % Replace with your image path
        % \caption{sparsity = 0.97}
        % \label{fig:resnet20_w1_sp_0.97_fig:1}
    \end{subfigure}\\%
    \begin{subfigure}{1.5em}
        \makebox[20pt]{\raisebox{50pt}{\rotatebox[origin=c]{90}{$w=4$}}}%
    \end{subfigure}
    \begin{subfigure}{0.235\textwidth}
        \centering
        \includegraphics[width=\linewidth]{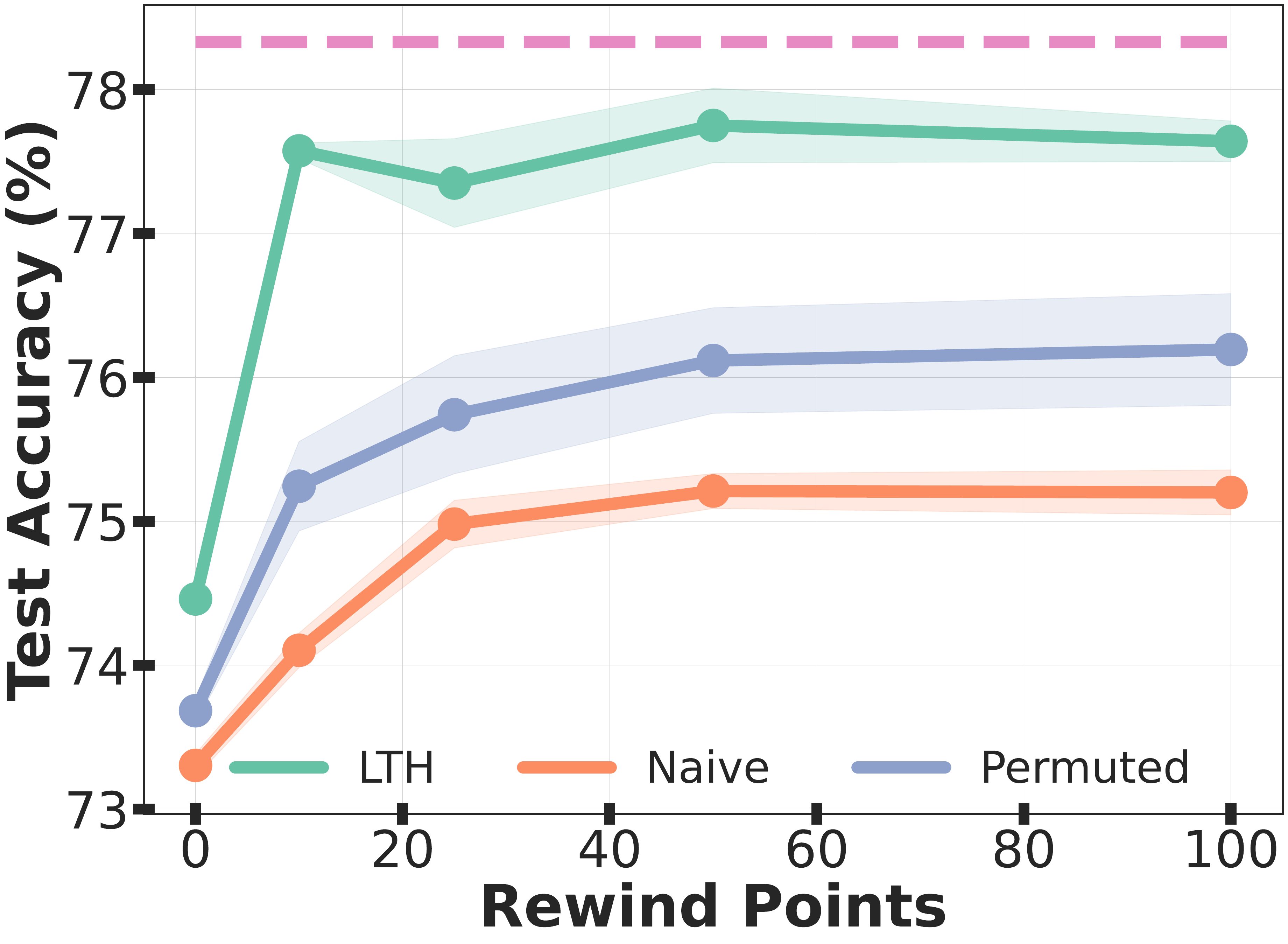}  % Replace with your image path
        % \caption{sparsity = 0.80}
        % \label{fig:resnet_w4_sp_fig:1}
    \end{subfigure}
     % First row, second figure
    \begin{subfigure}{0.235\textwidth}
        \centering
        \includegraphics[width=\linewidth]{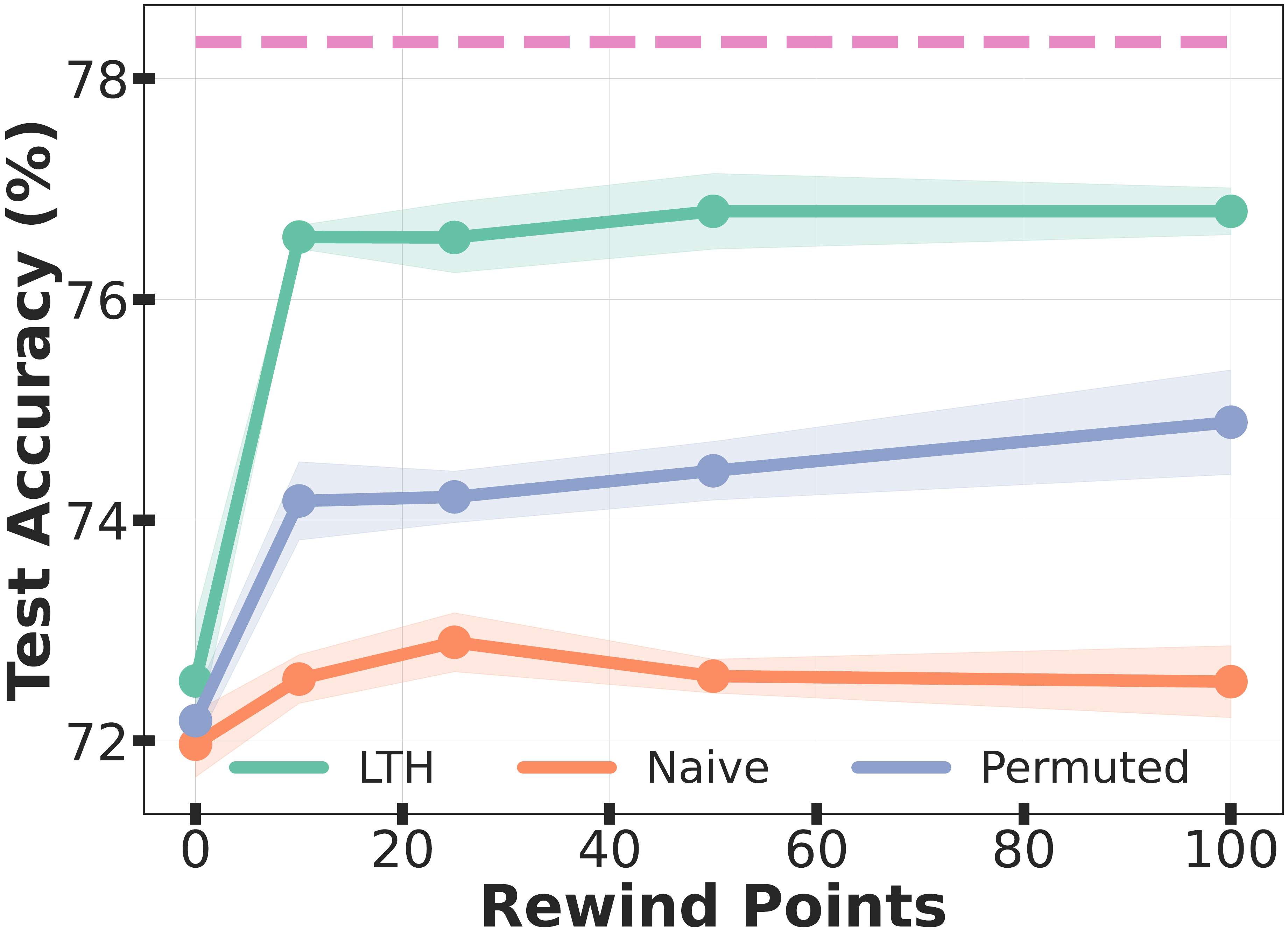}  % Replace with your image path
        % \caption{sparsity = 0.90}
        % \label{fig:resnet_w4_sp_fig:2}
    \end{subfigure}
    \begin{subfigure}{0.235\textwidth}
        \centering
        \includegraphics[width=\linewidth]{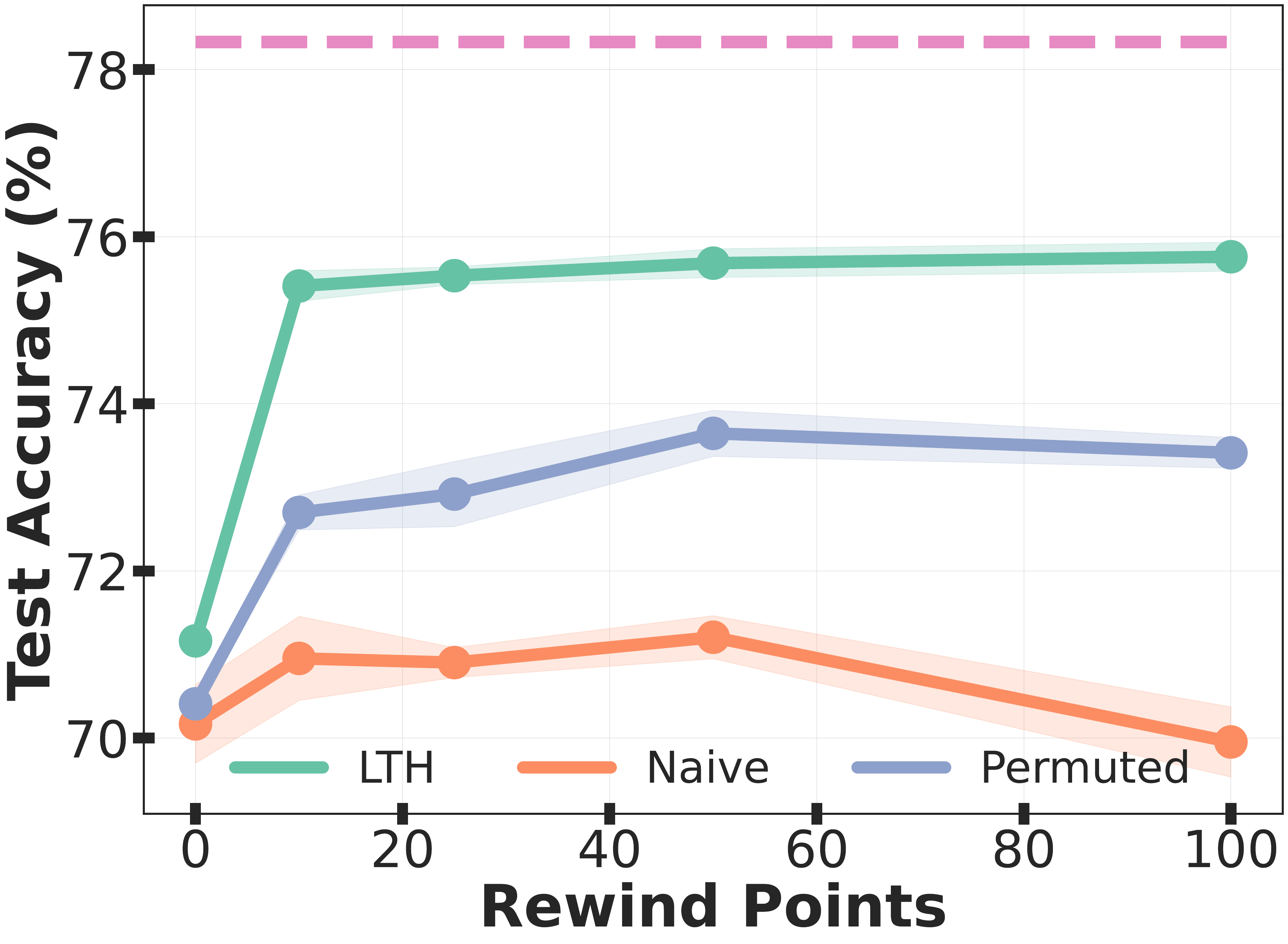}  % Replace with your image path
        % \caption{sparsity = 0.95}
        % \label{fig:resnet_w4_sp_fig:3}
    \end{subfigure}
    % Second row, second figure
    \begin{subfigure}{0.235\textwidth}
        \centering
        \includegraphics[width=\linewidth]{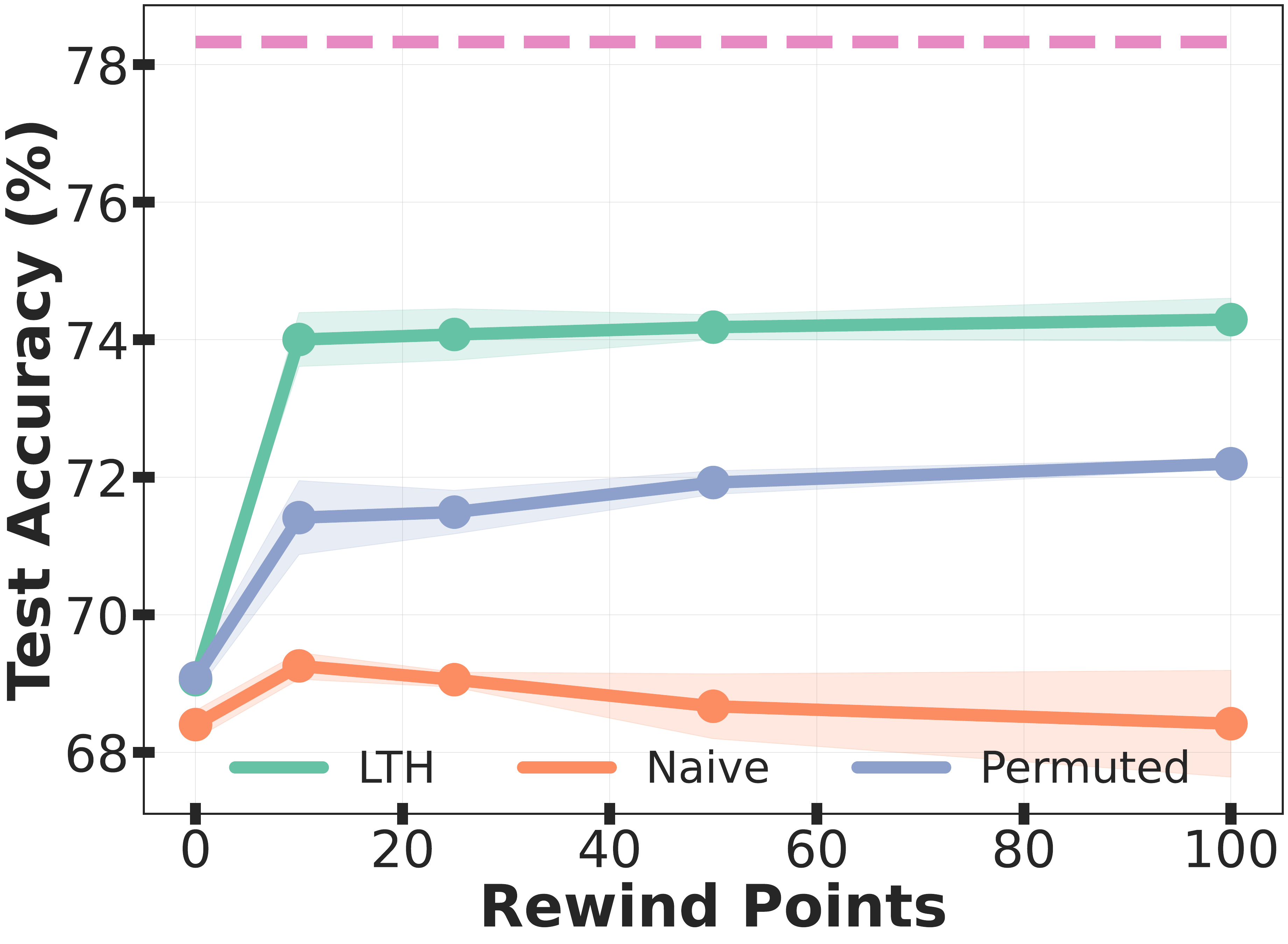}  % Replace with your image path
        % \caption{sparsity = 0.97}
        % \label{fig:resnet_w4_sp_fig:4}
    \end{subfigure}\\
    \begin{subfigure}{1.5em}
        \makebox[20pt]{\raisebox{50pt}{\rotatebox[origin=c]{90}{$w=8$}}}%
    \end{subfigure}
    \begin{subfigure}{0.235\textwidth}
        \centering
        \includegraphics[width=\linewidth]{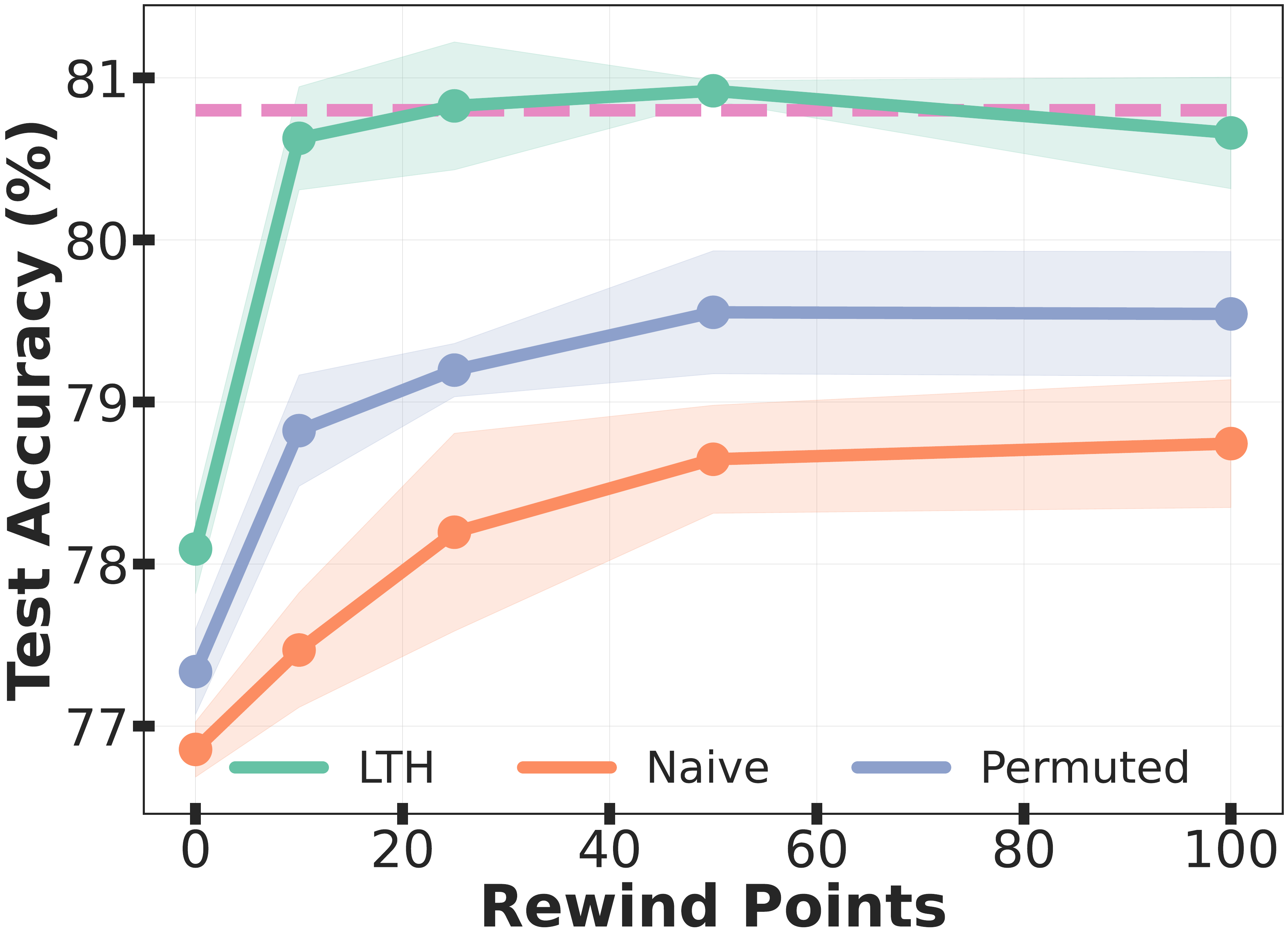}
        % \caption{sparsity = 0.80}
        % \label{fig:resnet_c100_w8_s80:1}
    \end{subfigure}
    \begin{subfigure}{0.235\textwidth}
        \centering
        \includegraphics[width=\linewidth]{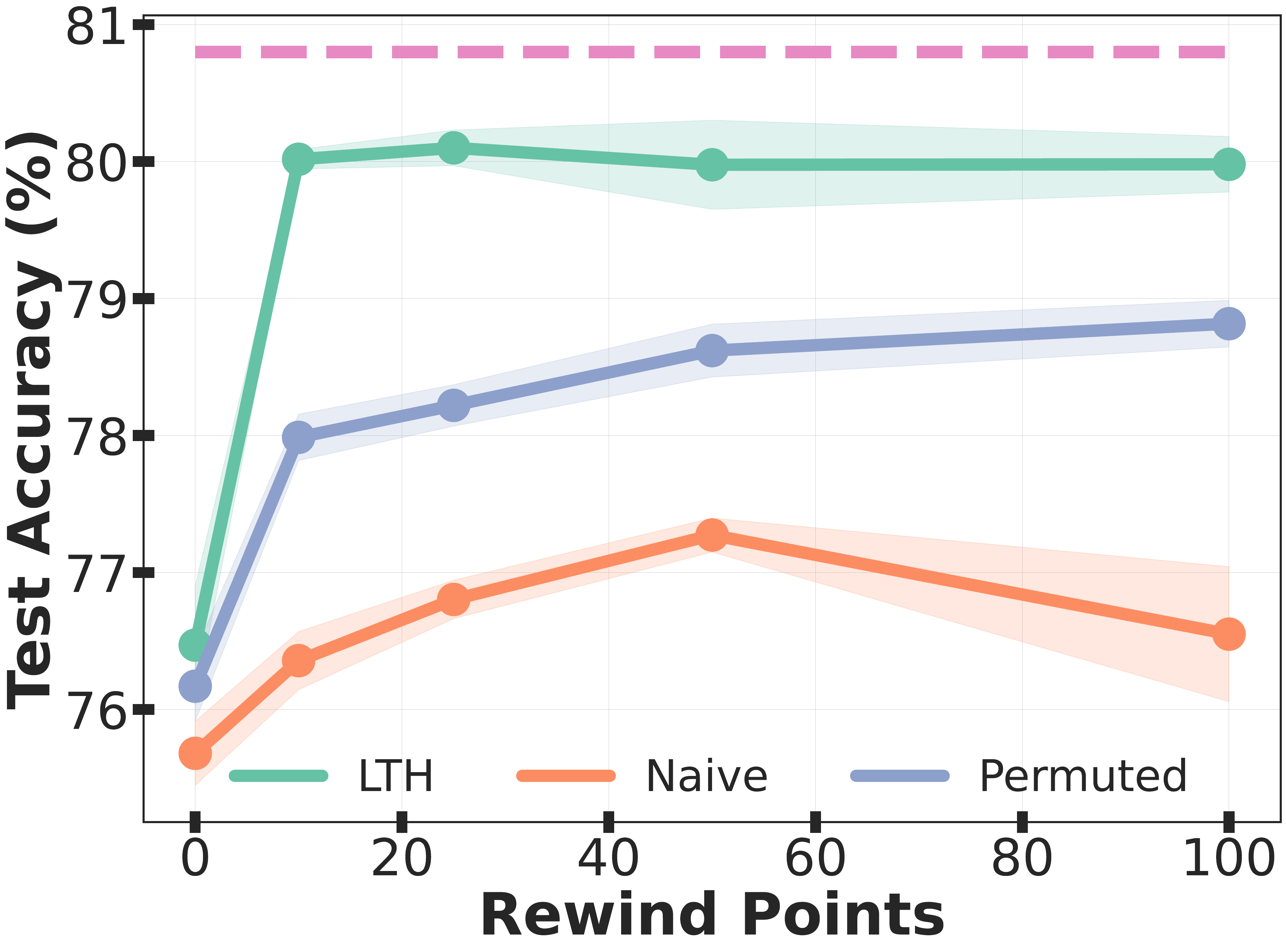}
        % \caption{sparsity = 0.90}
        % \label{fig:resnet_c100_w8_s90:2}
    \end{subfigure}
    \begin{subfigure}{0.235\textwidth}
        \centering
        \includegraphics[width=\linewidth]{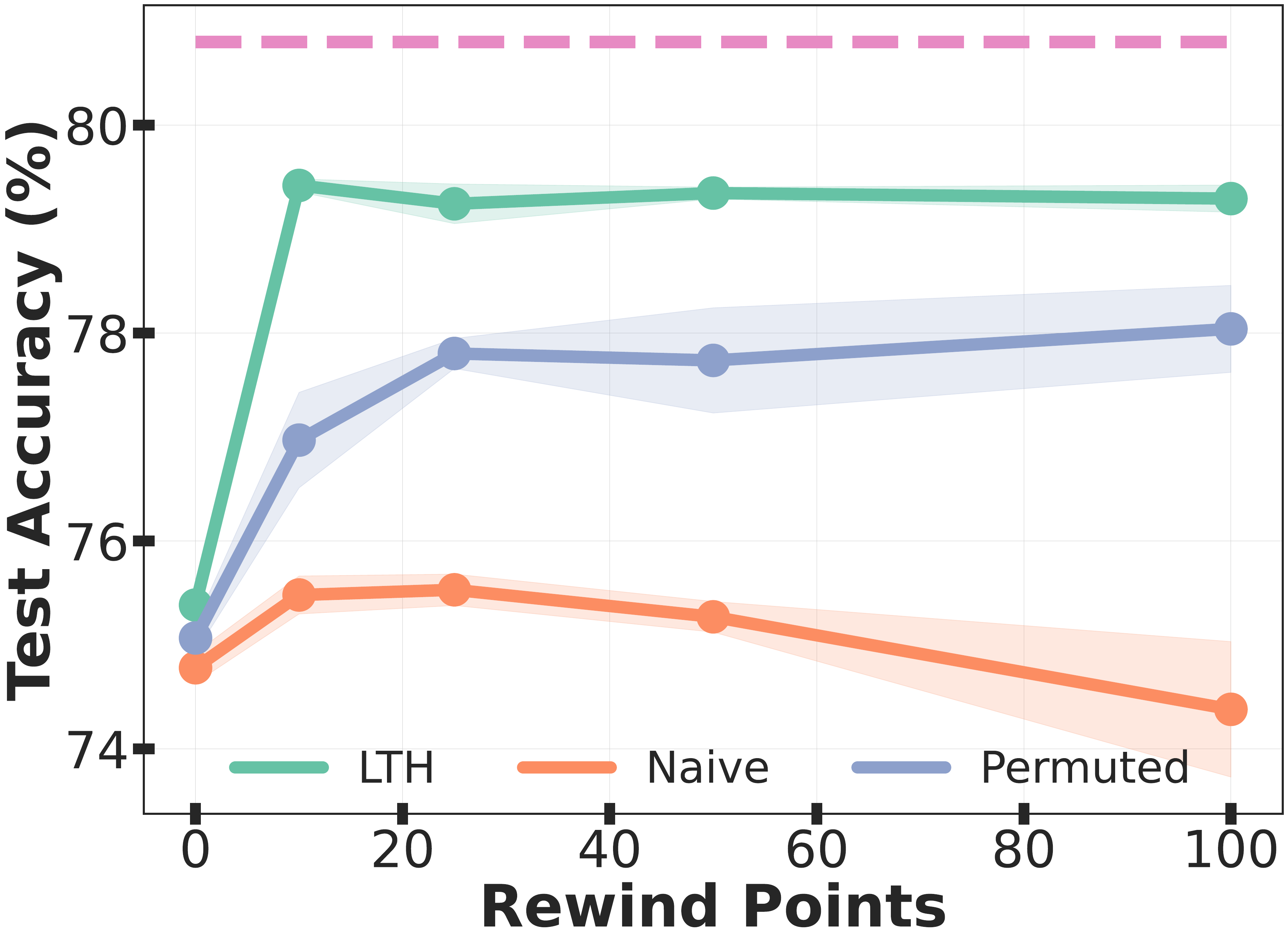}
        % \caption{sparsity = 0.95}
        % \label{fig:resnet_c100_w8_s95:3}
    \end{subfigure}
    \begin{subfigure}{0.235\textwidth}
        \centering
        \includegraphics[width=\linewidth]{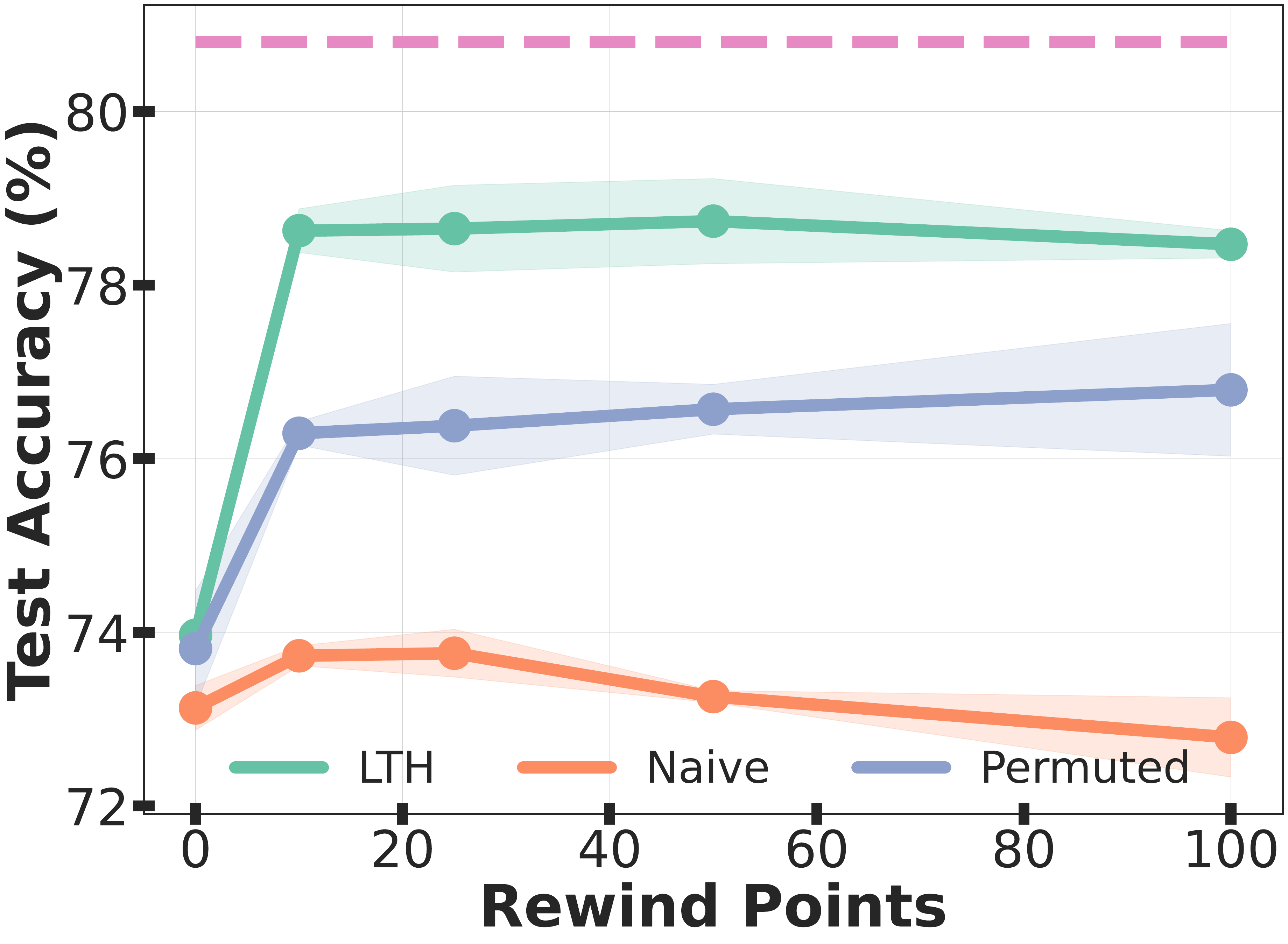}
        % \caption{sparsity = 0.97}
        % \label{fig:resnet_c100_w8_s97:4}
    \end{subfigure}\\
    \begin{subfigure}{1.5em}
        \makebox[20pt]{\raisebox{50pt}{\rotatebox[origin=c]{90}{$w=16$}}}%
    \end{subfigure}
    \begin{subfigure}{0.235\textwidth}
        \centering
        \includegraphics[width=\linewidth]{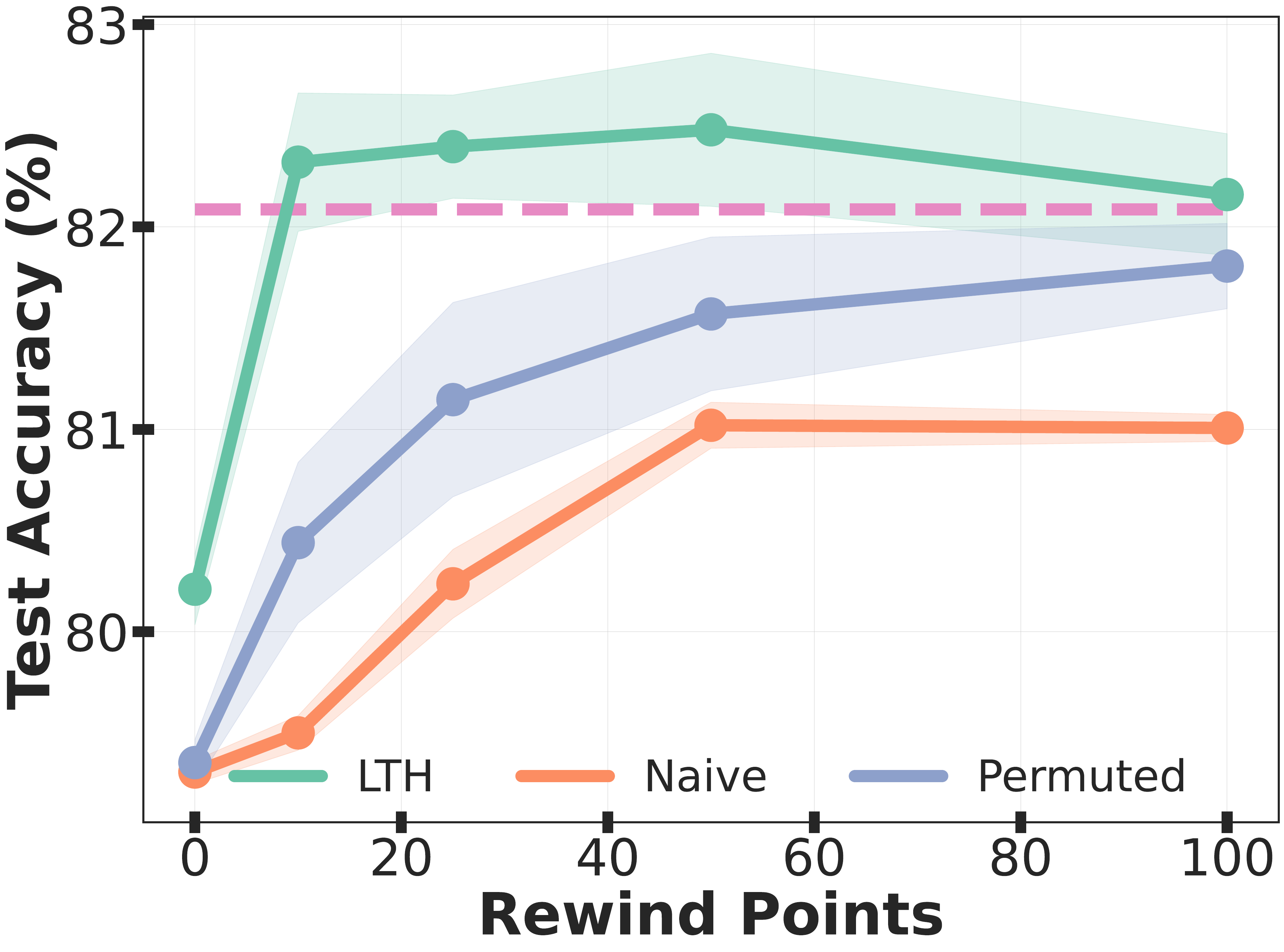}  % Replace with your image path
        \caption{sparsity = 0.80}
        % \label{fig:c100_w16_80:1}
        \label{fig:c100_80}
    \end{subfigure}
    \begin{subfigure}{0.235\textwidth}
        \centering
        \includegraphics[width=\linewidth]{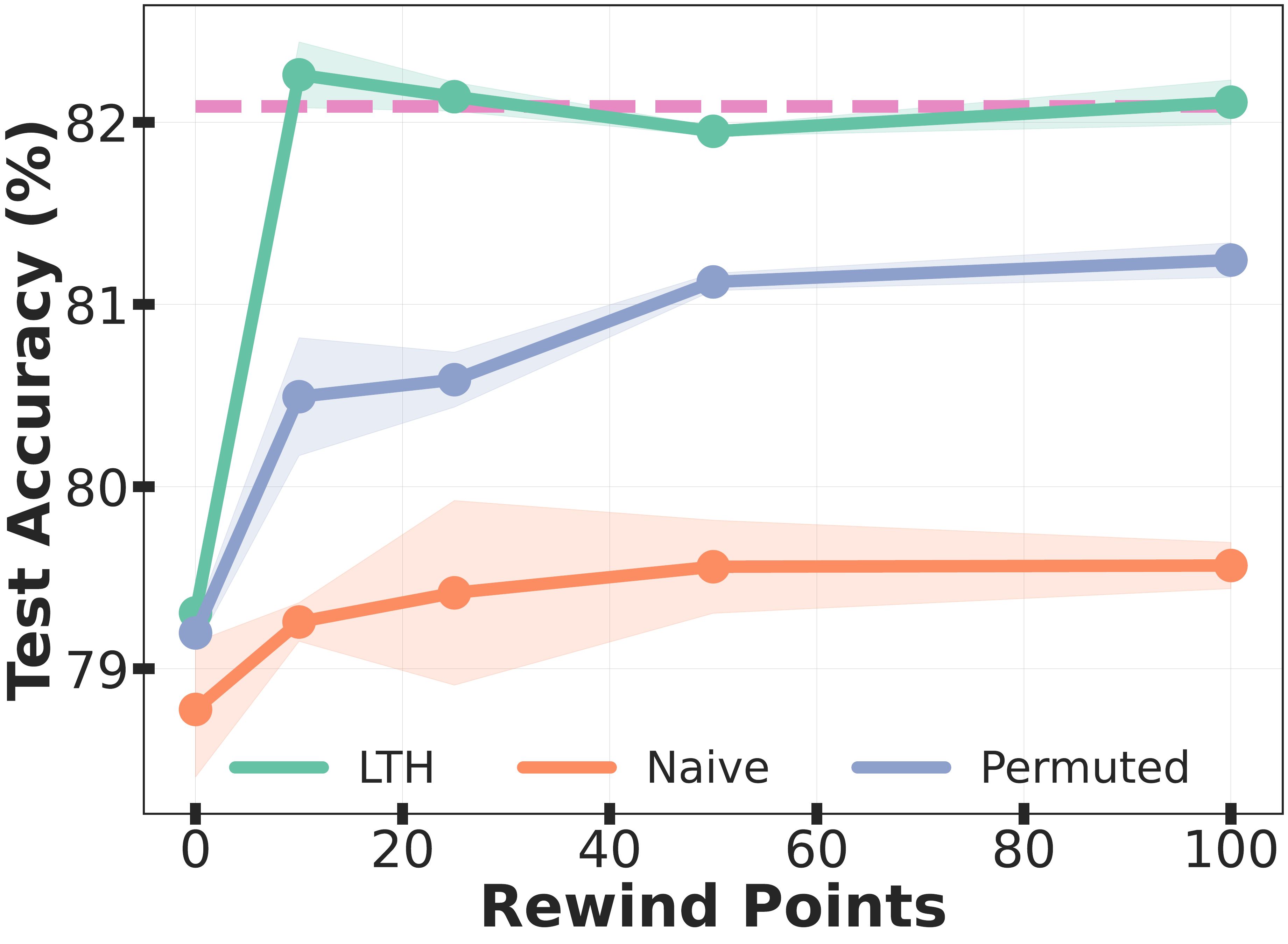}  % Replace with your image path
        \caption{sparsity = 0.90}
        % \label{fig:c100_w16_90:2}
        \label{fig:c100_90}
    \end{subfigure}
    \begin{subfigure}{0.235\textwidth}
        \centering
        \includegraphics[width=\linewidth]{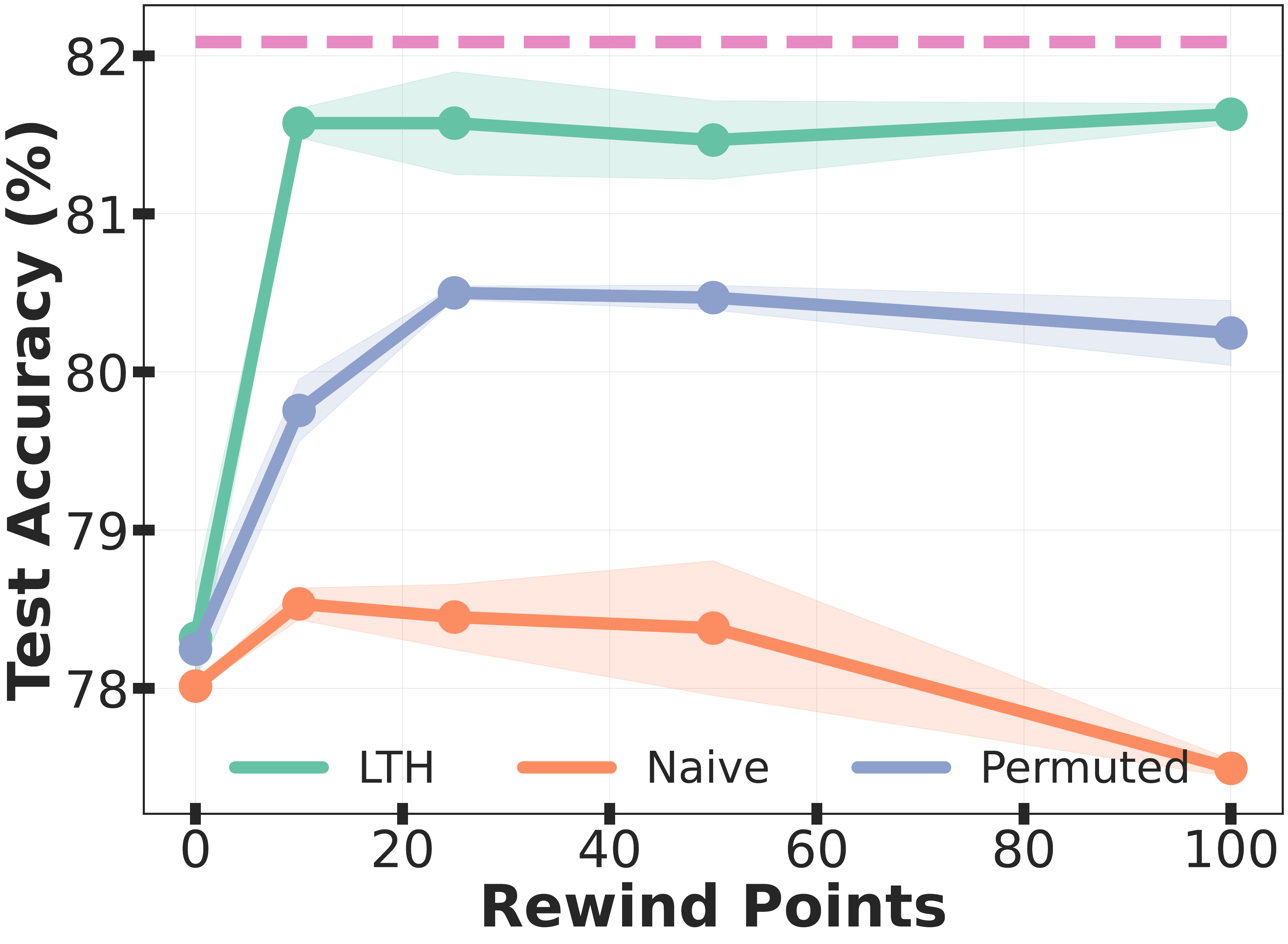}  % Replace with your image path
        \caption{sparsity = 0.95}
        % \label{fig:c100_w16_95:3}
        \label{fig:c100_95}
    \end{subfigure}
    \begin{subfigure}{0.235\textwidth}
        \centering
        \includegraphics[width=\linewidth]{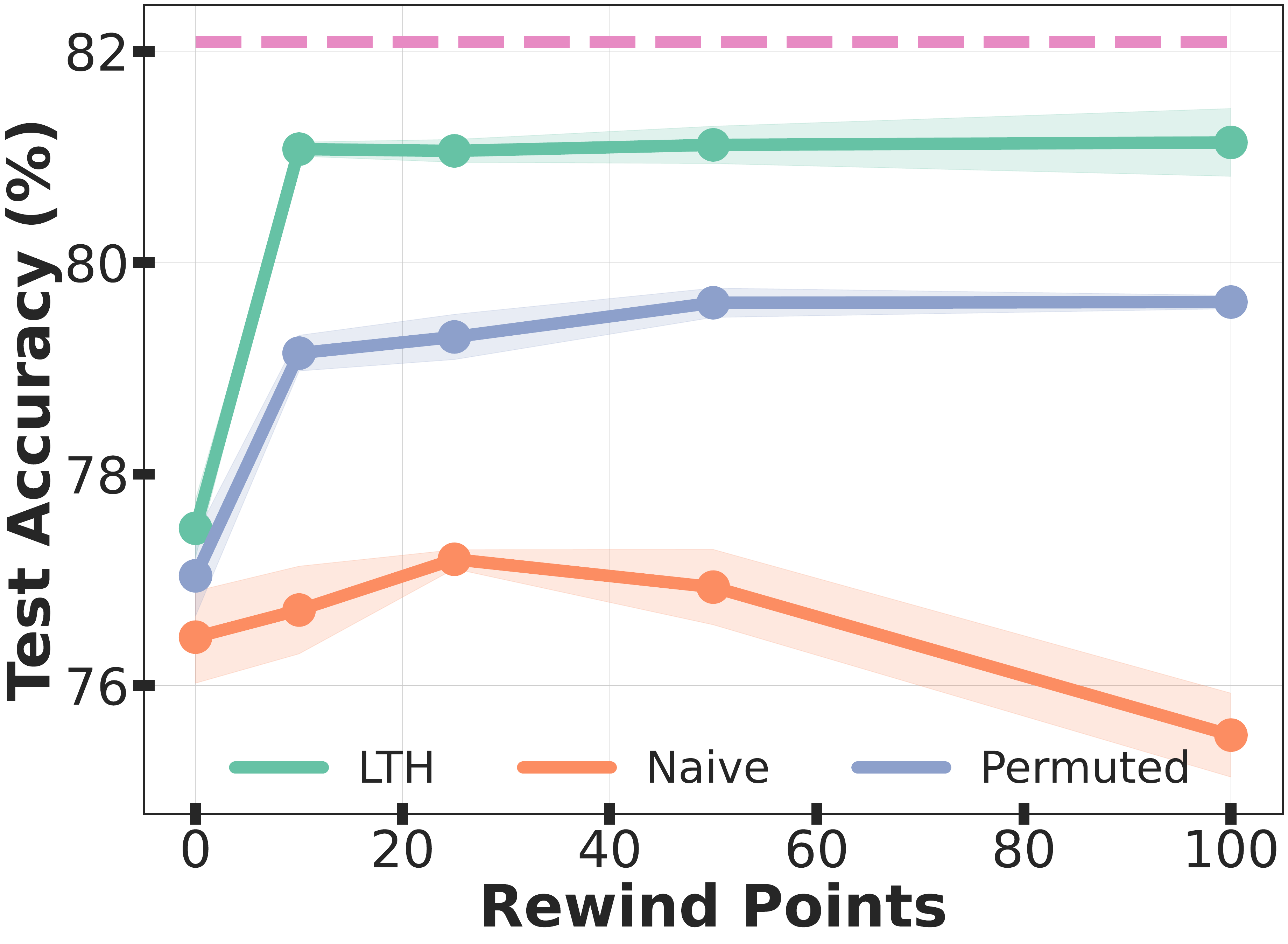}  % Replace with your image path
        \caption{sparsity = 0.97}
        % \label{fig:c100_w16_97:4}
        \label{fig:c100_97}
    \end{subfigure}
    \caption{{\textbf{ResNet20$\times\{w\}$/CIFAR-100}. Test accuracy of sparse network solutions vs.\ increasing rewind points for different sparsity levels and widths, $w$. The dashed ({\textbf{- -}}) line shows the dense model accuracy. The effect of the rewind points on the test accuracy for different sparsities is shown. As the width increases, the gap between training from a random initialization with the permuted mask and the LTH/dense baseline (dashed line) decreases, unlike training with the non-permuted mask (naive), showing model trained with the permuted model generalizes better than naive.}}
    \label{fig:cifar100_w4_rewind}
\end{figure*}

\subsection{Experimental Results.}
% \subsection{CIFAR-10/100}
\paragraph{ResNet20/CIFAR-10 \& CIFAR-100.}
We trained ResNet20 on the CIFAR-10/100 datasets. As shown in~\cref{fig:cifar10_allwidth_rewind,fig:cifar100_w4_rewind}, the permuted solution outperforms the naive baseline across all model widths and rewind points. Since it is more difficult to train models with higher sparsity, the gap between naive and permuted solutions increases as sparsity increases, as shown in~\cref{fig:resnet_c10_s97} for width multiplier 1,4,8, and 16. It can also be observed that at higher sparsity increasing the rewind point improves both the \gls{lth} and permuted solution but not the naive solution. The improved performance of the permuted solution over naive supports our hypothesis and shows that misalignment of the LTH mask and loss basin corresponding to the new random initialization could explain why LTH masks do not transfer to different initializations. We also show accuracy vs.\ sparsity plots for $k=\{10,25,50,100\}$ (details in~\cref{subsec:add_plots}); as sparsity increases, the gap between permuted and naive solution increases for all rewind points. As illustrated in figure~\cref{fig:cifar10_allwidth_rewind}, neither the \gls{lth} nor the permuted solution performs effectively with random initialization ($k=0$) but improves on increasing the rewind point up to a certain point, beyond which it plateaus. Detailed results are presented in~\cref{table:resnet20_1_CIFAR10,table:resnet20_4_CIFAR10,table:resnet20_8_CIFAR10,table:resnet20_16_CIFAR10} in~\cref{subsec:tables}.
% 
% consistent across varying width and different datasets both LTH and permuted solution improve as the rewind point increases in contrast to the naive solution, which does not improve on increasing the rewind point. We observed that naive performance saturates after $k \geq 50$ and does not yield further improvement. Since it is more difficult to train models with higher sparsity, the gap between naive and permuted solutions increases as sparsity increases, as shown in
% ~\cref{fig:resnet_c10_w1_s97:4,fig:resnet_c10_w4_s97:4,fig:resnet_c10_w8_s97:4}. The improved performance of the permuted solution over naive supports our hypothesis and shows that misalignment of LTH masks and loss basin corresponding to new random initialization could explain why LTH masks do not transfer to different initializations. We also show accuracy vs sparsity plots for $k=\{10,25,50,100\}$ (details in~\cref{subsec:add_plots}); as sparsity increases, the gap between permuted and naive solution increases for all rewind points. 

We also validated our hypothesis on CIFAR-100 using ResNet20 with varying widths. As shown in~\cref{fig:cifar100_w4_rewind}, the permuted solution consistently outperforms the naive solution, showing that our hypothesis holds true across different models and datasets. Similar to the CIFAR-10 dataset, as we increase the model width multiplier, the gap between the permuted and naive solution increases, showing the efficacy of our method. Detailed results are presented in~\cref{table:resnet20_1_CIFAR100,table:resnet20_4_CIFAR100,table:resnet20_8_CIFAR100,table:resnet20_16_CIFAR100} in~\cref{subsec:tables}.
% and the permuted solution gradually approaches the LTH baseline.

\begin{figure}[tbp]
    \centering
    \begin{subfigure}[b]{0.49\linewidth}
        \centering
        \includegraphics[width=\textwidth]{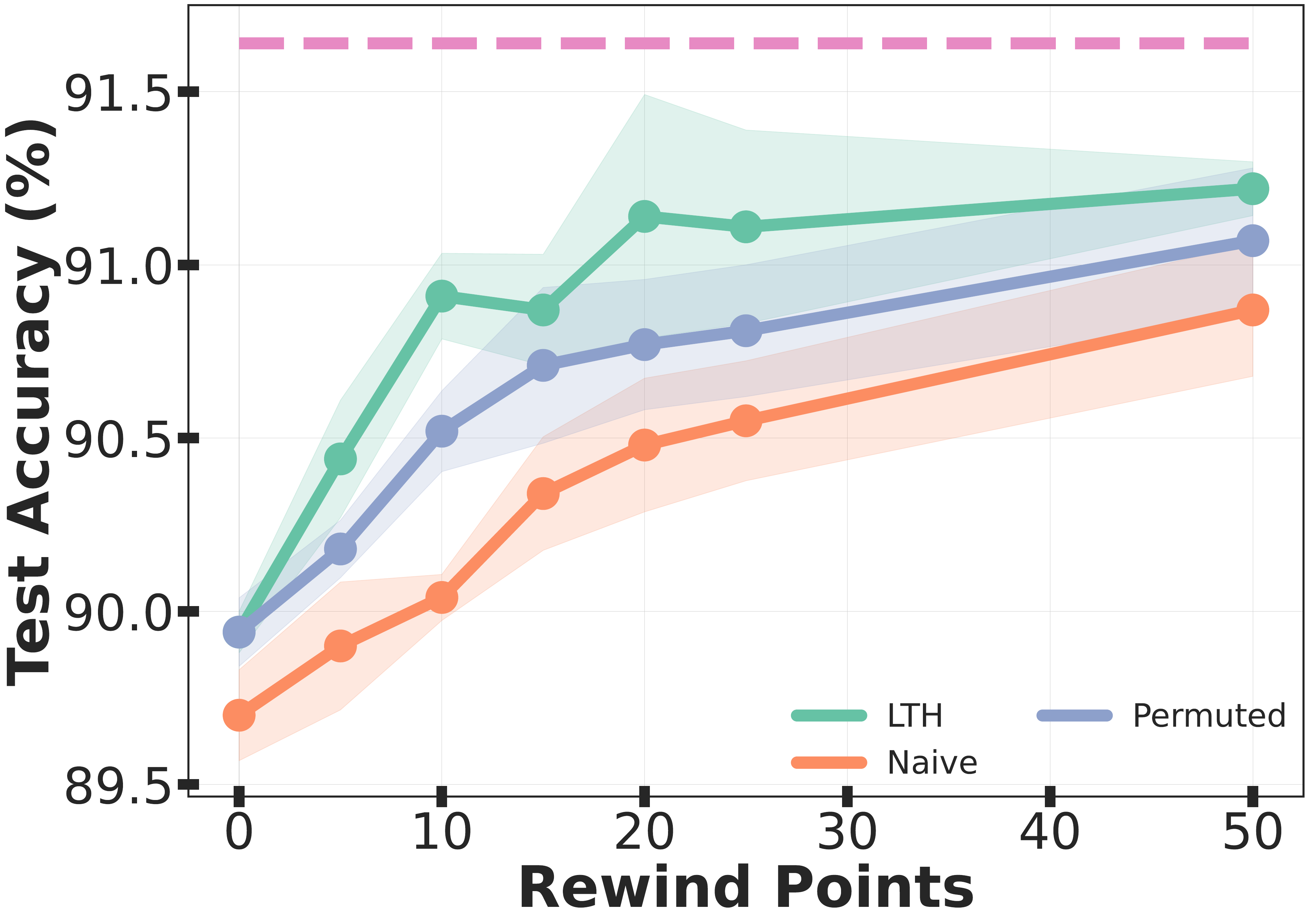}
        \caption{sparsity = 0.80}
        \label{fig:vgg11-sparsity-08}
    \end{subfigure}
    \begin{subfigure}[b]{0.49\linewidth}
        \centering
        \includegraphics[width=\textwidth]{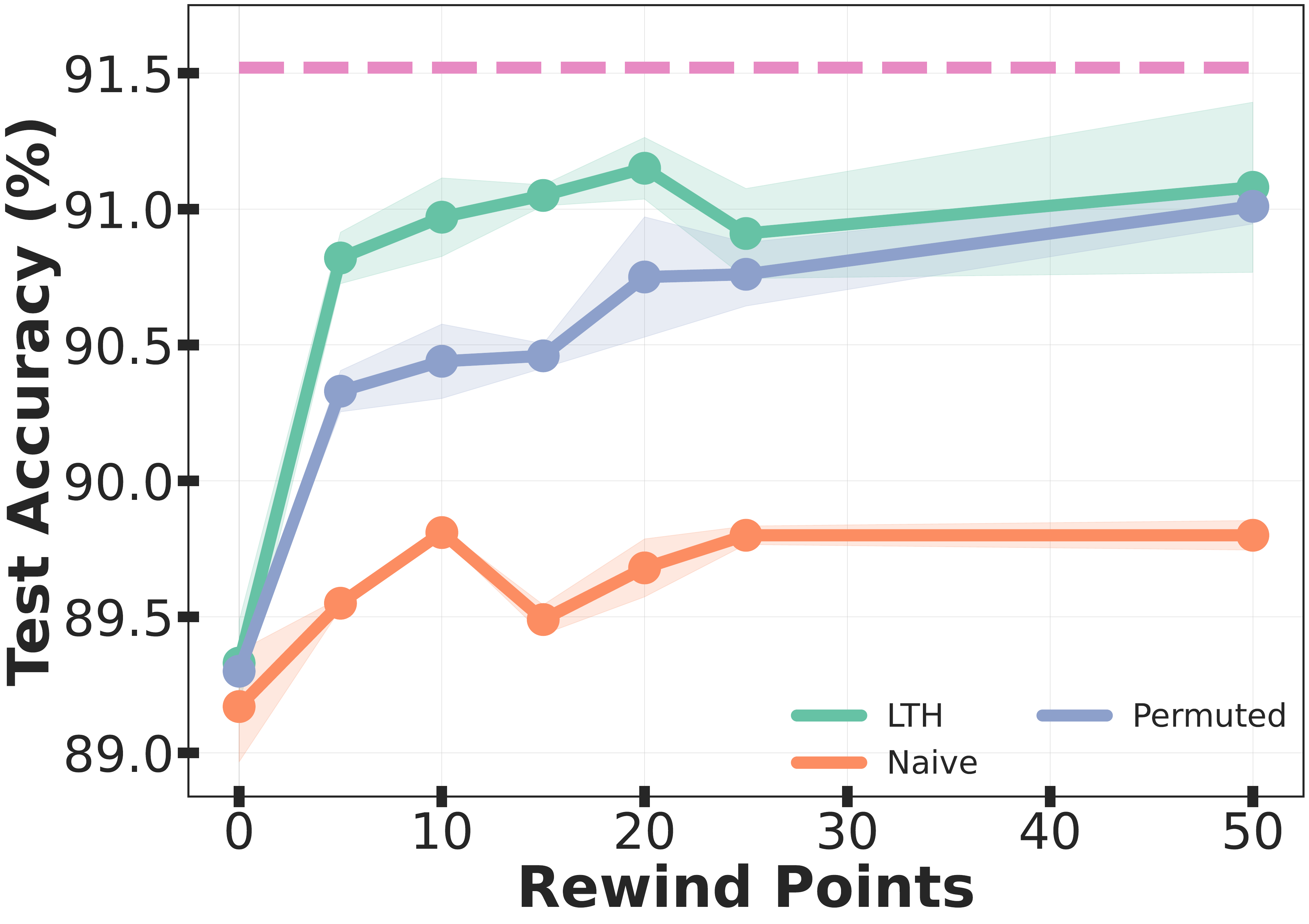}
         \caption{sparsity = 0.90}
        \label{fig:vgg11-sparsity-090}
    \end{subfigure}
    \caption{\textbf{VGG11$\times \{1\}$/CIFAR-10.} Test accuracy of sparse solutions at increasing rewind points for different sparsity levels.
    The dashed ({\textbf{- -}) line shows the dense model accuracy}. In ~\cref{fig:vgg11-sparsity-090}, the permuted solution closely matches the \gls{lth} solution. However, beyond a certain rewind point, i.e.\ for $k \geq 20$ the performance of the naive solution plateaus. Resulting in a more noticeable gap between the permuted and naive solutions compared to ~\cref{fig:vgg11-sparsity-08}.
    % \textbf{Note}, we do not include the plots for sparsity = \{0.95, 0.97\}, because the naive solution performs poorly yielding consistent metrics of test accuracy $= 10\%$ and test loss $ \approx \log_e(10)$ throughout all rewind points.
    }
    \label{fig:vgg-sparsity-comparison}
\end{figure}

\paragraph{VGG11/CIFAR-10.} We utilize the modified VGG11 architecture implemented by \citet{jordan2023repair} trained on CIFAR-10 (details in \cref{implementation}). We observe that for a moderate sparsity ($80\%$) in~\cref{fig:vgg11-sparsity-08}, the gap between the permuted and the naive baseline is not large, however for a higher sparsity level~($90\%$), the permuted solution significantly outperforms the naive solution as shown in~\cref{fig:vgg11-sparsity-090}. For the VGG11 model, on increasing the rewind point, the permuted solution closely matches the accuracy of \gls{lth}, while the naive solution significantly plateaus and does not improve on increasing the rewind point. For higher sparsities, the naive baseline was unstable in training as the modified VGG11 architecture does not have BatchNorm layers~\citep{batchnorm2015}; we omit those results in the discussion for a fair comparison. Detailed results are presented in~\cref{table:vgg11_CIFAR10} in~\cref{subsec:tables}.

% the permuted and the naive solution are relatively similar and steadily increasing together as we increase the rewind points with the permuted solution consistently taking the slight edge over naive. 
% As sparsity increases in ~\cref{fig:vgg11-sparsity-090}, a significant gap begins to emerge between the permuted and naive solutions. %indicating that the permutation plays a stronger role in aligning with the desired optimization basin. 
% As the rewind point increases, the permuted solution gradually improves and approaches the performance of LTH, while the naive solution significantly plateaus for $k \geq 20$ and performance subsides.
%
%

%
% We further provide evidence in \cref
%
%
% \newpage
%

\begin{figure}[tbp]
    \centering
    \begin{subfigure}{0.9\linewidth}
        \centering
        \includegraphics[width=\linewidth]{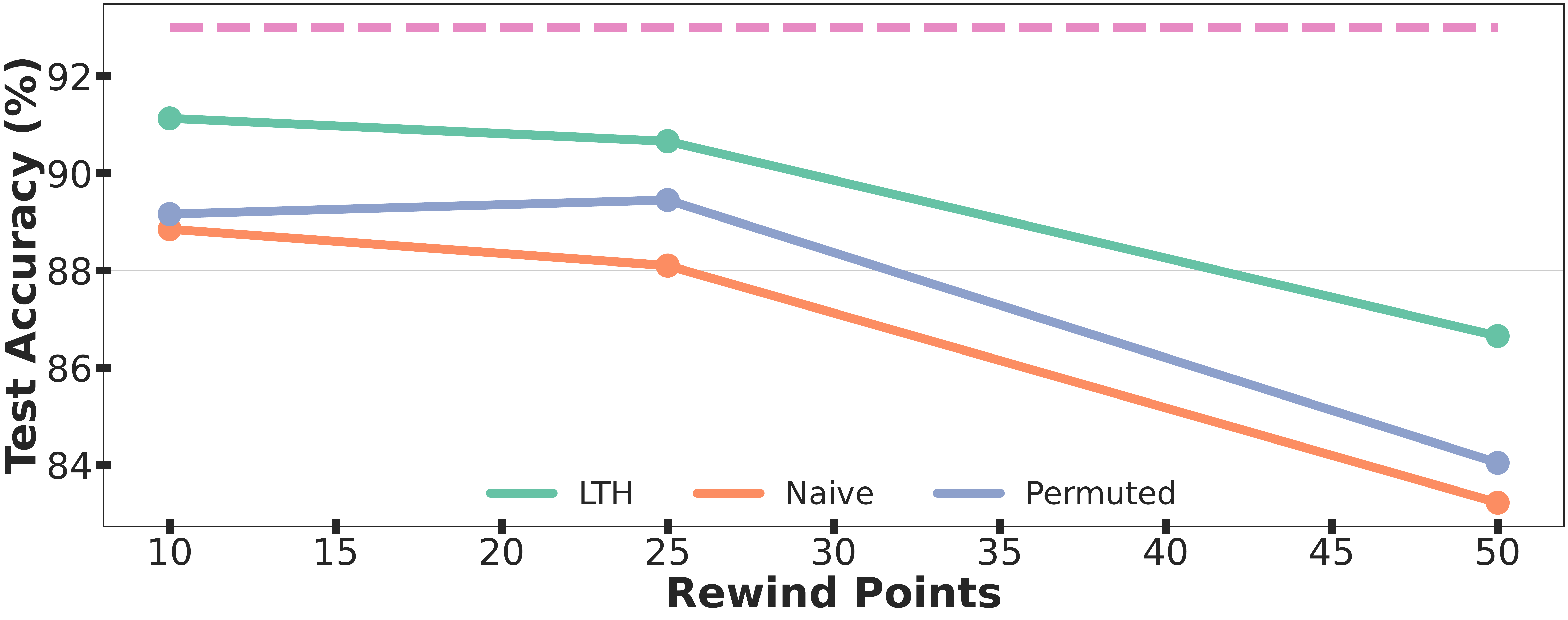}
        \caption{sparsity = 0.80}
        \label{fig:imagenet_80_plot_top5}
    \end{subfigure}
    \begin{subfigure}{0.9\linewidth}
        \centering
        \includegraphics[width=\linewidth]{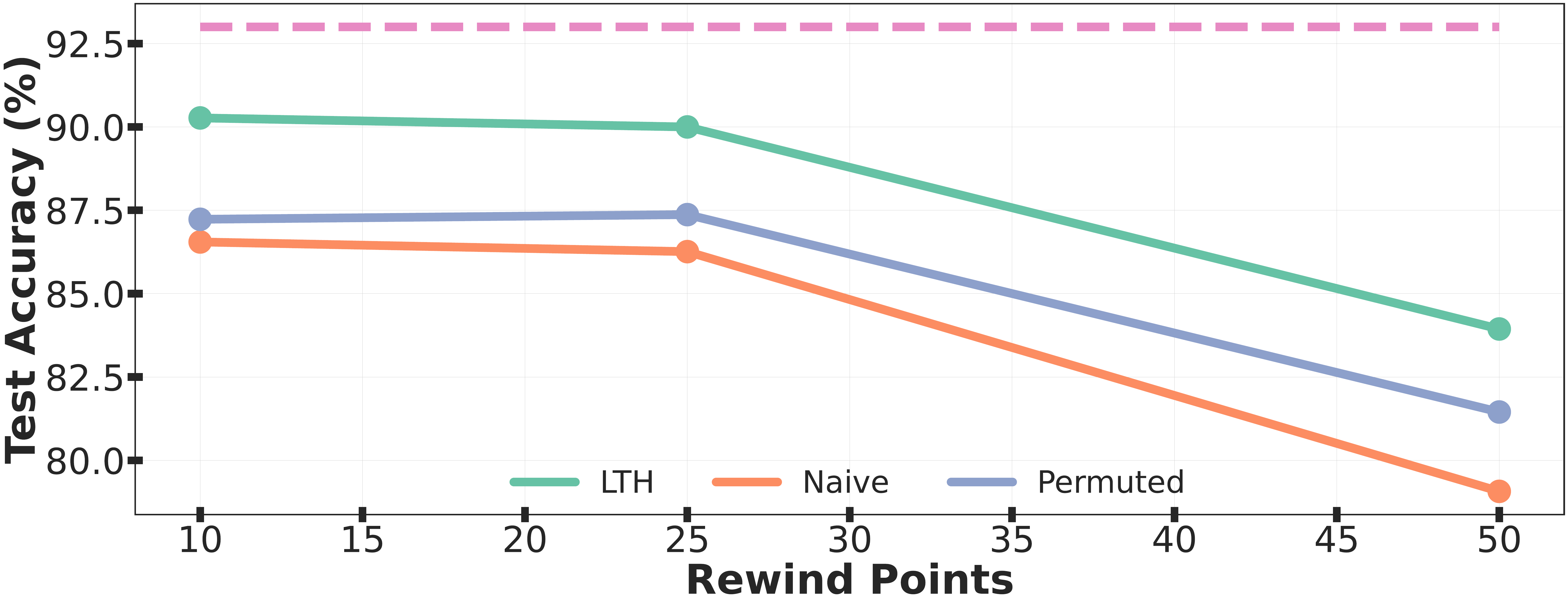}
        \caption{sparsity = 0.90}
        \label{fig:imagenet_90_plot_top5}
    \end{subfigure}
    \begin{subfigure}{0.9\linewidth}
        \centering
        \includegraphics[width=\linewidth]{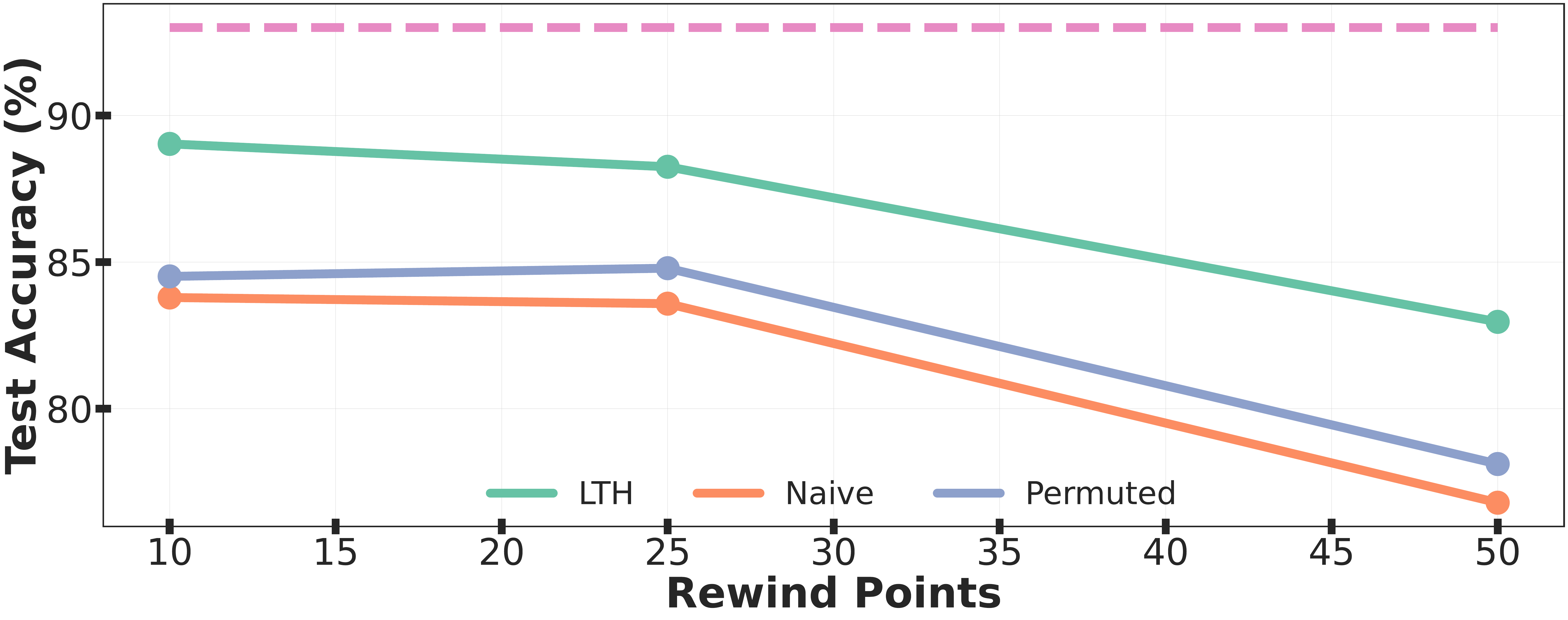}
        \caption{sparsity = 0.95}
        \label{fig:imagenet_95_plot_top5}
    \end{subfigure}
    \caption{{\textbf{ResNet50$\times\{1\}$/ImageNet}. Top-5 test accuracy vs.\ rewinds points of sparse network solutions at various sparsity levels. We observe the permuted solution consistently performing better than the naive solution for all sparsities. The dashed ({\textbf{- -}}) line shows the dense model accuracy. }}
    \label{fig:imagenet_plots_top5}
\end{figure}
% \vspace{-1cm}
% \subsection{ImageNet}
\paragraph{ResNet50/ImageNet.}
We also validated our hypothesis on the ILSVRC 2012 (ImageNet) dataset, which consists of 1.28 million images across 1,000 classes \citep{imagenet}. We used the ResNet50 model to evaluate the performance of the permuted mask at different sparsity levels. As observed in~\cref{fig:imagenet_plots_top5}, the permuted solution outperforms the naive solution across all sparsity levels, showing that our hypothesis holds true on large-scale datasets as well. While the permuted solution performs better than the naive solution, there is still a significant gap between \gls{lth} and the permuted solution in the case of the ImageNet dataset as compared to the CIFAR-10/100 dataset.
This could be due to permutation matching not being accurate enough, as only a small subset of the training dataset was used for activation matching. This can also be visualized in terms of the loss barrier in~\cref{fig:interp_resnet50_imagenet} between the permuted model $A$ and model $B$; the loss barrier after permutation is more prominent compared to the CIFAR dataset~(\cref{fig:interp_resnet20_c10,fig:interp_resnet20_c100}). Thus, the permutation mapping $\pi$ cannot match the models perfectly in the case of ImageNet since the permutation matching algorithm uses a greedy search algorithm to find the permutation mapping. However, given a better mapping, it may be possible to further improve the performance of the permuted solution as discussed in~\cref{subsec:effect}. Detailed results are presented in~\cref{table:resnet50_imagenet} in~\cref{subsec:tables}. As demonstrated in~\cref{table:resnet50_imagenet}, the permuted solution outperforms the naive approach by nearly $2\%$ at higher sparsity levels.

\subsection{Diversity Analysis of Permuted Models.}
\label{ensemble}
A limitation of \gls{lth} is that it consistently converges to very similar  solutions to the original pruned model~\citep{evci2022gradient}. \citet{evci2022gradient} speculate this occurs because the \gls{lth} is always trained with the same initialization/rewind point, and effectively relearns the same solution. 
Our hypothesis is that permuted \gls{lth} masks, trained with distinct initialization/rewind points and subject to approximation errors in permutation matching, may learn more diverse functions than the \gls{lth} itself.
We analyze the diversity of sparse models trained at $90\%$ sparsity, with either a permuted mask (permuted), the \gls{lth} mask (naive), \gls{lth} mask \& init.\ and the original pruned solution (IMP) on which the \gls{lth} is based. We follow the same analysis as~\citet{evci2022gradient} and compare the diversity of the resulting models, over five different training runs, using disagreement score, KL divergence and JS divergence. We also compare with an ensemble of five models trained independently with different random seeds.
%We also train an ensemble for each set of models, over 5 independent runs. 
As shown in~\cref{tab:diversity}, an ensemble of permuted models shows higher diversity across all the metrics than the \gls{lth}, showing that the permuted models learn a more diverse set of solutions. We provide additional details in~\cref{ensemble_details}.

\begin{table}[tbp]
    \centering
    \small
    \caption{\textbf{Ensemble Diversity Metrics for CIFAR-10/CIFAR-100}. Although the mean test accuracy of \gls{lth} is higher, the ensemble of permuted models achieves better test accuracy due to better functional diversity of permuted models. Here we compare several measurements of function space similarity between the models including disagreement, which measures prediction differences ~\citep{fort2020deepensembleslosslandscape}, and Kullback–Leibler (KL)/Jenson-Shannon (JS) divergence, which quantify how much the output distributions of different models differ ~\citep{evci2022gradient}. As shown, the permuted masks achieve similar diversity as computational expensive IMP solutions, also resulting in ensembles with a similar increase in generalization.}\label{tab:diversity}
    \begin{tabular}{@{}p{4.2em}p{5.5em}p{3em}p{3em}p{2em}p{2em}@{}}
        \toprule
        Mask & Test Accuracy (\%) & Ensemble Acc.~(\%) & Disagree-ment & KL & JS \\
        \cmidrule{1-6}
        \multicolumn{6}{c}{ResNet20$\times\{1\}$/CIFAR\nobreakdashes-10}\\
        \cmidrule{1-6}
        none\,(dense) & 92.76 $\pm$ 0.106 & \multicolumn{1}{c}{-} & \multicolumn{1}{c}{-} & \multicolumn{1}{c}{-} & \multicolumn{1}{c}{-}\\
        IMP & 91.09 $\pm$ 0.041 & 93.25 & {0.093} & {0.352}& {0.130}\\
        \cmidrule(l){2-6}
        LTH & \textbf{91.15 $\pm$ 0.163} & 91.43 & 0.035 & 0.038 & 0.011\\
        permuted & 89.38 $\pm$ 0.170 & \textbf{91.75} & {0.107} & \textbf{0.273}& \textbf{0.091} \\
        naive & 88.68 $\pm$ 0.205 & {91.07} & \textbf{0.113} & {0.271}& {0.089}\\
        \cmidrule{1-6}
        \multicolumn{6}{c}{ResNet20$\times\{4\}$/CIFAR\nobreakdashes-100}\\
        \cmidrule{1-6}
        none\,(dense) & {78.37 $\pm$ 0.059} & \multicolumn{1}{c}{-} & \multicolumn{1}{c}{-} & \multicolumn{1}{c}{-} & \multicolumn{1}{c}{-}\\
        IMP & 74.46 $\pm$ 0.321 & 79.27 & {0.259} & {1.005}& {0.372}\\
        \cmidrule(l){2-6}
        LTH & \textbf{75.35 $\pm$ 0.204} & 75.99 & 0.117 & 0.134 & 0.038\\
        permuted & 72.48 $\pm$ 0.356 & \textbf{77.85} & {0.278} & {0.918}& {0.327} \\
        naive & 71.05 $\pm$ 0.366 & {76.15} & \textbf{0.290} & \textbf{0.970}& \textbf{0.348}\\
        \bottomrule
    \end{tabular}
\end{table}
\subsection{Effect of Model Width Multiplier.}
\label{subsec:effect}
Permutation matching is an NP-hard problem; the activation matching algorithm proposed by \citet{ainsworth2023git} does not find the global optimum; rather, it uses a greedy search to explore a restricted solution space. Therefore, in practice, permutation matching does not perfectly align two models. However, it has been observed that for wider models, the algorithm can more closely align two models~\citep{ainsworth2023git,sharma2024simultaneous}.
To understand how the performance of the permuted model is affected by the approximation error of the matching algorithm, we evaluated the \gls{lmc} and the accuracy of the permuted solution on ResNet20 models with varying layer widths. As shown in~\cref{fig:interp_plots}, on increasing the layer width, the loss barrier of the interpolated network reduces, showing that permutation mapping becomes more accurate and aligns two models better.
Also, it can be observed in~\cref{fig:cifar10_allwidth_rewind,fig:cifar100_w4_rewind} that the permuted solution becomes close to the LTH solution on increasing the model width, showing that as the permutation matching becomes more accurate, the gap between the \gls{lth} and the permuted solution reduces.

% \newpage

\section{Conclusion}
\glsresetall
% Sparse training and \gls{lth} have generated a lot of traction over the past years. In this work, we further try to gain insights on sparse training from random initialization, and the \gls{lth}, leveraging permutation invariance in DNNs. We empirically observed evidence across multiple models and datasets that appears to support our hypothesis that misalignment between the mask and loss basin makes it impossible to use LTH masks with a different initialization. One limitation of our work is that activation matching is not very accurate for narrow models --- in future work, we will explore more efficient matching algorithms.
% Sparse training and the \gls{lth} have gained significant traction in recent years. 
In this work, we demonstrate new insights into sparse training from random initialization and the \gls{lth} by leveraging weight symmetry in \glspl{dnn}.
Our empirical findings across various models and datasets support the hypothesis that misalignment between the mask and loss basin prevents effective use of \gls{lth} masks with new initialization.
 Although finding a permutation to align dense models is computationally expensive, the goal of our work is to develop insights into the working of \gls{lth} and how the sparse mask can be reused, not to improve the efficiency of \gls{lth}. We hope that our work will spur future work in this direction and will be useful to the research community working in the realm of sparse training.
% One limitation is that activation matching is weaker for narrow models; future work will explore more efficient matching algorithms.
%For the extended version of the manuscript, we plan to add results with the ImageNet dataset~\citep{imagenet}. 

\clearpage 
\section*{Impact Statement}
Our work focuses on improving sparse training and reducing the computational cost of training \gls{dnn}. By reducing the computational and memory requirements of training and inference, sparse training can facilitate the deployment of deep learning models on resource-constrained devices. However, most current hardware cannot leverage unstructured sparsity, which currently limits the impact of this work. Furthermore, model pruning often introduces algorithmic bias in the model~\citep{hooker2020characterisingbiascompressedmodels}, it is important to evaluate algorithmic bias of sparse models before deploying for real-world applications. 

\section*{Acknowledgements}
We would like to acknowledge the assistance of Nayan Saxena with an early code prototype for the project, and Yigit Yargic with an early iteration of \cref{fig:symmetry-illustration}. 

We gratefully acknowledge the support of Alberta Innovates (ALLRP-577350-22, ALLRP-222301502), the Natural Sciences and Engineering Research Council of Canada (NSERC) (RGPIN-2022-03120, DGECR-2022-00358), and Defence Research and Development Canada (DGDND-2022-03120). 

This research was enabled in part by support provided by the Digital Research Alliance of Canada (alliancecan.ca). Resources used in preparing this research were provided, in part, by the Province of Ontario, the Government of Canada through
CIFAR, and companies sponsoring the Vector Institute.

% Individual acknowledgements, keep in order of author order.
MA is supported by the NSERC Postgraduate Scholarship, RBC Borealis through the Borealis AI
Global Fellowship Award and the Digital Research Alliance of Canada EDIA Champions program.
ES is supported by the Vector Research Grant at the Vector Institute. 
RGK gratefully acknowledges support from a Canada CIFAR AI Chair.
YI is supported by a Schulich Research Chair.

\section*{Contribution Statement}
All authors contributed to the writing of the paper. MA and RJ implemented the method, conducted most experiments, and contributed the majority of the writing. MA and YI played  key roles in designing the methodology and experimental setup. ES contributed to the design of the experimental setup, provided key insights, hypothesized the connection to the conclusion drawn by~\citet{paul2022unmasking}, as summarized in~\cref{fig:loss_landscape} and our contributions, and ran experiments to validate it. RGK and YI, as senior authors, provided feedback on the writing and methodology. YI conceptualized the research idea, designed \cref{fig:symmetry-illustration,fig:workflow,fig:symmetry-illustration-full}, and also contributed feedback on code and experiments throughout the project. 

\bibliography{reference}
\bibliographystyle{icml2025}

%%%%%%%%%%%%%%%%%%%%%%%%%%%%%%%%%%%%%%%%%%%%%%%%%%%%%%%%%%%%%%%%%%%%%%%%%%%%%%%
%%%%%%%%%%%%%%%%%%%%%%%%%%%%%%%%%%%%%%%%%%%%%%%%%%%%%%%%%%%%%%%%%%%%%%%%%%%%%%%
% APPENDIX
%%%%%%%%%%%%%%%%%%%%%%%%%%%%%%%%%%%%%%%%%%%%%%%%%%%%%%%%%%%%%%%%%%%%%%%%%%%%%%%
%%%%%%%%%%%%%%%%%%%%%%%%%%%%%%%%%%%%%%%%%%%%%%%%%%%%%%%%%%%%%%%%%%%%%%%%%%%%%%%
\newpage
\appendix
\onecolumn
\section{Appendix}

\label{appendix}

\subsection{Implementation Details for ResNet20 \& VGG11 on CIFAR-10 and CIFAR-100}
\label{implementation}

\paragraph{Architectures} For residual neural networks, we train the standard ResNet20 on CIFAR-10 and CIFAR-100 with varying width. We implemented a scalar, $w$, that adjusts the number of channels in each convolutional and fully connected layer:
\begin{itemize}
\item \textbf{First Convolution Layer}: The number of output channels is scaled from $16$ to $w \times 16$.
    \item \textbf{Layer 1,2,3}: The number of output channels for the convolutional blocks in these layers are scaled from $16$, $32$, and $64$ to $w \times 16$, $w \times 32$, and  $w \times 64$, respectively.
    \item \textbf{Fully Connected Layer}: The input dimension to the final linear layer is scaled to $w \times 64$.
\end{itemize}
For convolutional neural networks, we train a modified version of the standard VGG11 implemented by \citet{jordan2023repair} on CIFAR-10. Primary differences are: 
\begin{itemize}
    \item A single fully connected layer at the end which directly maps the flattened feature map output from the convolutional layers to the 10 classes for CIFAR-10 classification.
    \item The classifier is set up for CIFAR-10 with 10 output classes as originally VGG11 was designed for ImageNet with 1000 output classes \citep{imagenet}.
\end{itemize}

Each of our results for a given rewound point, $k$, is averaged over 3 runs.

\paragraph{Datasets}
\label{datasets}
For our set of experiments we used the CIFAR-10 and CIFAR-100 datasets~\citep{krizhevsky2009learning}. We apply the following standard data augmentation techniques to the training set:

\begin{itemize}
    \item \texttt{RandomHorizontalFlip}: Randomly flips the image horizontally with a given probability (by default, $50\%$).
    \item \texttt{RandomCrop}: Randomly crops the image to a size of $32 \times32$ pixels, with a padding of $4$ pixels around the image.

\end{itemize}
\paragraph{Optimizers}
\label{optim}
We use the following hyperparameters for ResNet20 and VGG11 trained on CIFAR-10/100, as outlined in~\cref{dense_and_sparse_hparams}.
% \begin{itemize}
%     \item \textbf{Optimizer}: SGD
%     \item \textbf{Momentum}: $0.9$
%     \item \textbf{Learning Rate}: $0.08$ (used for dense training)
%     \item \textbf{Sparse Learning Rate}: $0.02$ (used for sparse training our LTH, Naive and Permuted solutions)
%     \item \textbf{Weight Decay}: $5e-4$
%     \item \textbf{Batch Size}: $128$
% \end{itemize}
\begin{table}[ht]
\centering
\begin{tabular}{|l|l|}
\hline
\textbf{Hyperparameter}     & \textbf{Value} \\ \hline
Optimizer           & SGD            \\ \hline
Momentum            & 0.9            \\ \hline
Dense Learning Rate       & 0.08   \\ \hline
Sparse Learning Rate & 0.02  \\ \hline
Weight Decay        & $5 \times 10^{-4}$ \\ \hline
Batch Size          & 128            \\ \hline
Epochs ($T$)          & 200           \\ \hline
\end{tabular}
\vspace{0.5em}
\caption{Hyperparameters for dense and sparse training of both ResNet20 and VGG11.}
\label{dense_and_sparse_hparams}
\end{table}

\subsection{Implementation Details for ResNet50 on ImageNet}
\paragraph{Architecture} We utilize the standard ResNet50 implementation provided by torchvision and customize PyTorch's distributed data parallel codebase for training models on ImageNet~\citep{pytorch}.

\paragraph{Dataset} For our set of experiments we used the ImageNet dataset~\citep{imagenet}. We apply the following standard data augmentation techniques to the training set:
\begin{itemize}
    \item \texttt{RandomHorizontalFlip}: Randomly flips the image horizontally with a given probability (by default, $50\%$).
    \item \texttt{RandomResizedCrop}: Randomly crops a region from the image and resizes it to $224 \times 224$ pixels.
\end{itemize}

\paragraph{Optimizers}
% \label{optim}
We use the following hyperparameters for ResNet50 trained on ImageNet, as outlined in~\cref{dense_and_sparse_hparams_resnet50}.
\begin{table}[ht]
\centering
\begin{tabular}{|l|l|}
\hline
\textbf{Hyperparameter}     & \textbf{Value} \\ \hline
Optimizer           & SGD            \\ \hline
Momentum            & 0.9            \\ \hline
Dense Learning Rate       & 0.4  \\ \hline
Sparse Learning Rate & 0.4  \\ \hline
Weight Decay        & $1 \times 10^{-4}$ \\ \hline
Batch Size          & 1024            \\ \hline
Epochs ($T$)          & 80           \\ \hline
\end{tabular}
\vspace{0.5em}
\caption{Hyperparameters for dense and sparse training of ResNet50.}
\label{dense_and_sparse_hparams_resnet50}
\end{table}
\subsection{Pruning}
\label{pruning}
We apply standard \gls{impft}~\citep{frankle2019lottery,han2015learning, renda2020} to obtain our final mask, $\textbf{m}_A$, producing a sparse subnetwork $\textbf{w}_A^{t=T} \odot \textbf{m}_A$. For pruning, we utilize PyTorch's \texttt{torch.nn.utils.prune} library \citep{paganini2020streamlining}.
\begin{enumerate} 
    \item In an unstructured, global manner, we identify and mask (set to zero) the smallest 20\% of unpruned weights based on their magnitude.
    \item This process is repeated for $s$ rounds to achieve the target sparsity $S$, with each subsequent round pruning 20\% of the remaining weights.
    \item During each round, the model is trained for $\texttt{train\_epochs\_per\_prune}$ epochs.
\end{enumerate}
\begin{table}[ht]
\centering
\begin{tabular}{|l|l|l|}
\hline
\textbf{Hyperparameter}        & \textbf{ResNet20/VGG11} & \textbf{ResNet50} \\ \hline
\texttt{train\_epochs\_per\_prune} & 50                       & 20                            \\ \hline
Learning Rate  & 0.01          & 0.04                           \\ \hline
\end{tabular}
\vspace{0.5em}
\caption{\small{Hyperparameters used for pruning ResNet20/VGG11 on CIFAR-10/100 and ResNet50 on ImageNet.}}
\label{pruning_table}
\end{table}

% \subsection{Algorithm Details}
% The method outlined in \cref{fig:workflow}, is also detailed fully in \cref{algorithm}.
% \begin{algorithm}
% \caption{Our training paradigm}\label{algorithm}
% \begin{algorithmic}[1]
%     \State Randomly initialize two distinct neural networks from the same distribution: \( \textbf{w}_A^{t=0}, \textbf{w}_B^{t=0} \sim \mathcal{N} \).
%     \State Train both networks to convergence for $T$ epochs: \( \textbf{w}_A^{t=T}, \textbf{w}_B^{t=T} \).
%     \State Prune \(\textbf{w}_A^{t=T}\) via \gls{imp} (without weight-rewinding) producing a sparse subnetwork: \(\textbf{w}_A^{t=T} \odot \textbf{m}_A\).
%     \State Perform activation matching by aligning the activations of \( \mathbf{w}_A^{t=T} \) to \( \mathbf{w}_B^{t=T} \), such that \( \mathcal{B}(\pi(\mathbf{w}_A^{t=T}), \mathbf{w}_B^{t=T}) \leq \epsilon \implies \) loss barrier below some threshold \(\implies\) \gls{lmc}. 
%     \State Save checkpoints from 2. at some epoch $t =k \ll T$, resulting in rewound epochs: \( \textbf{w}_A^{t=k}, \textbf{w}_B^{t=k}\).
%     \State Sparse train the LTH solution: \(\mathbf{w}_A^{t=k} \odot \textbf{m}_A\) for $T -k$ epochs.
%     \State Sparse train the naive solution using the wrong initialization: \(\mathbf{w}_B^{t=k} \odot \textbf{m}_A\) for $T -k$ epochs.
%     \State Sparse train the permuted solution using the permuted mask: \(\mathbf{w}_B^{t=k} \odot \pi(\textbf{m}_A)\) for $T -k$ epochs.
% \end{algorithmic}
% \end{algorithm}
\subsection{Results}
\label{subsec:tables}
Detailed results for ResNet20$\times\{w\}$/CIFAR-10 are provided in~\cref{table:resnet20_1_CIFAR10,table:resnet20_4_CIFAR10,table:resnet20_8_CIFAR10,table:resnet20_16_CIFAR10}, for VGG11$\times\{1\}$/CIFAR-10 in~\cref{table:vgg11_CIFAR10}, for ResNet50$\times\{1\}$/ImageNet in~\cref{table:resnet50_imagenet}, and for ResNet20$\times\{w\}$/CIFAR-100 in~\cref{table:resnet20_1_CIFAR100,table:resnet20_4_CIFAR100,table:resnet20_8_CIFAR100,table:resnet20_16_CIFAR100}.

\begin{table}[tbp]
\centering
\caption{\small{\textbf{ResNet20$\times\{1\}$/CIFAR-10}. Results using the ResNet20$\times\{1\}$ trained on CIFAR-10, from a rewind point $k$, using various methods of sparse training with sparsity $S$. \Gls{lth} trains within the original dense/pruned solution basin, while naive/permuted train from a new random initialization.}}\label{table:resnet20_1_CIFAR10}
\resizebox{\textwidth}{!}{\begin{tabular}{@{}p{1.3em}lccccccccc@{}}
    
    \toprule
    & & \multicolumn{9}{c}{Rewind Epoch $k$} \\
         \cmidrule(l){3-11}
     $S$ & Method & $k=$ 0 & 5 & 10 & 15 & 20 & 25 & 50 & 75 & 100\\
     \midrule
    \multirow{3}{2em}{{80\%}} & {LTH} &90.41 $\pm$ 0.14 &92.12 $\pm$ 0.25 &92.08 $\pm$ 0.36 &92.10 $\pm$ 0.27 &92.25 $\pm$ 0.14 &92.32 $\pm$ 0.26 &92.15 $\pm$ 0.13 &92.26 $\pm$ 0.19 &92.21 $\pm$ 0.16 \\
    \cmidrule(lr){2-11}
    & {naive} &89.67 $\pm$ 0.35 &89.74 $\pm$ 0.69 &90.16 $\pm$ 0.14 &90.07 $\pm$ 0.09 &90.13 $\pm$ 0.11 &90.40 $\pm$ 0.11 &90.66 $\pm$ 0.12 &90.31 $\pm$ 0.27 &90.45 $\pm$ 0.22 \\
    & {perm.} &89.74 $\pm$ 0.05 &90.15 $\pm$ 0.16 &90.26 $\pm$ 0.08 &90.72 $\pm$ 0.12 &90.68 $\pm$ 0.18 &90.72 $\pm$ 0.28 &90.76 $\pm$ 0.27 &\textcolor{blue}{\textbf{91.13 $\pm$ 0.06}} &90.82 $\pm$ 0.21\\
    \midrule
    \multirow{3}{2em}{{90\%}} & {LTH} &89.45 $\pm$ 0.10 &91.27 $\pm$ 0.37 &91.34 $\pm$ 0.29 &91.34 $\pm$ 0.09 &91.18 $\pm$ 0.27 &91.43 $\pm$ 0.22 &91.44 $\pm$ 0.12 &91.36 $\pm$ 0.18 &91.68 $\pm$ 0.28 
    \\\cmidrule(lr){2-11}
                            & {naive} &88.47 $\pm$ 0.21 &88.70 $\pm$ 0.14 &88.77 $\pm$ 0.21 &88.84 $\pm$ 0.43 &88.83 $\pm$ 0.27 &88.78 $\pm$ 0.02 &88.99 $\pm$ 0.08 &88.81 $\pm$ 0.17 &88.82 $\pm$ 0.07  \\
                            & {perm.} &88.59 $\pm$ 0.11 &89.09 $\pm$ 0.22 &89.56 $\pm$ 0.28 &89.71 $\pm$ 0.12 &89.50 $\pm$ 0.27 &89.97 $\pm$ 0.13 &89.84 $\pm$ 0.15 &\textcolor{blue}{\textbf{90.03 $\pm$ 0.07}} &89.77 $\pm$ 0.15  \\
    \midrule
    \multirow{3}{2em}{{95\%}} & {LTH} &87.83 $\pm$ 0.38 &90.33 $\pm$ 0.22 &90.39 $\pm$ 0.28 &90.37 $\pm$ 0.21 &90.58 $\pm$ 0.26 &90.43 $\pm$ 0.20 &90.56 $\pm$ 0.29 &90.44 $\pm$ 0.26 &90.40 $\pm$ 0.19 
    \\\cmidrule(lr){2-11}
                            & {naive} &86.89 $\pm$ 0.21 &87.01 $\pm$ 0.23 &86.88 $\pm$ 0.13 &87.28 $\pm$ 0.19 &87.31 $\pm$ 0.36 &87.00 $\pm$ 0.19 &86.88 $\pm$ 0.08 &86.99 $\pm$ 0.29 &86.50 $\pm$ 0.22  \\
                            & {perm.} &87.24 $\pm$ 0.22 &87.70 $\pm$ 0.08 &87.92 $\pm$ 0.25 &88.23 $\pm$ 0.52 &\textcolor{blue}{\textbf{88.29 $\pm$ 0.52 }} & 88.24 $\pm$ 0.20 &88.21 $\pm$ 0.30 &88.21 $\pm$ 0.20 &88.04 $\pm$ 0.22 \\
    \midrule
    \multirow{3}{2em}{{97\%}} & {LTH} &86.03 $\pm$ 0.22 &88.00 $\pm$ 0.02 &88.73 $\pm$ 0.05 &89.00 $\pm$ 0.24 &89.21 $\pm$ 0.23 &89.27 $\pm$ 0.14 &89.03 $\pm$ 0.27 &89.12 $\pm$ 0.25 &89.06 $\pm$ 0.21
    \\\cmidrule(lr){2-11}
                            & {naive} &85.60 $\pm$ 0.38 &85.43 $\pm$ 0.40 &85.89 $\pm$ 0.37 &85.48 $\pm$ 0.13 &85.36 $\pm$ 0.14 &85.70 $\pm$ 0.21 &85.30 $\pm$ 0.32 &85.14 $\pm$ 0.29 &84.64 $\pm$ 0.34  \\
                            & {perm.} &85.61 $\pm$ 0.48 &85.93 $\pm$ 0.34 &86.26 $\pm$ 0.40 &\textcolor{blue}{\textbf{86.48 $\pm$ 0.39}} &86.12 $\pm$ 0.27 &86.16 $\pm$ 0.14 & 86.43 $\pm$ 0.27 &86.06 $\pm$ 0.26 &85.95 $\pm$ 0.14 \\

    \bottomrule
\end{tabular}}
\end{table}

\begin{table}[tbp]
\centering
\caption{\small{\textbf{ResNet20$\times\{4\}$/CIFAR-10}. Results using the ResNet20$\times\{4\}$ trained on CIFAR-10, from a rewind point $k$, using various methods of sparse training with sparsity $S$. \Gls{lth} trains within the original dense/pruned solution basin, while naive/permuted train from a new random initialization. Note this table is the same setting as~\cref{table:resnet20_1_CIFAR10} except $w = 4$.}}\label{table:resnet20_4_CIFAR10}
\resizebox{\textwidth}{!}{\begin{tabular}{@{}llccccccccc@{}}
    
    \toprule
    & & \multicolumn{9}{c}{Rewind Epoch $k$} \\
         \cmidrule(lr){3-11}
     $S$ & Method & $k$=0 & 5 & 10 & 15 & 20 & 25 & 50 & 75 & 100\\
     \midrule
    \multirow{3}{2em}{{80\%}} & {LTH} &94.67 $\pm$ 0.14 &95.57 $\pm$ 0.05 &95.84 $\pm$ 0.15 &95.80 $\pm$ 0.12 &95.88 $\pm$ 0.20 &95.72 $\pm$ 0.09 &95.81 $\pm$ 0.10 &95.83 $\pm$ 0.21 &95.71 $\pm$ 0.16 \\
        \cmidrule(lr){2-11}
    & {naive} &94.36 $\pm$ 0.04 &94.55 $\pm$ 0.14 &94.59 $\pm$ 0.29 &94.74 $\pm$ 0.13 &94.69 $\pm$ 0.09 &94.81 $\pm$ 0.06 &95.07 $\pm$ 0.17 &95.02 $\pm$ 0.11 &94.97 $\pm$ 0.21 \\
    & {perm.} &94.39 $\pm$ 0.19 &94.88 $\pm$ 0.28 &95.15 $\pm$ 0.14 &95.20 $\pm$ 0.16 &95.17 $\pm$ 0.21 &95.28 $\pm$ 0.29 &\textcolor{blue}{\textbf{95.43 $\pm$ 0.14}} & 95.40 $\pm$ 0.10 &95.30 $\pm$ 0.08 \\
    \midrule
    \multirow{3}{2em}{{90\%}} & {LTH} &94.43 $\pm$ 0.17 &95.53 $\pm$ 0.21 &95.63 $\pm$ 0.07 &95.65 $\pm$ 0.30 &95.66 $\pm$ 0.07 &95.61 $\pm$ 0.14 &95.56 $\pm$ 0.16 &95.62 $\pm$ 0.14 &95.50 $\pm$ 0.04 
\\
    \cmidrule(lr){2-11}

                            & {naive} &93.79 $\pm$ 0.15 &93.96 $\pm$ 0.05 &94.09 $\pm$ 0.11 &94.20 $\pm$ 0.29 &94.35 $\pm$ 0.25 &94.20 $\pm$ 0.13 &94.27 $\pm$ 0.19 &94.23 $\pm$ 0.08 &94.19 $\pm$ 0.27 \\
                            & {perm.} &93.97 $\pm$ 0.29 &94.64 $\pm$ 0.13 &94.73 $\pm$ 0.17 &94.93 $\pm$ 0.12 &94.92 $\pm$ 0.11 &94.90 $\pm$ 0.07 &95.04 $\pm$ 0.14 &\textcolor{blue}{\textbf{95.07 $\pm$ 0.18}} &94.91 $\pm$ 0.19 \\
    \midrule
    \multirow{3}{2em}{{95\%}} & {LTH} &93.65 $\pm$ 0.12 &95.26 $\pm$ 0.08 &95.39 $\pm$ 0.05 &95.32 $\pm$ 0.18 &95.26 $\pm$ 0.03 &95.33 $\pm$ 0.07 &95.40 $\pm$ 0.14 &95.19 $\pm$ 0.05 &95.37 $\pm$ 0.21  \\
        \cmidrule(lr){2-11}

                            & {naive} &93.27 $\pm$ 0.07 &93.30 $\pm$ 0.11 &93.63 $\pm$ 0.04 &93.61 $\pm$ 0.21 &93.66 $\pm$ 0.13 &93.67 $\pm$ 0.14 &93.43 $\pm$ 0.21 &93.51 $\pm$ 0.32 &93.14 $\pm$ 0.03 \\
                            & {perm.} &93.54 $\pm$ 0.24 &94.17 $\pm$ 0.07 &94.46 $\pm$ 0.10 &94.27 $\pm$ 0.19 &94.61 $\pm$ 0.07 &94.54 $\pm$ 0.07 &\textcolor{blue}{\textbf{94.75 $\pm$ 0.11}} &\textcolor{blue}{\textbf{94.75 $\pm$ 0.09}} &94.54 $\pm$ 0.27\\
    \midrule
    \multirow{3}{2em}{{97\%}} & {LTH} &93.00 $\pm$ 0.11 &94.77 $\pm$ 0.09 &94.86 $\pm$ 0.06 &94.94 $\pm$ 0.17 &94.96 $\pm$ 0.06 &94.89 $\pm$ 0.21 &95.00 $\pm$ 0.24 &94.94 $\pm$ 0.10 &94.97 $\pm$ 0.13 \\
        \cmidrule(lr){2-11}

                            & {naive} &92.63 $\pm$ 0.12 &92.80 $\pm$ 0.10 &92.85 $\pm$ 0.21 &92.66 $\pm$ 0.21 &92.74 $\pm$ 0.11 &92.69 $\pm$ 0.14 &92.28 $\pm$ 0.09 &92.02 $\pm$ 0.18 &91.87 $\pm$ 0.10 \\
                            & {perm.} &92.81 $\pm$ 0.27 &93.54 $\pm$ 0.08 &93.83 $\pm$ 0.12 &93.75 $\pm$ 0.34 &94.00 $\pm$ 0.33 &94.12 $\pm$ 0.04 &94.07 $\pm$ 0.31 &\textcolor{blue}{\textbf{94.32 $\pm$ 0.24}} &94.14 $\pm$ 0.04

\\

    \bottomrule
\end{tabular}}
\end{table}

\begin{table}[tbp]
\centering
\caption{\small{\textbf{ResNet20$\times\{8\}$/CIFAR-10}. Results using the ResNet20$\times\{8\}$ trained on CIFAR-10, from a rewind point $k$, using various methods of sparse training with sparsity $S$. \Gls{lth} trains within the original dense/pruned solution basin, while naive/permuted train from a new random initialization. Note this table is the same setting as~\cref{table:resnet20_1_CIFAR10} except $w = 8$.}}\label{table:resnet20_8_CIFAR10}
\begin{tabular}{@{}llccccc@{}}
    
    \toprule
    & & \multicolumn{5}{c}{Rewind Epoch $k$} \\
         \cmidrule(lr){3-7}
     $S$ & Method & $k$=0 & 10 & 25 & 50 & 100\\
     \midrule
    \multirow{3}{2em}{{80\%}} & {LTH} &95.35 $\pm$ 0.07 &95.98 $\pm$ 0.14 &96.12 $\pm$ 0.04 &96.10 $\pm$ 0.20 &96.21 $\pm$ 0.06 \\
        \cmidrule(lr){2-7}

                            & {naive} &95.17 $\pm$ 0.17 &95.32 $\pm$ 0.13 &95.63 $\pm$ 0.13 &95.62 $\pm$ 0.08 &95.79 $\pm$ 0.15 \\
                            & {perm.} &95.36 $\pm$ 0.14 &95.60 $\pm$ 0.15 &95.89 $\pm$ 0.19 &\textcolor{blue}{\textbf{95.94 $\pm$ 0.17}} &\textcolor{blue}{\textbf{95.94 $\pm$ 0.06}} \\
    \midrule
    \multirow{3}{2em}{{90\%}} & {LTH} &94.96 $\pm$ 0.18 &95.97 $\pm$ 0.15 &96.02 $\pm$ 0.05 &96.00 $\pm$ 0.19 &96.12 $\pm$ 0.10 \\
        \cmidrule(lr){2-7}

                            & {naive} &95.05 $\pm$ 0.07 &95.12 $\pm$ 0.03 &95.20 $\pm$ 0.22 &95.44 $\pm$ 0.14 &95.06 $\pm$ 0.25 \\
                            & {perm.} &95.05 $\pm$ 0.05 &95.58 $\pm$ 0.06 &95.78 $\pm$ 0.12 &\textcolor{blue}{\textbf{95.87 $\pm$ 0.13}} &95.85 $\pm$ 0.11\\
    \midrule
    \multirow{3}{2em}{{95\%}} & {LTH} &94.86 $\pm$ 0.08 &95.90 $\pm$ 0.15 &95.93 $\pm$ 0.26 &96.07 $\pm$ 0.25 &96.00 $\pm$ 0.25 \\
        \cmidrule(lr){2-7}

                            & {naive} &94.60 $\pm$ 0.14 &94.84 $\pm$ 0.13 &94.93 $\pm$ 0.17 &95.01 $\pm$ 0.33 &94.59 $\pm$ 0.52 \\
                            & {perm.} &94.85 $\pm$ 0.19 &95.29 $\pm$ 0.27 &95.63 $\pm$ 0.11 &\textcolor{blue}{\textbf{95.67 $\pm$ 0.16}} &95.59 $\pm$ 0.22 \\
    \midrule
    \multirow{3}{2em}{{97\%}} & {LTH} &94.54 $\pm$ 0.23 &95.79 $\pm$ 0.14 &95.87 $\pm$ 0.03 &95.78 $\pm$ 0.21 &95.90 $\pm$ 0.04 \\
        \cmidrule(lr){2-7}

                            & {naive} &94.39 $\pm$ 0.04 &94.39 $\pm$ 0.04 &94.49 $\pm$ 0.18 &94.19 $\pm$ 0.11 &93.83 $\pm$ 0.08 \\
                            & {perm.} &94.46 $\pm$ 0.14 &95.26 $\pm$ 0.10 &95.16 $\pm$ 0.26 &\textcolor{blue}{\textbf{95.56 $\pm$ 0.06}} &95.45 $\pm$ 0.05 \\

    \bottomrule
\end{tabular}
\end{table}

\begin{table}[tbp]
\centering
\caption{\small{\textbf{ResNet20$\times\{16\}$/CIFAR-10}. Results using the ResNet20$\times\{8\}$ trained on CIFAR-10, from a rewind point $k$, using various methods of sparse training with sparsity $S$. \Gls{lth} trains within the original dense/pruned solution basin, while naive/permuted train from a new random initialization. Note this table is the same setting as~\cref{table:resnet20_1_CIFAR10} except $w = 16$.}}\label{table:resnet20_16_CIFAR10}
\begin{tabular}{@{}llccccc@{}}
    
    \toprule
    & & \multicolumn{5}{c}{Rewind Epoch $k$} \\
         \cmidrule(lr){3-7}
     $S$ & Method & $k$=0 & 10 & 25 & 50 & 100\\
     \midrule
    \multirow{3}{2em}{{80\%}} & {LTH} &95.62 $\pm$ 0.19 &95.84 $\pm$ 0.36 &96.05 $\pm$ 0.34 &96.31 $\pm$ 0.18 &96.36 $\pm$ 0.24 \\
        \cmidrule(lr){2-7}

                            & {naive} &95.47 $\pm$ 0.15 &95.71 $\pm$ 0.22 &95.71 $\pm$ 0.26 &96.09 $\pm$ 0.04 &95.99 $\pm$ 0.21 \\
                            & {perm.} &95.77 $\pm$ 0.11 &95.79 $\pm$ 0.29 &96.00 $\pm$ 0.14 &\textcolor{blue}{\textbf{96.24 $\pm$ 0.11}} &96.21 $\pm$ 0.06\\
    \midrule
    \multirow{3}{2em}{{90\%}} & {LTH} &95.59 $\pm$ 0.22 &96.10 $\pm$ 0.48 &96.19 $\pm$ 0.49 &96.18 $\pm$ 0.20 &96.41 $\pm$ 0.14 \\
        \cmidrule(lr){2-7}

                            & {naive} &95.37 $\pm$ 0.09 &95.47 $\pm$ 0.13 &95.66 $\pm$ 0.01 &95.70 $\pm$ 0.13 &95.76 $\pm$ 0.14 \\
                            & {perm.} &95.58 $\pm$ 0.22 &95.80 $\pm$ 0.14 &96.11 $\pm$ 0.13 &\textcolor{blue}{\textbf{96.17 $\pm$ 0.17}} &96.04 $\pm$ 0.05\\
    \midrule
    \multirow{3}{2em}{{95\%}} & {LTH} &95.08 $\pm$ 0.21 &95.96 $\pm$ 0.39 &96.12 $\pm$ 0.21 &96.16 $\pm$ 0.30 &96.26 $\pm$ 0.23 \\
        \cmidrule(lr){2-7}

                            & {naive} &95.27 $\pm$ 0.13 &95.43 $\pm$ 0.09 &95.57 $\pm$ 0.37 &95.63 $\pm$ 0.25 &95.27 $\pm$ 0.55 \\
                            & {perm.} &95.39 $\pm$ 0.26 &96.02 $\pm$ 0.22 &96.12 $\pm$ 0.18 &\textcolor{blue}{\textbf{96.18 $\pm$ 0.18}} &96.06 $\pm$ 0.09\\
    \midrule
    \multirow{3}{2em}{{97\%}} & {LTH} &95.19 $\pm$ 0.27 &95.84 $\pm$ 0.25 &96.14 $\pm$ 0.30 &96.12 $\pm$ 0.27 &96.17 $\pm$ 0.33 \\
        \cmidrule(lr){2-7}

                            & {naive} &94.94 $\pm$ 0.04 &95.06 $\pm$ 0.17 &95.29 $\pm$ 0.15 &95.13 $\pm$ 0.19 &94.35 $\pm$ 0.45 \\
                            & {perm.} &95.07 $\pm$ 0.06 &95.51 $\pm$ 0.22 &95.88 $\pm$ 0.14 &\textcolor{blue}{\textbf{95.90 $\pm$ 0.24}} &95.88 $\pm$ 0.09 \\

    \bottomrule
\end{tabular}
\end{table}

\begin{table}[tbp]
\centering
\caption{\small{\textbf{VGG11$\times \{1\}$/CIFAR-10}. Results using the VGG11 trained on CIFAR-10, from a rewind point $k$, using various methods of sparse training with sparsity $S$. \Gls{lth} trains within the original dense/pruned solution basin, while naive/permuted train from a new random initialization. }}\label{table:vgg11_CIFAR10}
\begin{tabular}{@{}llccccccc@{}}
    \toprule
    & & \multicolumn{7}{c}{Rewind Epoch $k$} \\
    \cmidrule(l){3-9}
    $S$ & Method & $k=$ 0 & 5 & 10 & 15 & 20 & 25 & 50\\
    \midrule
    \multirow{3}{2em}{80\%} & LTH & 89.94 $\pm$ 0.06 & 90.44 $\pm$ 0.17 & 90.91 $\pm$ 0.12 & 90.87 $\pm$ 0.16 & 91.14 $\pm$ 0.28 & 91.11 $\pm$ 0.08 & 91.22 $\pm$ 0.08\\
    \cmidrule(lr){2-9}
    & naive & 89.70 $\pm$ 0.13 & 89.90 $\pm$ 0.18 & 90.04 $\pm$ 0.07 & 90.34 $\pm$ 0.16 & 90.48 $\pm$ 0.19 & 90.55 $\pm$ 0.17 & 90.87 $\pm$ 0.19 \\
    & perm. & 89.94 $\pm$ 0.1 & 90.18 $\pm$ 0.08 & 90.52 $\pm$ 0.17 & 90.71 $\pm$ 0.22 & 90.77 $\pm$ 0.19 & 90.81 $\pm$ 0.19 & \textcolor{blue}{\textbf{91.07 $\pm$ 0.21}} \\
    \midrule
    \multirow{3}{2em}{90\%} & LTH & 89.33 $\pm$ 0.16 & 90.82 $\pm$ 0.09 & 90.97 $\pm$ 0.14& 91.05 $\pm$ 0.04 & 91.15 $\pm$ 0.11 & 90.91 $\pm$ 0.17 & 91.08 $\pm$ 0.31 \\
    \cmidrule(lr){2-9}
    & naive & 89.17 $\pm$ 0.2 & 89.55 $\pm$ 0.02 & 89.81 $\pm$ 0.02 & 89.49 $\pm$ 0.05& 89.68 $\pm$ 0.11 & 89.80 $\pm$ 0.03 & 89.80 $\pm$ 0.05 \\
    & perm. & 89.30 $\pm$ 0.02& 90.33 $\pm$ 0.08 & 90.44 $\pm$ 0.14 & 90.46 $\pm$ 0.04 & 90.75 $\pm$ 0.22 & 90.76 $\pm$ 0.12 & \textcolor{blue}{\textbf{91.01 $\pm$ 0.06}} \\
    \bottomrule
\end{tabular}
\end{table}

\begin{table}[tbp]
\centering
\caption{\small{\textbf{ResNet50$\times \{1\}$/ImageNet}. Top-1 and Top-5 Accuracies of ResNet50$\times \{1\}$ trained on ImageNet, from a rewind point $k$, using various methods of sparse training with sparsity $S$.}}\label{table:resnet50_imagenet}
\vspace{0.5cm} 
\begin{tabular}{@{}llccc|ccc@{}}
    \toprule
    & & \multicolumn{3}{c|}{Top-1 Accuracy} & \multicolumn{3}{c}{Top-5 Accuracy} \\
    \cmidrule(lr){3-5} \cmidrule(l){6-8}
    $S$ & Method & $k=$ 10 & 25 & 50 & $k=$ 10 & 25 & 50\\
    \midrule
    \multirow{3}{*}{80\%} 
    & LTH   & 72.87 & 72.16 & 65.23 & 91.13 & 90.66 & 86.65 \\
    \cmidrule(lr){2-8}
    & naive & 69.13 & 68.94 & 60.30 & 88.85 & 88.1 & 83.22 \\
    & perm. & \textcolor{blue}{\textbf{69.87}} & 69.85 & 61.14 & 89.16 & \textcolor{blue}{\textbf{89.45}} & 84.04 \\
    \midrule
    \multirow{3}{*}{90\%} 
    & LTH   & 71.40 & 70.74 & 60.62 & 90.27 & 90.00 & 83.94 \\
    \cmidrule(lr){2-8}
    & naive & 65.49 & 64.77 & 54.46 & 86.55 & 86.26 & 79.07 \\
    & perm. & 66.25 & \textcolor{blue}{\textbf{66.37}} & 57.40 & 87.23 & \textcolor{blue}{\textbf{87.37}} & 81.45 \\
    \midrule
    \multirow{3}{*}{95\%} 
    & LTH   & 68.61 & 68.07 & 59.83 & 89.03 & 88.25 & 82.96 \\
    \cmidrule(lr){2-8}
    & naive & 61.39 & 60.77 & 51.78 & 83.79 & 83.58 & 76.79 \\
    & perm. & 62.48 & \textcolor{blue}{\textbf{62.77}} & 52.98 & 84.51 & \textcolor{blue}{\textbf{84.79}} & 78.11 \\
    \bottomrule
\end{tabular}
\end{table}

% \begin{table}[tbp]
% \centering
% \caption{\small{\textbf{ResNet50$\times \{2\}$/ImageNet}. Top-1 and Top-5 Accuracies of ResNet50$\times \{2\}$ trained on ImageNet, from a rewind point $k$, using various methods of sparse training with sparsity $S$.}}\label{table:resnet50x2_imagenet}
% \vspace{0.5cm} 
% \begin{tabular}{@{}llccc|ccc@{}}
%     \toprule
%     & & \multicolumn{3}{c|}{Top-1 Accuracy} & \multicolumn{3}{c}{Top-5 Accuracy} \\
%     \cmidrule(lr){3-5} \cmidrule(l){6-8}
%     $S$ & Method & $k=$ 10 & 25 & 50 & $k=$ 10 & 25 & 50 \\
%     \midrule
%     \multirow{3}{*}{80\%} 
%     & LTH   & 76.20 & -- & -- & 92.65 & -- & -- \\
%     \cmidrule(lr){2-8}
%     & naive & 76.15 & -- & -- & 92.85 & -- & -- \\
%     & perm. & \textcolor{blue}{\textbf{76.52}} & -- & -- & \textcolor{blue}{\textbf{92.99}} & -- & -- \\
%     \midrule
%     \multirow{3}{*}{90\%} 
%     & LTH   & 74.96 & -- & -- & 92.23 & -- & -- \\
%     \cmidrule(lr){2-8}
%     & naive & 75.11 & -- & -- & 92.25 & -- & -- \\
%     & perm. & \textcolor{blue}{\textbf{75.38}} & -- & -- & \textcolor{blue}{\textbf{92.45}} & -- & -- \\
%     \bottomrule
% \end{tabular}
% \end{table}

\begin{table}[tbp]
\centering
\caption{\small{\textbf{ResNet20$\times\{1\}$/CIFAR-100}. Results using the ResNet20$\times\{1\}$ trained on CIFAR-100, from a rewind point $k$, using various methods of sparse training with sparsity $S$. \Gls{lth} trains within the original dense/pruned solution basin, while naive/permuted train from a new random initialization.}}\label{table:resnet20_1_CIFAR100}
\begin{tabular}{@{}llccccc@{}}
    
    \toprule
    & & \multicolumn{5}{c}{Rewind Epoch $k$} \\
         \cmidrule(lr){3-7}
     $S$ & Method & $k$=0 & 10 & 25 & 50 & 100\\
     \midrule
    \multirow{3}{2em}{{80\%}} & {LTH} &63.69 $\pm$ 0.41 &67.67 $\pm$ 0.08 &67.66 $\pm$ 0.25 &67.82 $\pm$ 0.17 &67.73 $\pm$ 0.38 \\
        \cmidrule(lr){2-7}

                            & {naive} &62.89 $\pm$ 0.16 &63.37 $\pm$ 0.09 &63.07 $\pm$ 0.44 &63.36 $\pm$ 0.27 &63.33 $\pm$ 0.35 \\
                            & {perm.} &63.04 $\pm$ 0.24 &64.07 $\pm$ 0.15 &\textcolor{blue}{\textbf{64.71 $\pm$ 0.10}} &64.52 $\pm$ 0.78 &64.57 $\pm$ 0.49\\
    \midrule
    \multirow{3}{2em}{{90\%}} & {LTH} &59.81 $\pm$ 0.29 &65.21 $\pm$ 0.17 &65.15 $\pm$ 0.28 &65.10 $\pm$ 0.30 &65.17 $\pm$ 0.21 \\
        \cmidrule(lr){2-7}

                            & {naive} &58.77 $\pm$ 0.28 &59.59 $\pm$ 0.18 &59.44 $\pm$ 0.27 &59.19 $\pm$ 0.41 &58.58 $\pm$ 0.16 \\
                            & {perm.} &59.32 $\pm$ 0.32 &60.60 $\pm$ 0.79 &61.32 $\pm$ 0.33 &\textcolor{blue}{\textbf{61.53 $\pm$ 0.65}} &60.93 $\pm$ 0.51\\
    \midrule
    \multirow{3}{2em}{{95\%}} & {LTH} &55.71 $\pm$ 0.52 &61.08 $\pm$ 0.54 &61.73 $\pm$ 0.18 &61.65 $\pm$ 0.37 &61.68 $\pm$ 0.18 \\
        \cmidrule(lr){2-7}

                            & {naive} &54.04 $\pm$ 0.29 &55.20 $\pm$ 0.39 &54.65 $\pm$ 0.38 &54.96 $\pm$ 0.57 &53.97 $\pm$ 0.91\\
                            & {perm.} &55.12 $\pm$ 0.17 &56.93 $\pm$ 0.26 &\textcolor{blue}{\textbf{57.64 $\pm$ 0.36}} &57.47 $\pm$ 0.66 &57.13 $\pm$ 0.34 \\
    \midrule
    \multirow{3}{2em}{{97\%}} & {LTH} &51.10 $\pm$ 0.34 &56.14 $\pm$ 0.56 &56.92 $\pm$ 0.25 &56.94 $\pm$ 0.13 &56.93 $\pm$ 0.06 \\
        \cmidrule(lr){2-7}

                            & {naive} &49.70 $\pm$ 0.64 &49.60 $\pm$ 0.25 &49.49 $\pm$ 0.32 &49.16 $\pm$ 0.21 &47.70 $\pm$ 0.83 \\
                            & {perm.} &50.34 $\pm$ 0.21 &51.55 $\pm$ 0.69 &51.88 $\pm$ 1.08 &\textcolor{blue}{\textbf{52.64 $\pm$ 0.34}} &50.96 $\pm$ 1.15 \\

    \bottomrule
\end{tabular}
\end{table}

\begin{table}[tbp]
\centering
\caption{\small{\textbf{ResNet20$\times\{4\}$/CIFAR-100}. Results using the ResNet20$\times\{4\}$ trained on CIFAR-100, from a rewind point $k$, using various methods of sparse training with sparsity $S$. \Gls{lth} trains within the original dense/pruned solution basin, while naive/permuted train from a new random initialization. Note this table is the same setting as~\cref{table:resnet20_1_CIFAR100} except $w = 4$.}}\label{table:resnet20_4_CIFAR100}
\begin{tabular}{@{}llccccc@{}}
    
    \toprule
    & & \multicolumn{5}{c}{Rewind Epoch $k$} \\
         \cmidrule(lr){3-7}
     $S$ & Method & $k$=0 & 10 & 25 & 50 & 100\\
     \midrule
    \multirow{3}{2em}{{80\%}} & {LTH} &74.46 $\pm$ 0.12 &77.57 $\pm$ 0.06 &77.35 $\pm$ 0.31 &77.75 $\pm$ 0.26 &77.64 $\pm$ 0.14 \\
        \cmidrule(lr){2-7}

                            & {naive} &73.30 $\pm$ 0.08 &74.10 $\pm$ 0.12 &74.98 $\pm$ 0.17 &75.21 $\pm$ 0.12 &75.20 $\pm$ 0.16 \\
                            & {perm.} &73.68 $\pm$ 0.09 &75.24 $\pm$ 0.31 &75.74 $\pm$ 0.41 &76.12 $\pm$ 0.37 &\textcolor{blue}{\textbf{76.19 $\pm$ 0.39}} \\
    \midrule
    \multirow{3}{2em}{{90\%}} & {LTH} &72.54 $\pm$ 0.57 &76.56 $\pm$ 0.11 &76.56 $\pm$ 0.32 &76.80 $\pm$ 0.34 &76.80 $\pm$ 0.21 \\
        \cmidrule(lr){2-7}

                            & {naive} &71.97 $\pm$ 0.30 &72.56 $\pm$ 0.22 &72.89 $\pm$ 0.27 &72.59 $\pm$ 0.15 &72.54 $\pm$ 0.33 \\
                            & {perm.} &72.18 $\pm$ 0.23 &74.17 $\pm$ 0.35 &74.21 $\pm$ 0.23 &74.45 $\pm$ 0.27 &\textcolor{blue}{\textbf{74.89 $\pm$ 0.47}}\\
    \midrule
    \multirow{3}{2em}{{95\%}} & {LTH} &71.16 $\pm$ 0.23 &75.41 $\pm$ 0.18 &75.53 $\pm$ 0.11 &75.68 $\pm$ 0.17 &75.76 $\pm$ 0.17 \\
        \cmidrule(lr){2-7}

                            & {naive} &70.17 $\pm$ 0.47 &70.95 $\pm$ 0.50 &70.90 $\pm$ 0.18 &71.21 $\pm$ 0.26 &69.95 $\pm$ 0.42 \\
                            & {perm.} &70.41 $\pm$ 0.07 &72.70 $\pm$ 0.21 &72.92 $\pm$ 0.39 &\textcolor{blue}{\textbf{73.65 $\pm$ 0.28}} &73.41 $\pm$ 0.18\\
    \midrule
    \multirow{3}{2em}{{97\%}} & {LTH} &69.06 $\pm$ 0.03 &74.00 $\pm$ 0.39 &74.08 $\pm$ 0.37 &74.18 $\pm$ 0.18 &74.29 $\pm$ 0.31 \\
        \cmidrule(lr){2-7}

                            & {naive} &68.40 $\pm$ 0.21 &69.26 $\pm$ 0.19 &69.06 $\pm$ 0.11 &68.67 $\pm$ 0.47 &68.42 $\pm$ 0.78 \\
                            & {perm.} &69.08 $\pm$ 0.22 &71.41 $\pm$ 0.54 &71.49 $\pm$ 0.32 &71.92 $\pm$ 0.17 &\textcolor{blue}{\textbf{72.20 $\pm$ 0.08}} \\

    \bottomrule
\end{tabular}
\end{table}

\begin{table}[tbp]
\centering
\caption{\small{\textbf{ResNet20$\times\{8\}$/CIFAR-100}. Results using the ResNet20$\times\{8\}$ trained on CIFAR-100, from a rewind point $k$, using various methods of sparse training with sparsity $S$. \Gls{lth} trains within the original dense/pruned solution basin, while naive/permuted train from a new random initialization. Note this table is the same setting as~\cref{table:resnet20_1_CIFAR100} except $w = 8$.}}\label{table:resnet20_8_CIFAR100}
\begin{tabular}{@{}llccccc@{}}
    
    \toprule
    & & \multicolumn{5}{c}{Rewind Epoch $k$} \\
         \cmidrule(lr){3-7}
     $S$ & Method & $k$=0 & 10 & 25 & 50 & 100\\
     \midrule
    \multirow{3}{2em}{{80\%}} & {LTH} &78.09 $\pm$ 0.28 &80.63 $\pm$ 0.32 &80.83 $\pm$ 0.39 &80.92 $\pm$ 0.06 &80.66 $\pm$ 0.34 \\
        \cmidrule(lr){2-7}

                            & {naive} &76.86 $\pm$ 0.17 &77.47 $\pm$ 0.35 &78.20 $\pm$ 0.61 &78.65 $\pm$ 0.33 &78.74 $\pm$ 0.39 \\
                            & {perm.} &77.34 $\pm$ 0.26 &78.82 $\pm$ 0.34 &79.20 $\pm$ 0.16 &\textcolor{blue}{\textbf{79.55 $\pm$ 0.38}} &79.54 $\pm$ 0.39\\
    \midrule
    \multirow{3}{2em}{{90\%}} & {LTH} &76.47 $\pm$ 0.43 &80.02 $\pm$ 0.07 &80.10 $\pm$ 0.13 &79.98 $\pm$ 0.33 &79.98 $\pm$ 0.20 \\
        \cmidrule(lr){2-7}

                            & {naive} &75.68 $\pm$ 0.23 &76.36 $\pm$ 0.21 &76.80 $\pm$ 0.14 &77.27 $\pm$ 0.12 &76.55 $\pm$ 0.49 \\
                            & {perm.} &76.17 $\pm$ 0.26 &77.99 $\pm$ 0.17 &78.22 $\pm$ 0.15 &78.62 $\pm$ 0.19 &\textcolor{blue}{\textbf{78.82 $\pm$ 0.17}}\\
    \midrule
    \multirow{3}{2em}{{95\%}} & {LTH} &75.38 $\pm$ 0.02 &79.42 $\pm$ 0.06 &79.24 $\pm$ 0.19 &79.35 $\pm$ 0.06 &79.29 $\pm$ 0.13 \\
        \cmidrule(lr){2-7}

                            & {naive} &74.78 $\pm$ 0.15 &75.48 $\pm$ 0.18 &75.53 $\pm$ 0.15 &75.27 $\pm$ 0.15 &74.38 $\pm$ 0.65 \\
                            & {perm.} &75.07 $\pm$ 0.14 &76.97 $\pm$ 0.46 &77.80 $\pm$ 0.14 &77.74 $\pm$ 0.51 &\textcolor{blue}{\textbf{78.04 $\pm$ 0.42}}\\
    \midrule
    \multirow{3}{2em}{{97\%}} & {LTH} &73.97 $\pm$ 0.21 &78.63 $\pm$ 0.25 &78.65 $\pm$ 0.50 &78.74 $\pm$ 0.49 &78.47 $\pm$ 0.16 \\
        \cmidrule(lr){2-7}

                            & {naive} &73.13 $\pm$ 0.26 &73.73 $\pm$ 0.12 &73.76 $\pm$ 0.27 &73.26 $\pm$ 0.07 &72.79 $\pm$ 0.46 \\
                            & {perm.} &73.81 $\pm$ 0.67 &76.29 $\pm$ 0.14 &76.38 $\pm$ 0.57 &76.57 $\pm$ 0.29 &\textcolor{blue}{\textbf{76.79 $\pm$ 0.76}} \\

    \bottomrule
\end{tabular}
\end{table}

\begin{table}[tbp]
\centering
\caption{\small{\textbf{ResNet20$\times\{16\}$/CIFAR-100}. Results using the ResNet20$\times\{16\}$ trained on CIFAR-100, from a rewind point $k$, using various methods of sparse training with sparsity $S$. \Gls{lth} trains within the original dense/pruned solution basin, while naive/permuted train from a new random initialization. Note this table is the same setting as~\cref{table:resnet20_1_CIFAR100} except $w = 16$.}}\label{table:resnet20_16_CIFAR100}
\begin{tabular}{@{}llccccc@{}}
    
    \toprule
    & & \multicolumn{5}{c}{Rewind Epoch $k$} \\
         \cmidrule(lr){3-7}
     $S$ & Method & $k$=0 & 10 & 25 & 50 & 100\\
     \midrule
    \multirow{3}{2em}{{80\%}} & {LTH} &80.21 $\pm$ 0.18 &82.32 $\pm$ 0.34 &82.40 $\pm$ 0.26 &82.48 $\pm$ 0.38 &82.16 $\pm$ 0.30 \\
        \cmidrule(lr){2-7}

                            & {naive} &79.31 $\pm$ 0.06 &79.50 $\pm$ 0.09 &80.24 $\pm$ 0.17 &81.02 $\pm$ 0.11 &81.01 $\pm$ 0.07 \\
                            & {perm.} &79.35 $\pm$ 0.11 &80.44 $\pm$ 0.40 &81.15 $\pm$ 0.48 &81.57 $\pm$ 0.38 &\textcolor{blue}{\textbf{81.81 $\pm$ 0.21}}\\
    \midrule
    \multirow{3}{2em}{{90\%}} & {LTH} &79.31 $\pm$ 0.16 &82.26 $\pm$ 0.18 &82.14 $\pm$ 0.08 &81.95 $\pm$ 0.03 &82.11 $\pm$ 0.12 \\
        \cmidrule(lr){2-7}

                            & {naive} &78.78 $\pm$ 0.37 &79.26 $\pm$ 0.11 &79.42 $\pm$ 0.51 &79.56 $\pm$ 0.26 &79.57 $\pm$ 0.13 \\
                            & {perm.} &79.20 $\pm$ 0.09 &80.49 $\pm$ 0.32 &80.59 $\pm$ 0.15 &81.12 $\pm$ 0.05 &\textcolor{blue}{\textbf{81.24 $\pm$ 0.09}}\\
    \midrule
    \multirow{3}{2em}{{95\%}} & {LTH} &78.32 $\pm$ 0.34 &81.57 $\pm$ 0.09 &81.57 $\pm$ 0.32 &81.47 $\pm$ 0.25 &81.63 $\pm$ 0.07 \\
        \cmidrule(lr){2-7}

                            & {naive} &78.01 $\pm$ 0.02 &78.53 $\pm$ 0.10 &78.45 $\pm$ 0.21 &78.38 $\pm$ 0.43 &77.49 $\pm$ 0.06 \\
                            & {perm.} &78.25 $\pm$ 0.20 &79.76 $\pm$ 0.20 &\textcolor{blue}{\textbf{80.50 $\pm$ 0.04}} &80.47 $\pm$ 0.08 &80.25 $\pm$ 0.21\\
    \midrule
    \multirow{3}{2em}{{97\%}} & {LTH} &77.49 $\pm$ 0.27 &81.07 $\pm$ 0.07 &81.06 $\pm$ 0.11 &81.11 $\pm$ 0.18 &81.14 $\pm$ 0.32 \\
        \cmidrule(lr){2-7}

                            & {naive} &76.46 $\pm$ 0.44 &76.71 $\pm$ 0.41 &77.19 $\pm$ 0.09 &76.93 $\pm$ 0.36 &75.53 $\pm$ 0.40 \\
                            & {perm.} &77.04 $\pm$ 0.38 &79.14 $\pm$ 0.17 &79.30 $\pm$ 0.21 &79.62 $\pm$ 0.14 &\textcolor{blue}{\textbf{79.63 $\pm$ 0.06}} \\

    \bottomrule
\end{tabular}
\end{table}

\vspace{-0.5em}
\subsection{Additional Plots}
\begin{figure}[tbp]
    \centering
    % First row, first figure
    \begin{subfigure}{1.5em}
        \makebox[20pt]{\raisebox{50pt}{\rotatebox[origin=c]{90}{$w=1$}}}%
    \end{subfigure}
    \begin{subfigure}{0.235\textwidth}
        \centering
        \includegraphics[width=\linewidth]{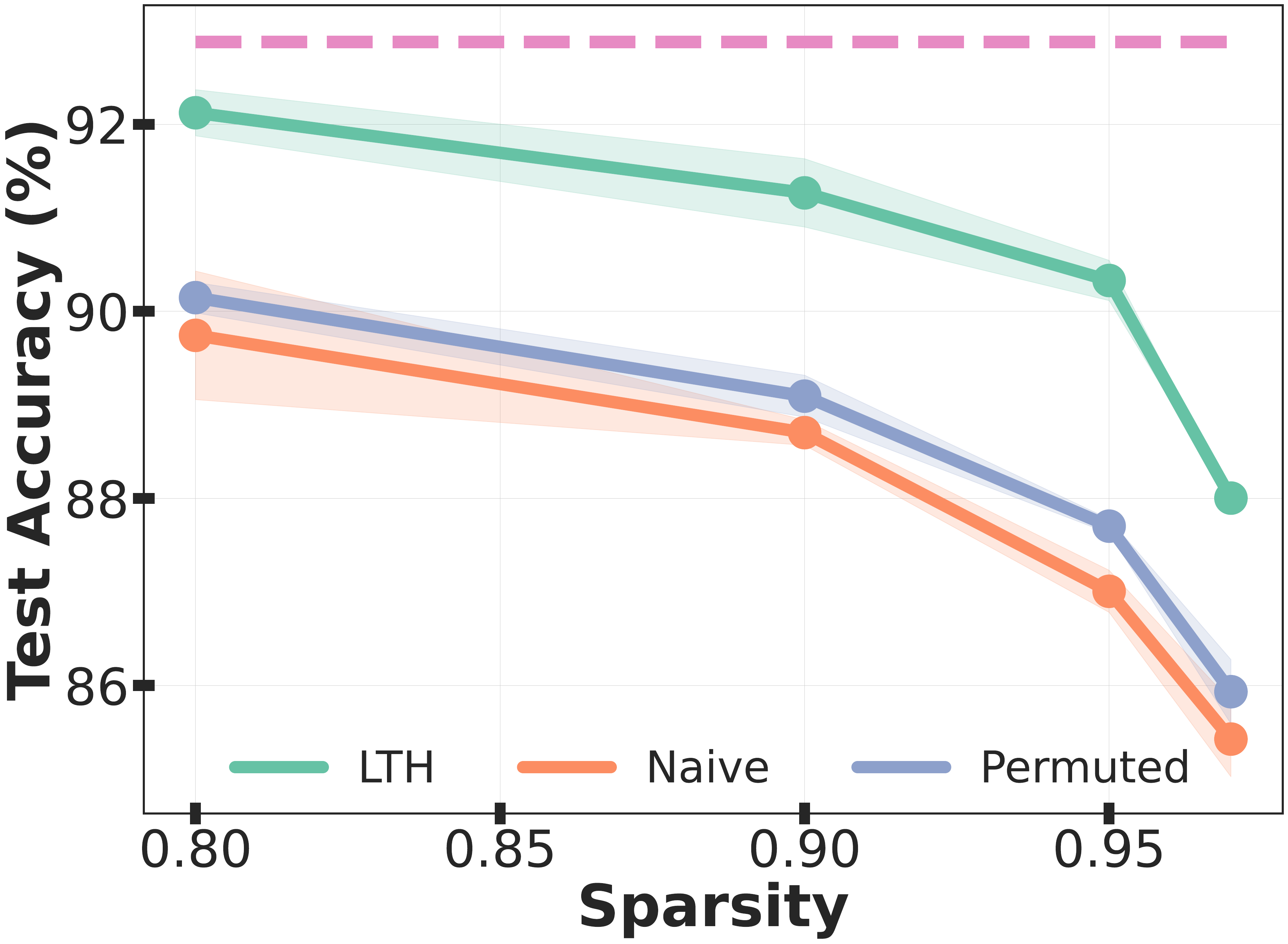}
        \caption{rewind=10}
        \label{fig:resnet_w1_s_fig:1}
    \end{subfigure}
    \begin{subfigure}{0.235\textwidth}
        \centering
        \includegraphics[width=\linewidth]{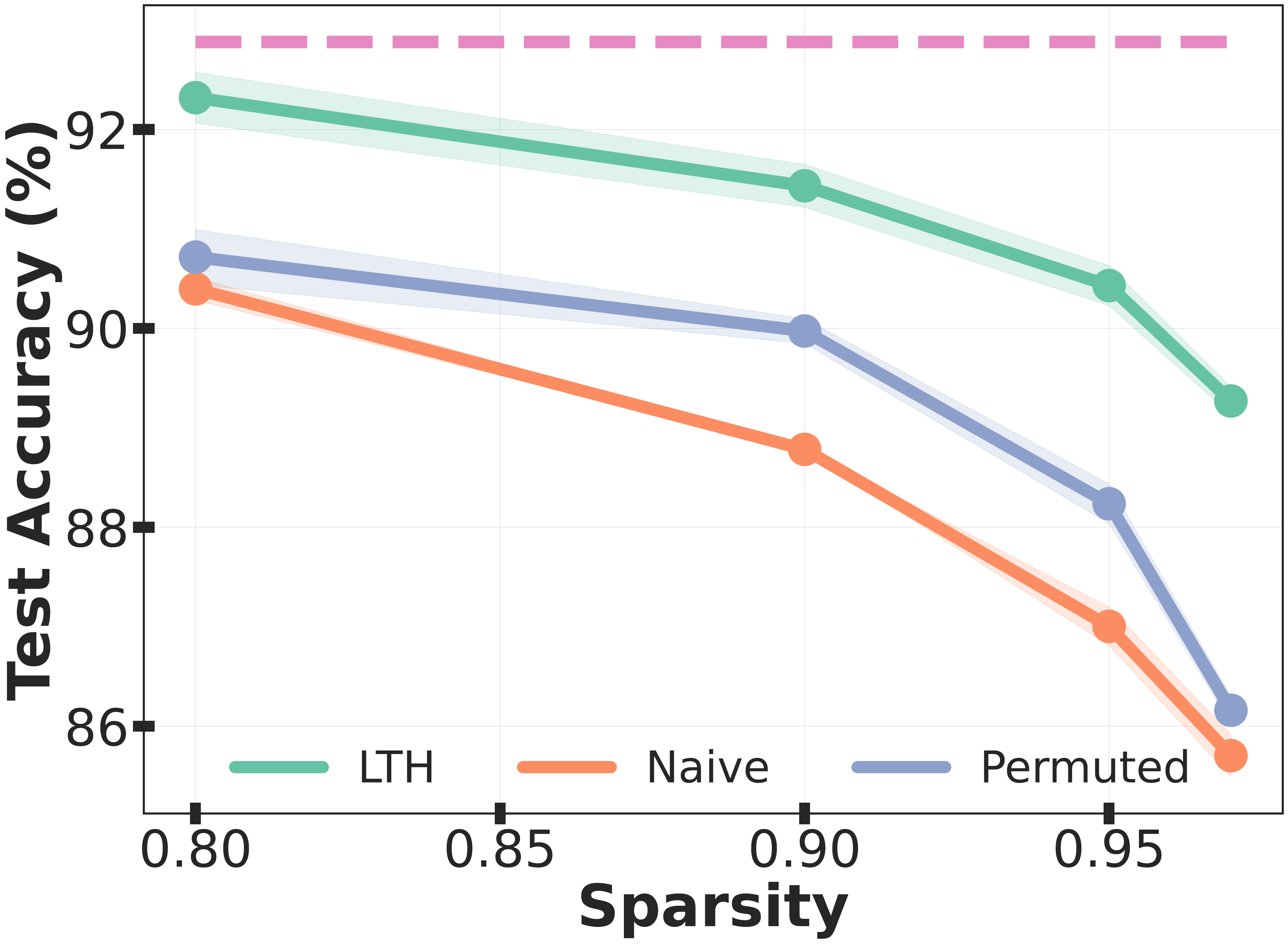} 
        \caption{rewind = 25}
        \label{fig:resnet_w1_s_fig:2}
    \end{subfigure}
    \begin{subfigure}{0.235\textwidth}
        \centering
        \includegraphics[width=\linewidth]{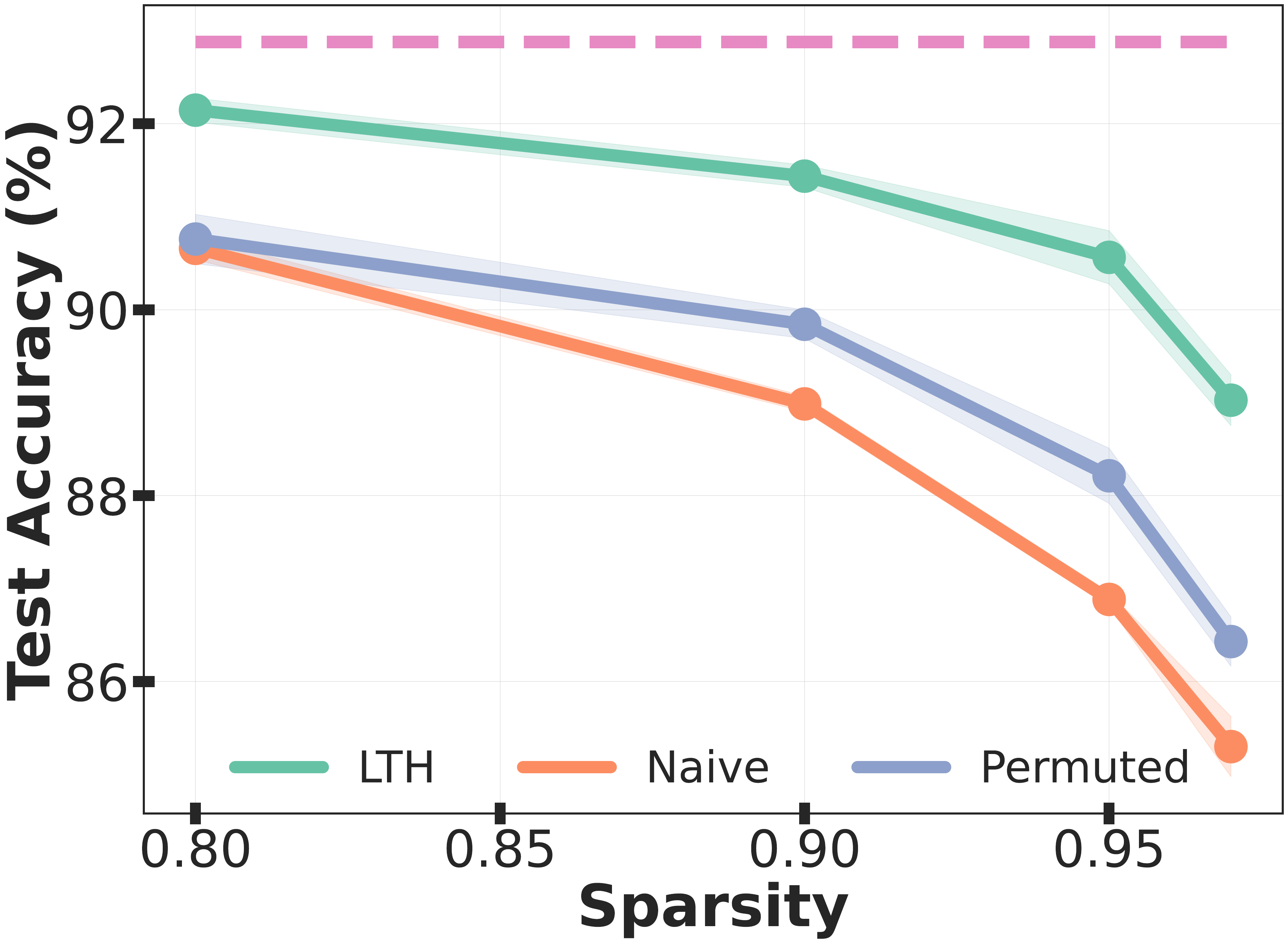} 
        \caption{rewind = 50}
        \label{fig:resnet_w1_s_fig:3}
    \end{subfigure}
    \begin{subfigure}{0.235\textwidth}
        \centering
        \includegraphics[width=\linewidth]{plots/acc_sp/rew_5_w_1.pdf} 
        \caption{rewind = 100}
        \label{fig:resnet_w1_s_fig:4}
    \end{subfigure}\\
    \begin{subfigure}{1.5em}
        \makebox[20pt]{\raisebox{50pt}{\rotatebox[origin=c]{90}{$w=4$}}}%
    \end{subfigure}
    \begin{subfigure}{0.235\textwidth}
        \centering
        \includegraphics[width=\linewidth]{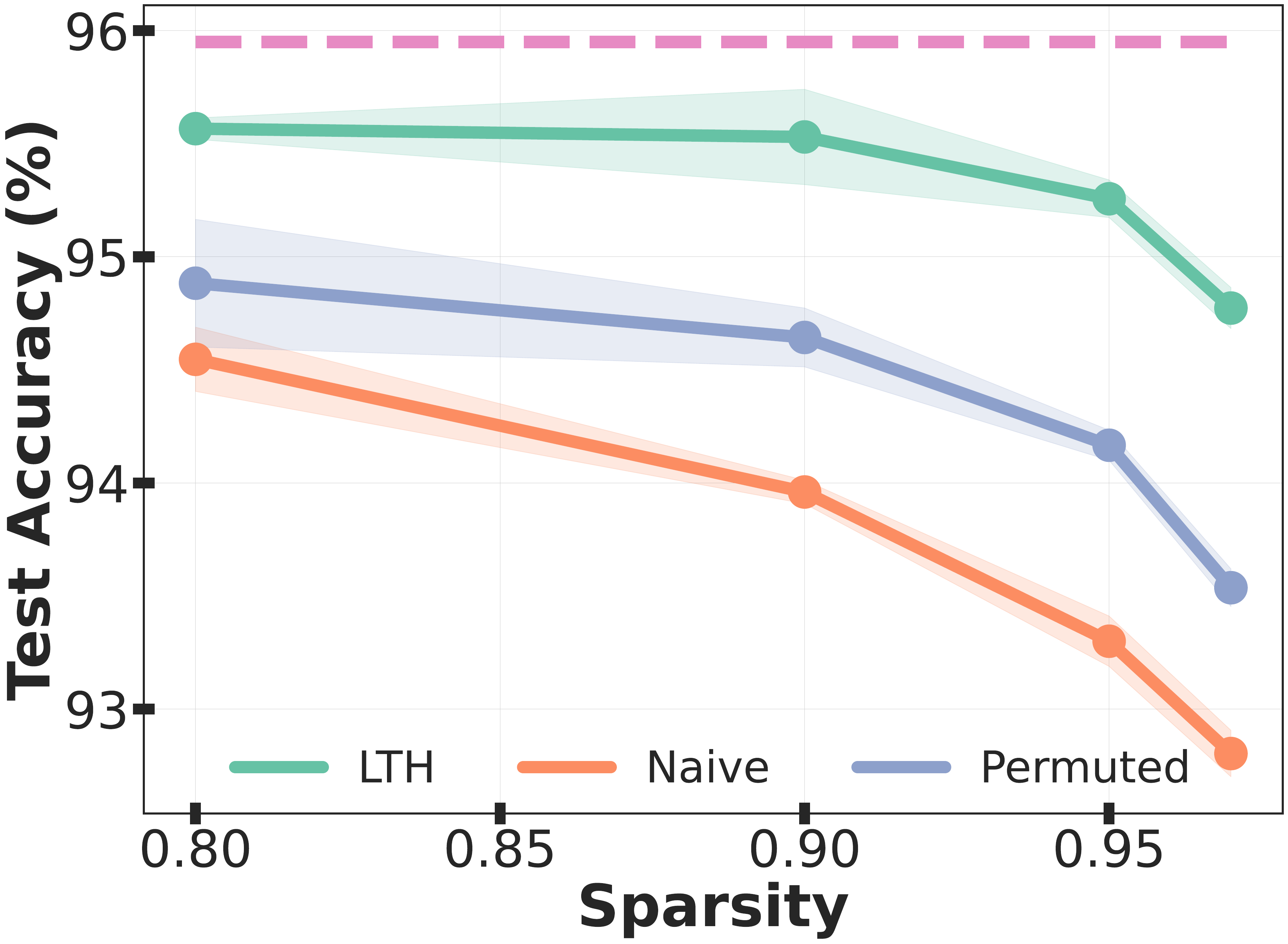}  
        \caption{rewind=10}
        \label{fig:resnet_w1_sp_fig:1}
    \end{subfigure}
    \begin{subfigure}{0.235\textwidth}
        \centering
        \includegraphics[width=\linewidth]{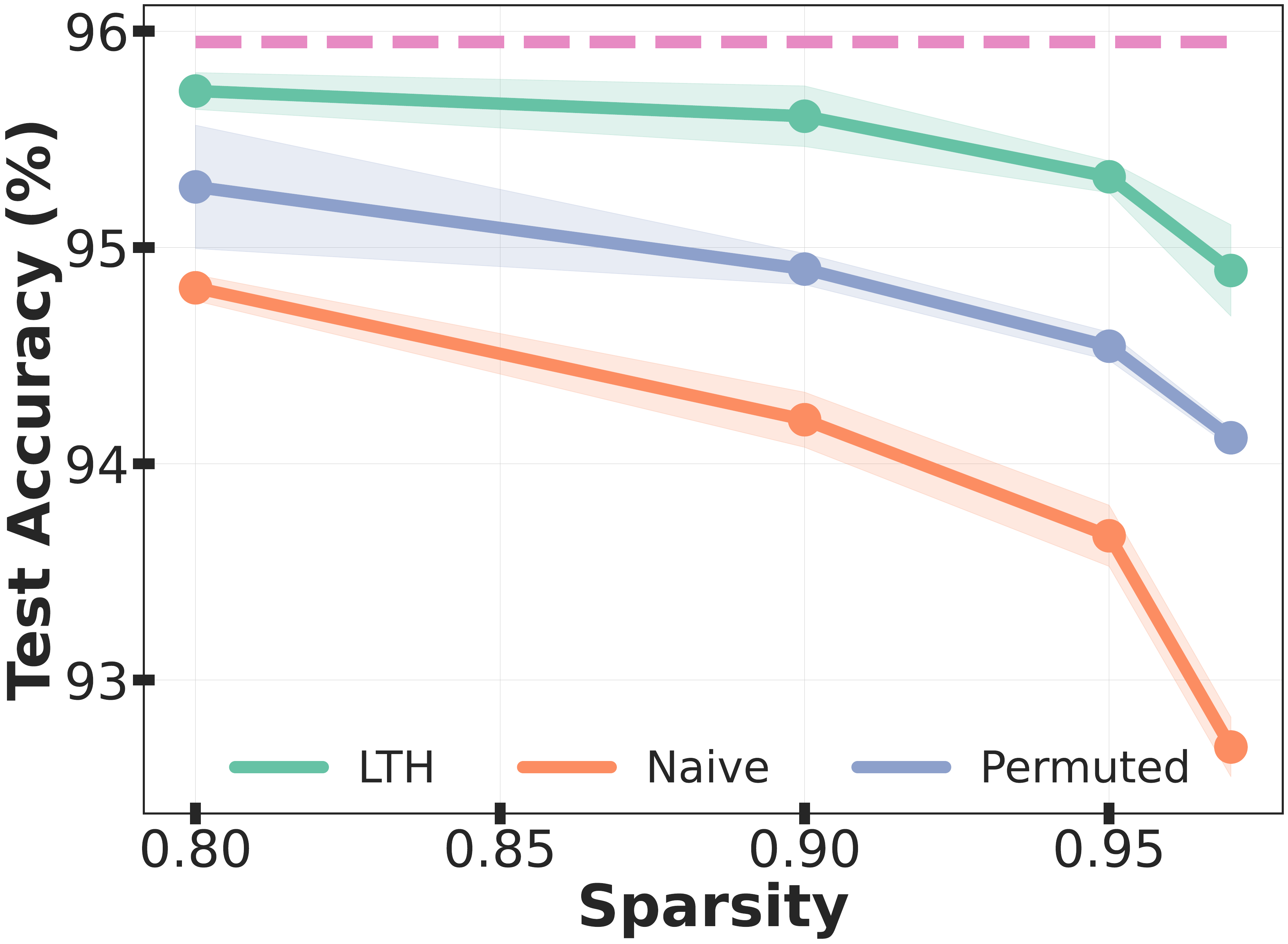}  
        \caption{rewind = 25}
        \label{fig:resnet_w1_sp_fig:2}
    \end{subfigure}
    \begin{subfigure}{0.235\textwidth}
        \centering
        \includegraphics[width=\linewidth]{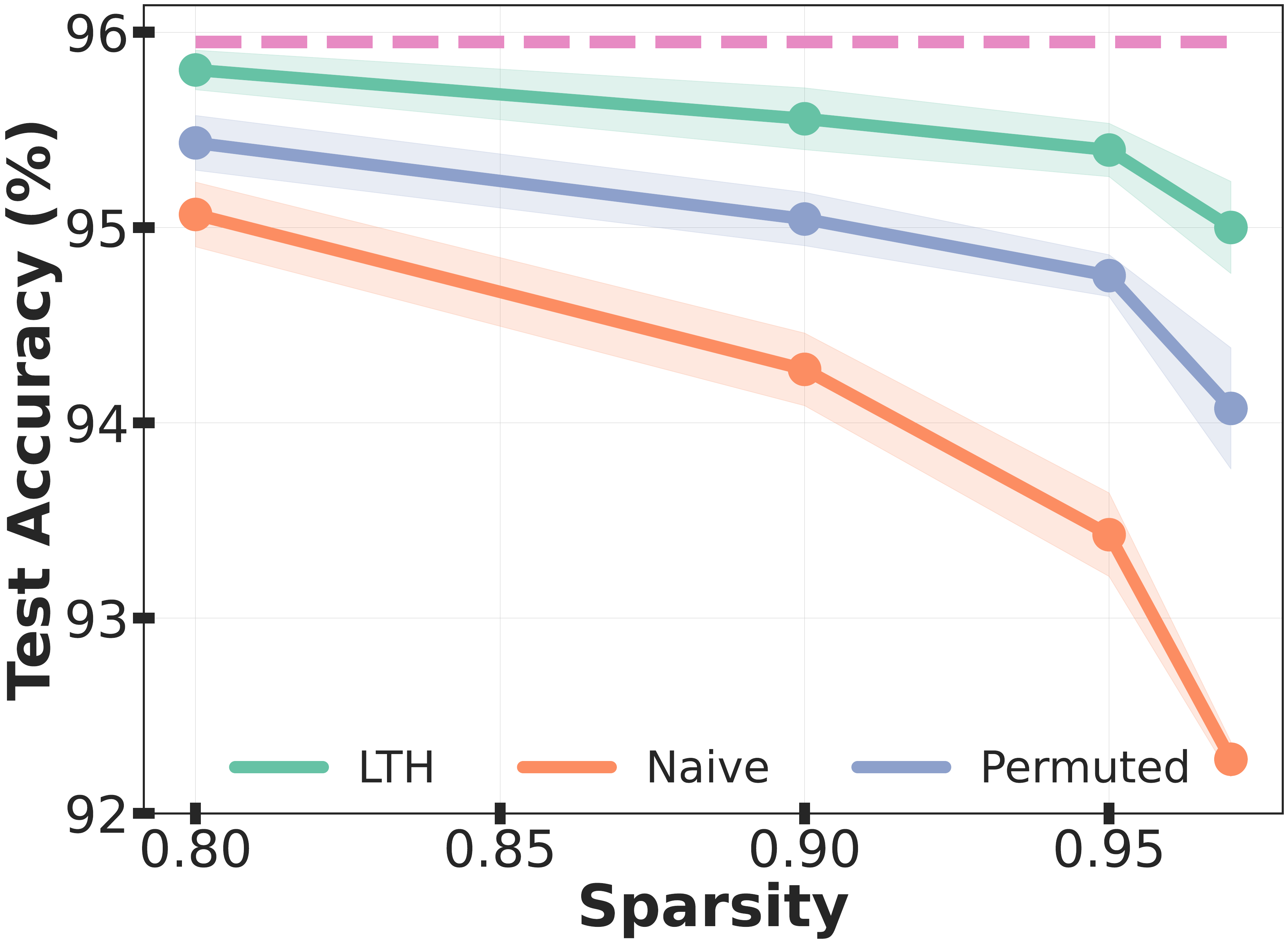}  
        \caption{rewind = 50}
        \label{fig:resnet_w1_sp_fig:3}
    \end{subfigure}
    \begin{subfigure}{0.235\textwidth}
        \centering
        \includegraphics[width=\linewidth]{plots/acc_sp/rew_5_w_4.pdf} 
        \caption{rewind = 100}
        \label{fig:resnet_w1_sp_fig:4}
    \end{subfigure}\\
    \begin{subfigure}{1.5em}
        \makebox[20pt]{\raisebox{50pt}{\rotatebox[origin=c]{90}{$w=8$}}}%
    \end{subfigure}
    \begin{subfigure}{0.235\textwidth}
        \centering
        \includegraphics[width=\linewidth]{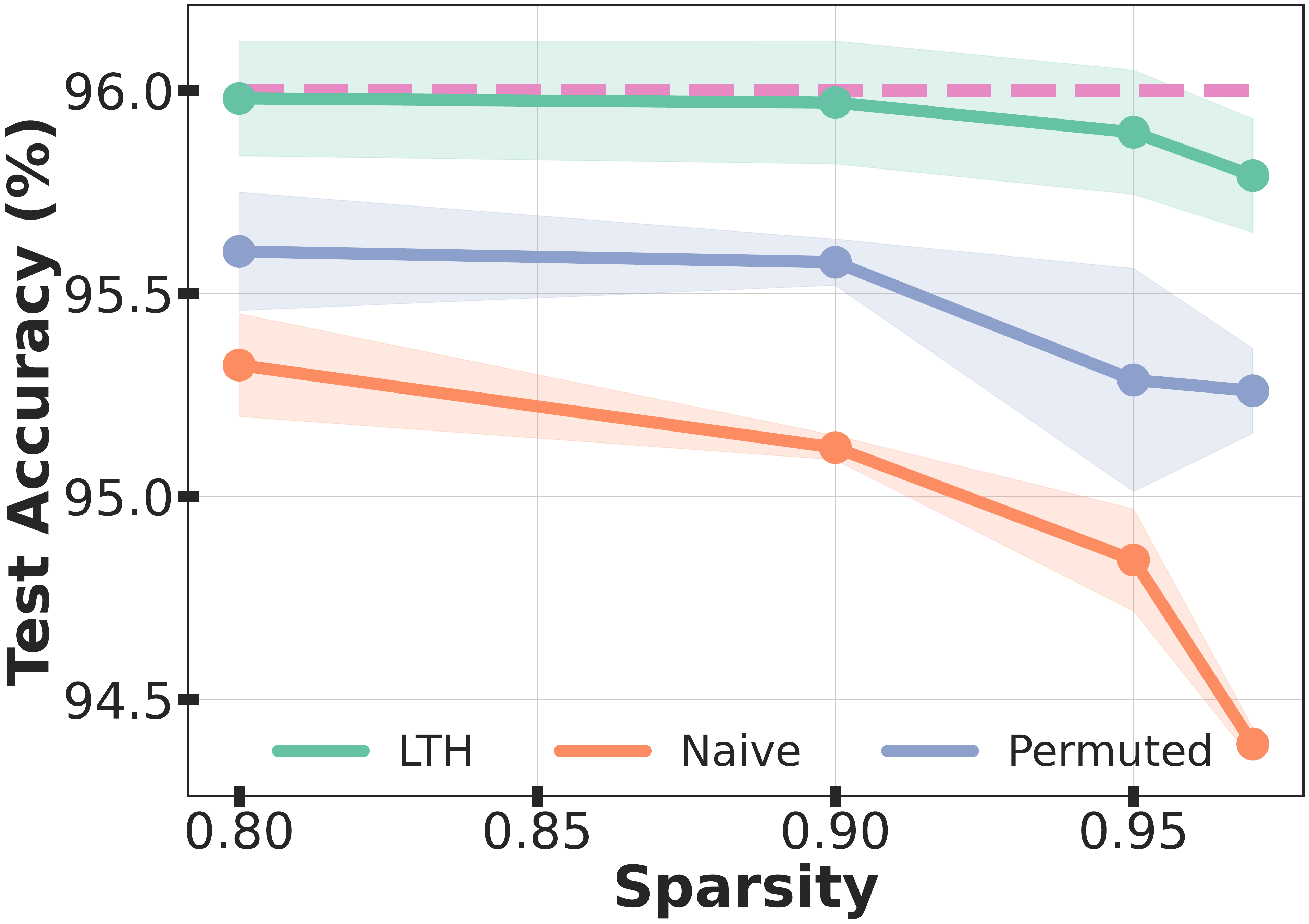}  
        \caption{rewind=10}
        \label{fig:rw_1_10:1}
    \end{subfigure}
    \begin{subfigure}{0.235\textwidth}
        \centering
        \includegraphics[width=\linewidth]{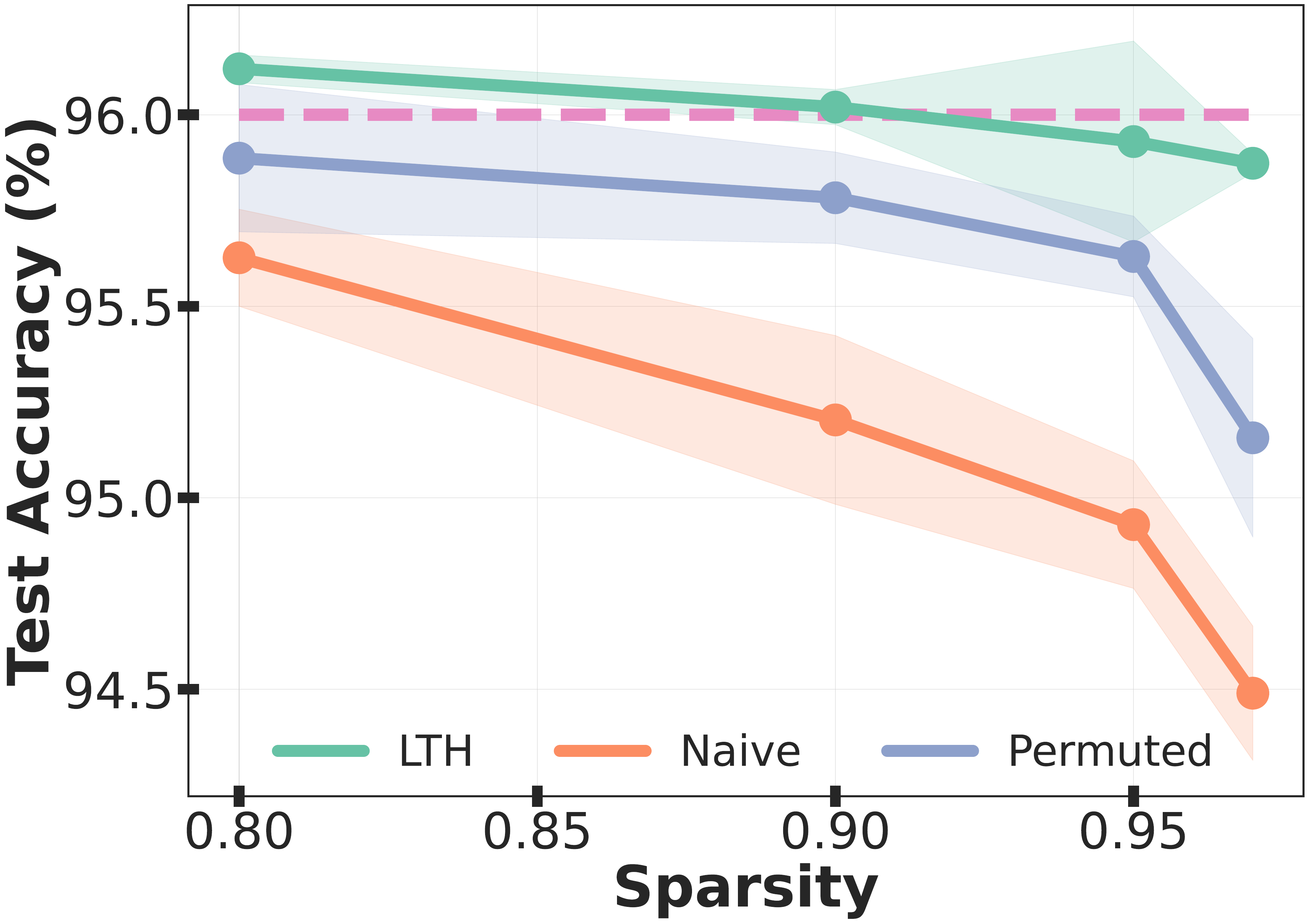}  
        \caption{rewind = 25}
        \label{fig:rw_1_25:2}
    \end{subfigure}
    \begin{subfigure}{0.235\textwidth}
        \centering
        \includegraphics[width=\linewidth]{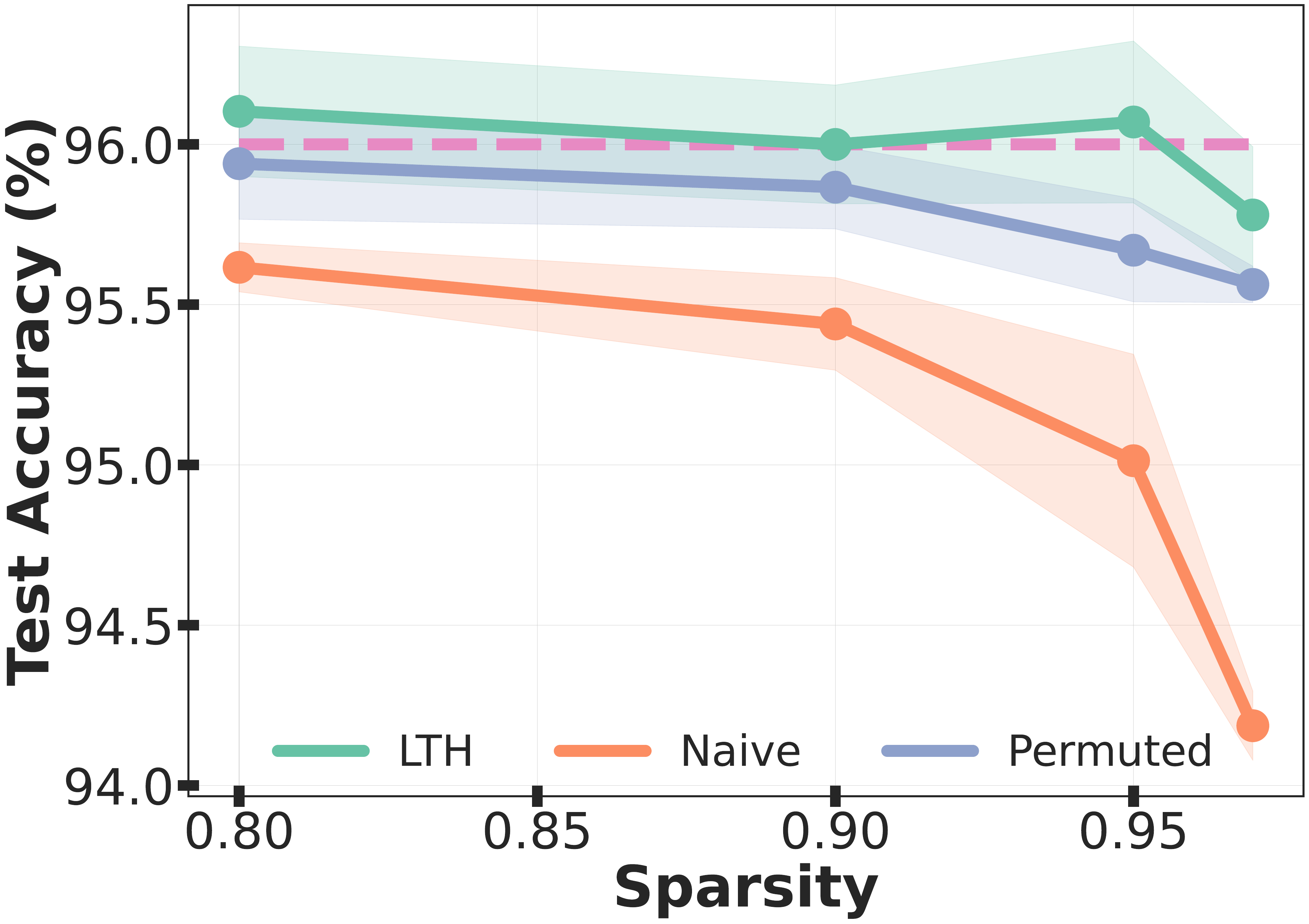}  
        \caption{rewind = 50}
        \label{fig:rw_1_50:3}
    \end{subfigure}
    \begin{subfigure}{0.235\textwidth}
        \centering
        \includegraphics[width=\linewidth]{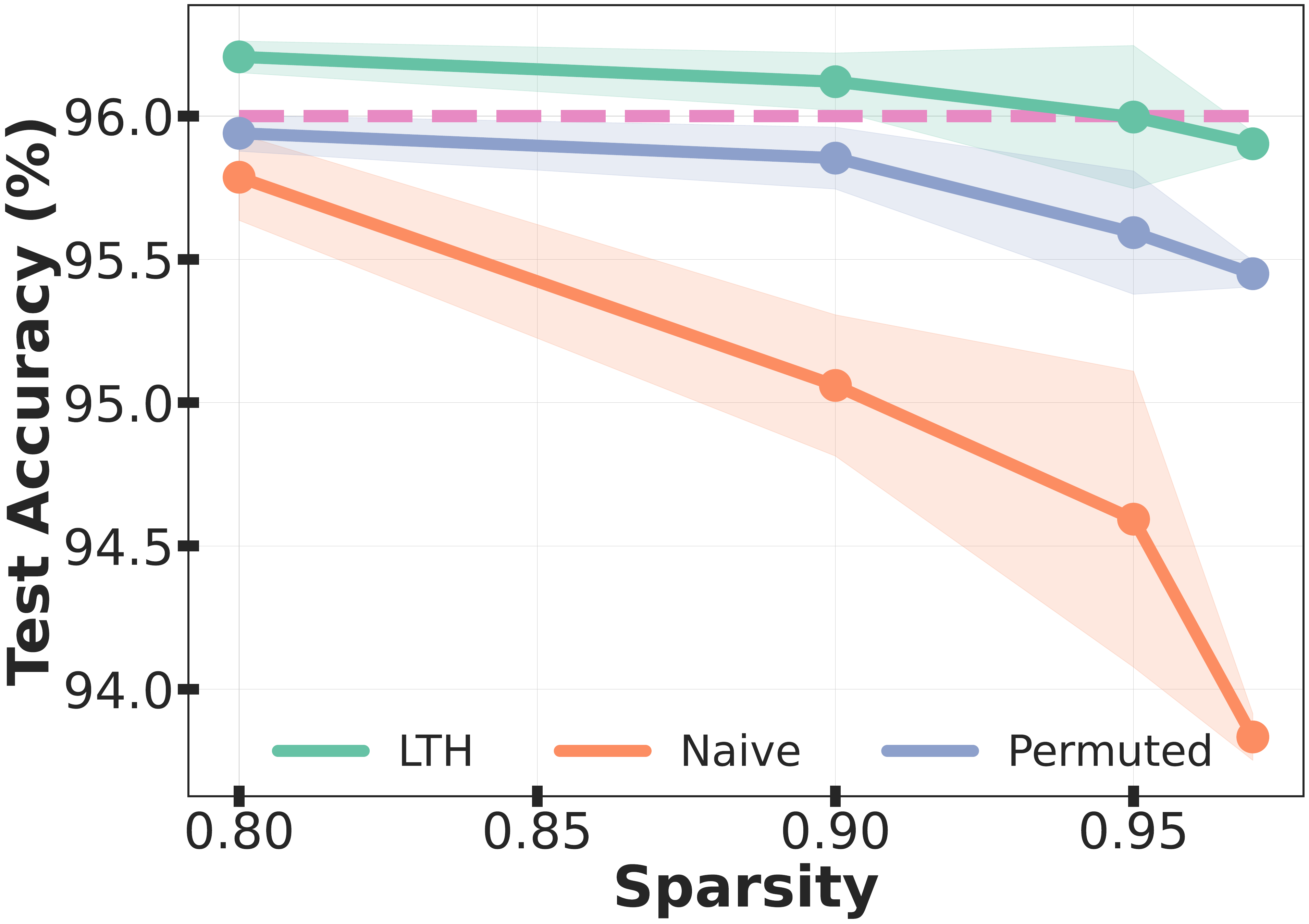} 
        \caption{rewind = 100}
        \label{fig:rw_1_100:4}
    \end{subfigure}\\
    \begin{subfigure}{1.5em}
        \makebox[20pt]{\raisebox{50pt}{\rotatebox[origin=c]{90}{$w=16$}}}%
    \end{subfigure}
    \begin{subfigure}{0.235\textwidth}
        \centering
        \includegraphics[width=\linewidth]{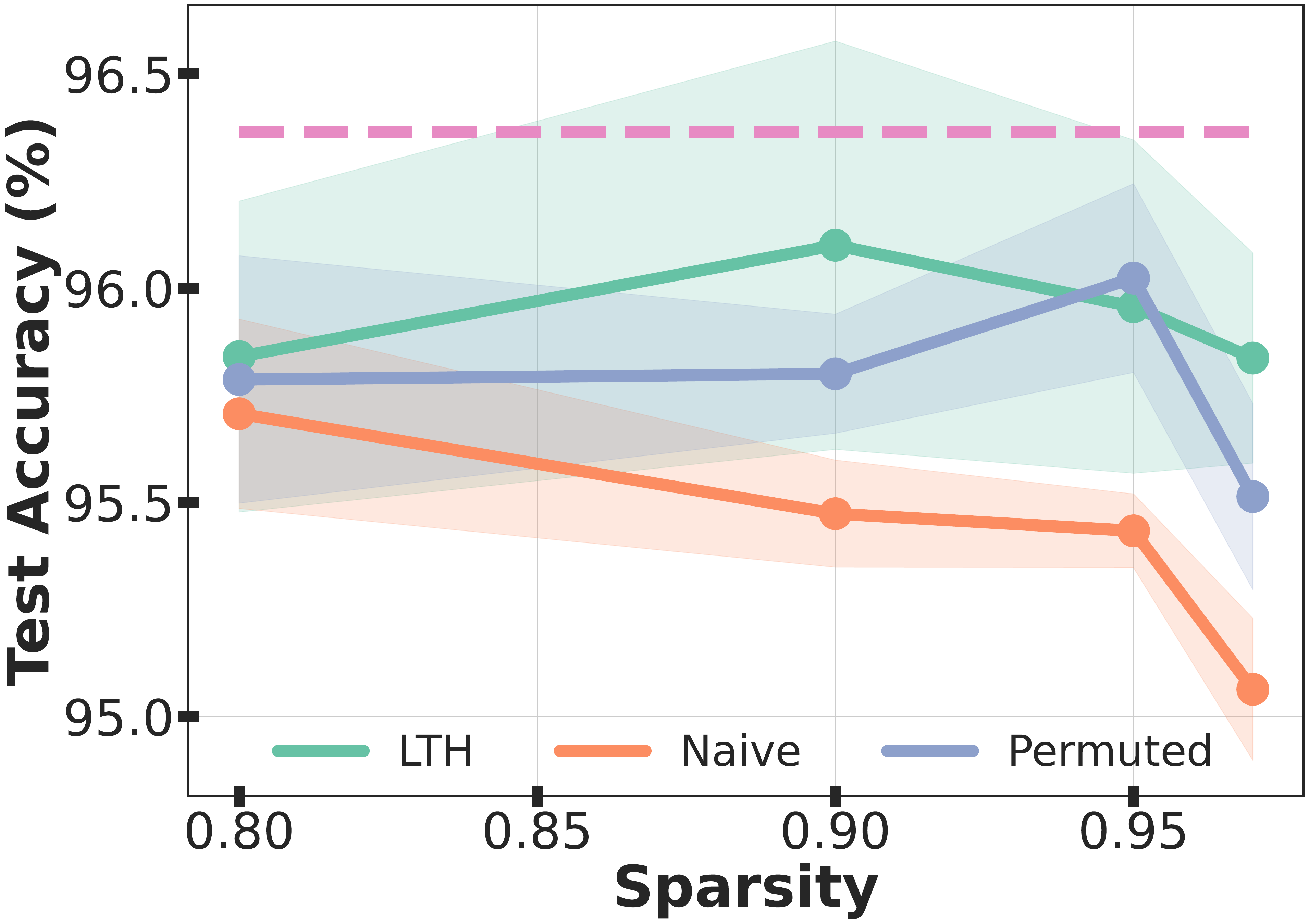}  
        \caption{rewind=10}
        \label{fig:rw_1_10_w16:1}
    \end{subfigure}
    \begin{subfigure}{0.235\textwidth}
        \centering
        \includegraphics[width=\linewidth]{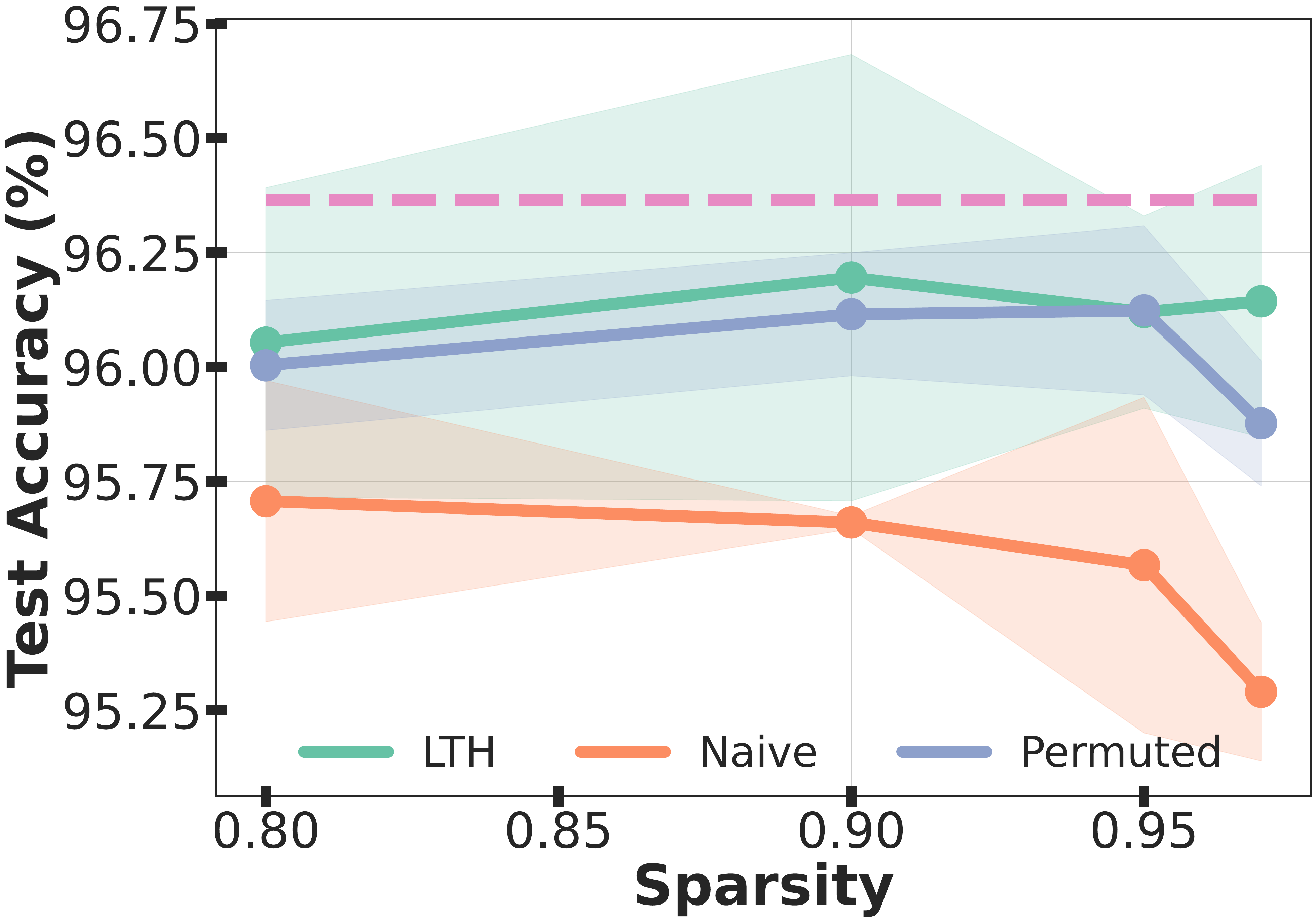}  
        \caption{rewind = 25}
        \label{fig:rw_1_25_w16:2}
    \end{subfigure}
    \begin{subfigure}{0.235\textwidth}
        \centering
        \includegraphics[width=\linewidth]{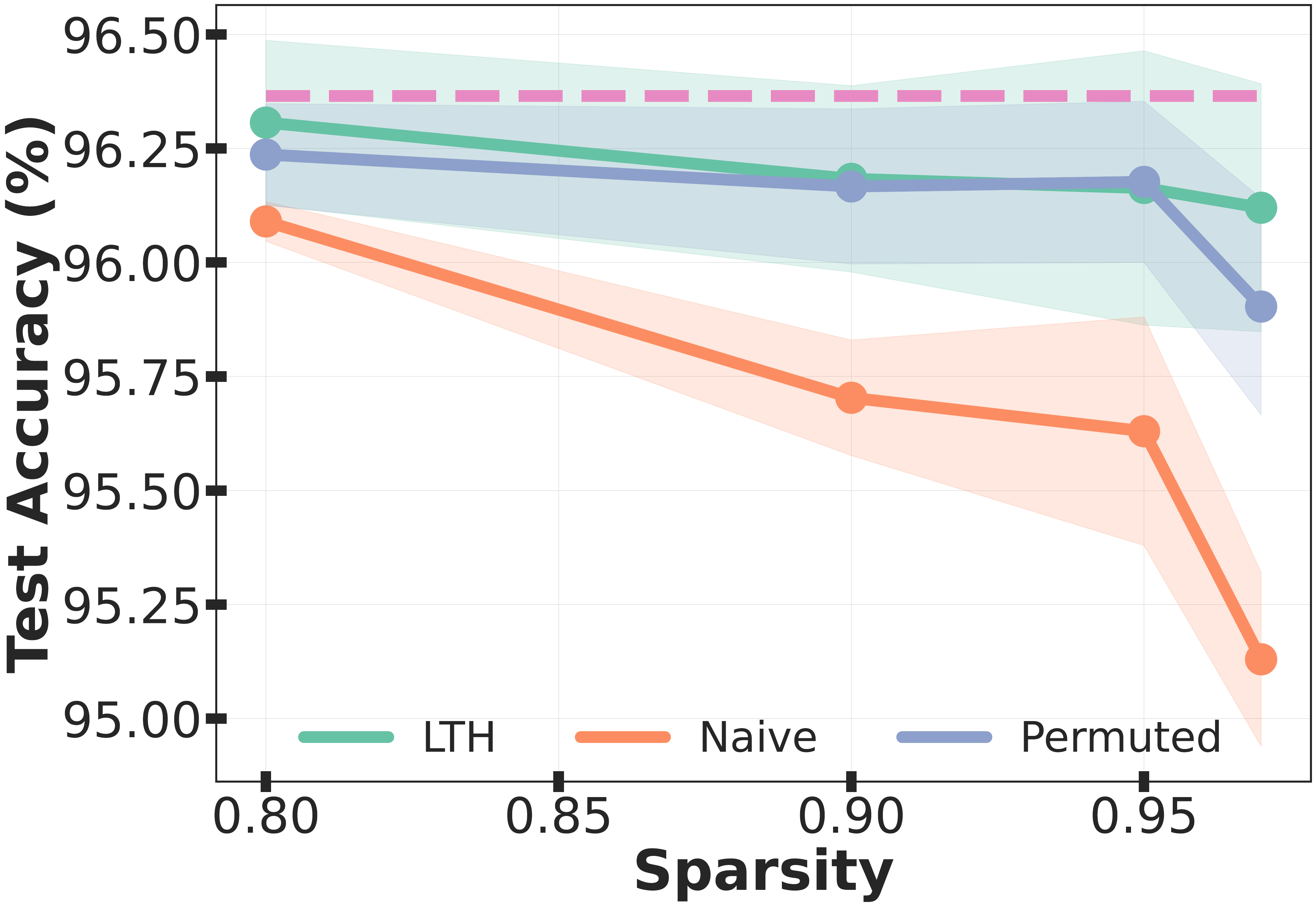}  
        \caption{rewind = 50}
        \label{fig:rw_1_50_w16:3}
    \end{subfigure}
    \begin{subfigure}{0.235\textwidth}
        \centering
        \includegraphics[width=\linewidth]{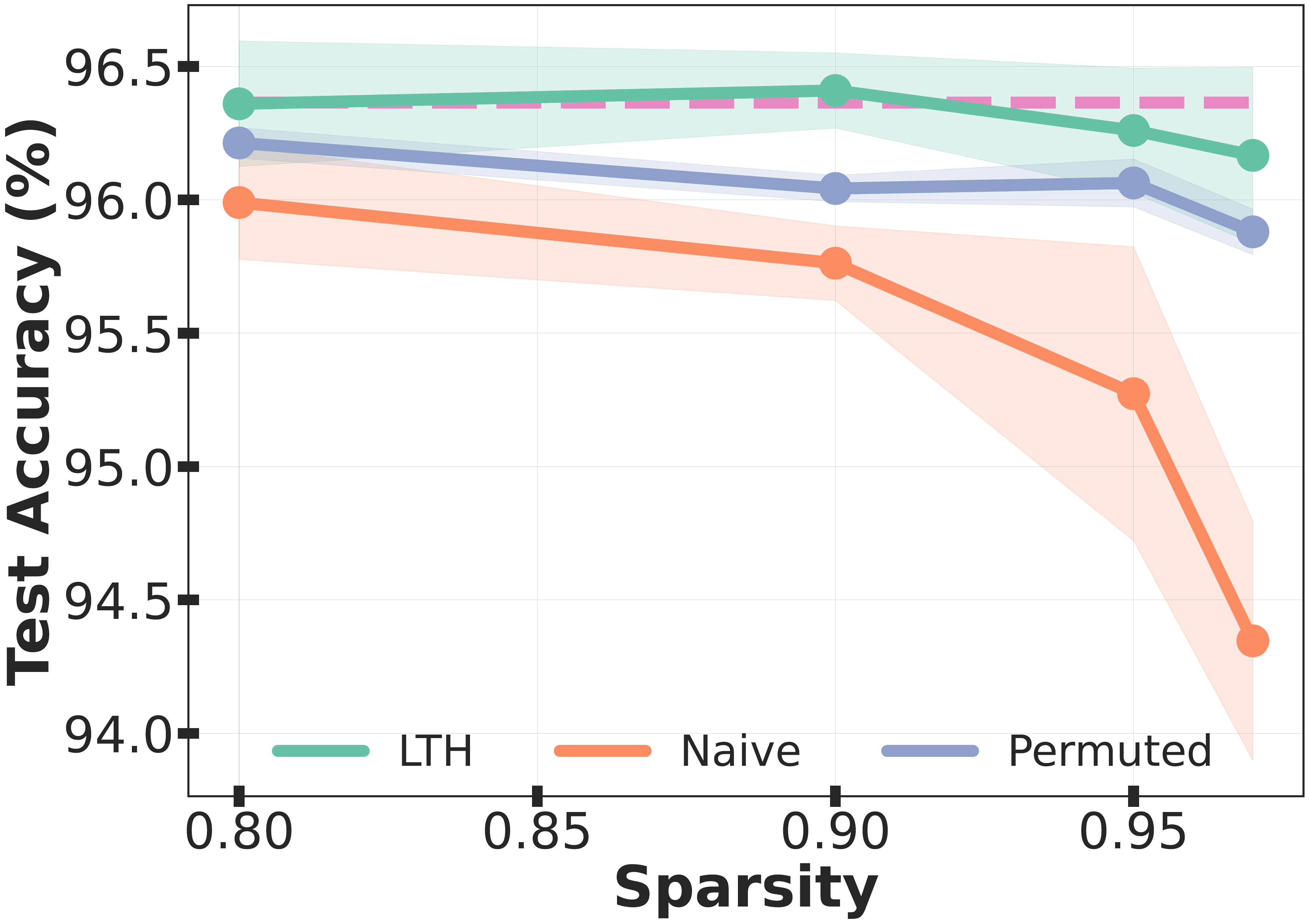} 
        \caption{rewind = 100}
        \label{fig:rw_1_100_w16:4}
    \end{subfigure}
    \caption{\small{\textbf{Accuracy vs sparsity trend for} \textbf{ResNet20$\times\{w\}$/CIFAR-10}.As the width increases, the gap between permuted and naive solutions increases, showing permuted masks help with sparse training. With increased width, we observe a more significant gap seen throughout ~\cref{fig:resnet_w1_s_fig:4,fig:resnet_w1_sp_fig:4,fig:rw_1_100:4,fig:rw_1_100_w16:4} and the permuted solution approaches the \gls{lth} solution. The dashed ({\textbf{- -}}) line shows the dense model accuracy.}}
    \label{fig:resnet_w1_sp}
\end{figure}

\begin{figure}[tbp]
    \centering
    % First row, first figure
    \begin{subfigure}{1.5em}
        \makebox[20pt]{\raisebox{50pt}{\rotatebox[origin=c]{90}{$w=1$}}}
    \end{subfigure}
    \begin{subfigure}{0.235\textwidth}
        \centering
        \includegraphics[width=\linewidth]{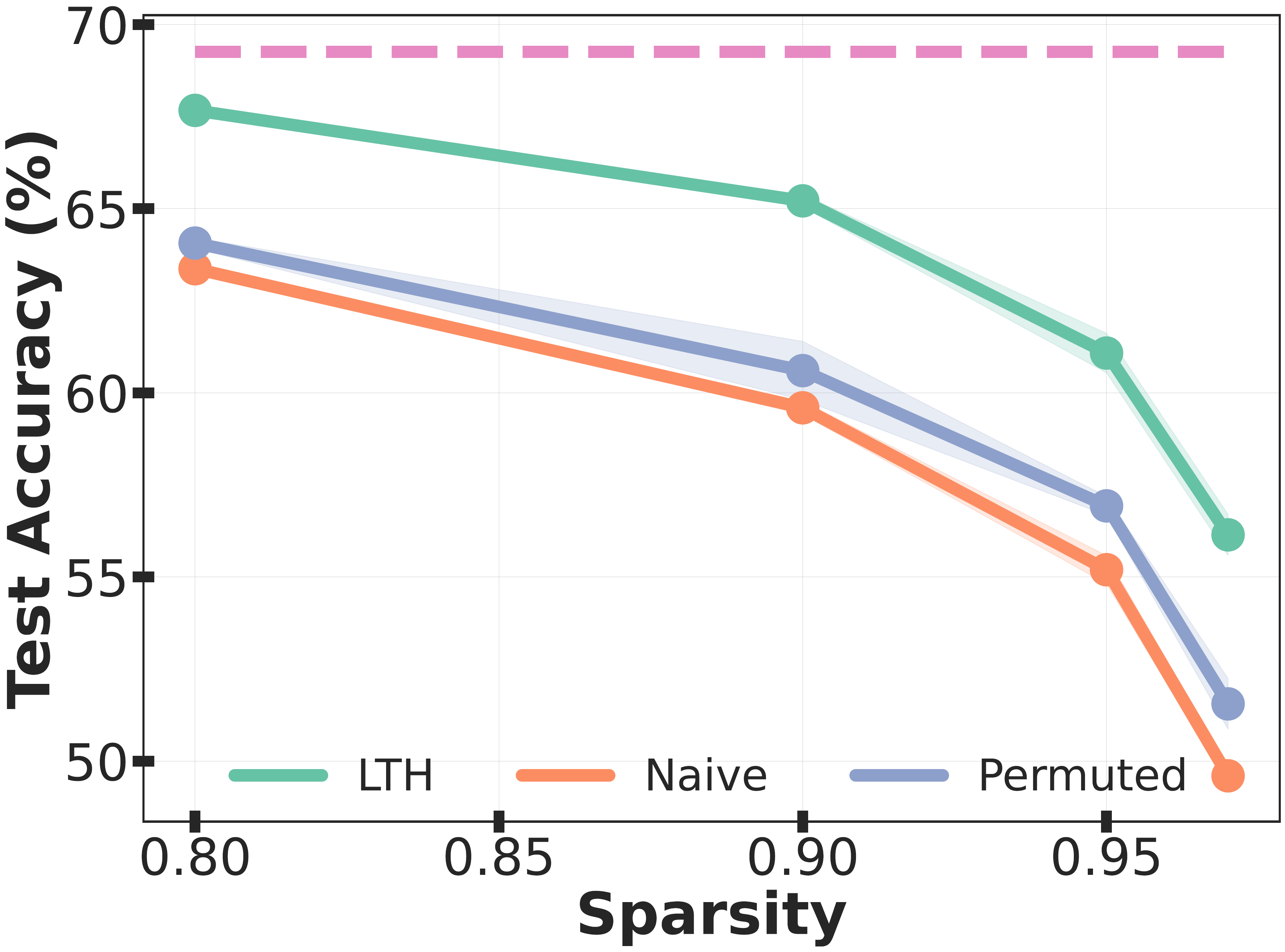}  % Replace with your image path
        \caption{rewind=10}
        \label{fig:accresnet_w1_sp_80_fig_rew10:1}
    \end{subfigure}
     % First row, second figure
    \begin{subfigure}{0.235\textwidth}
        \centering
        \includegraphics[width=\linewidth]{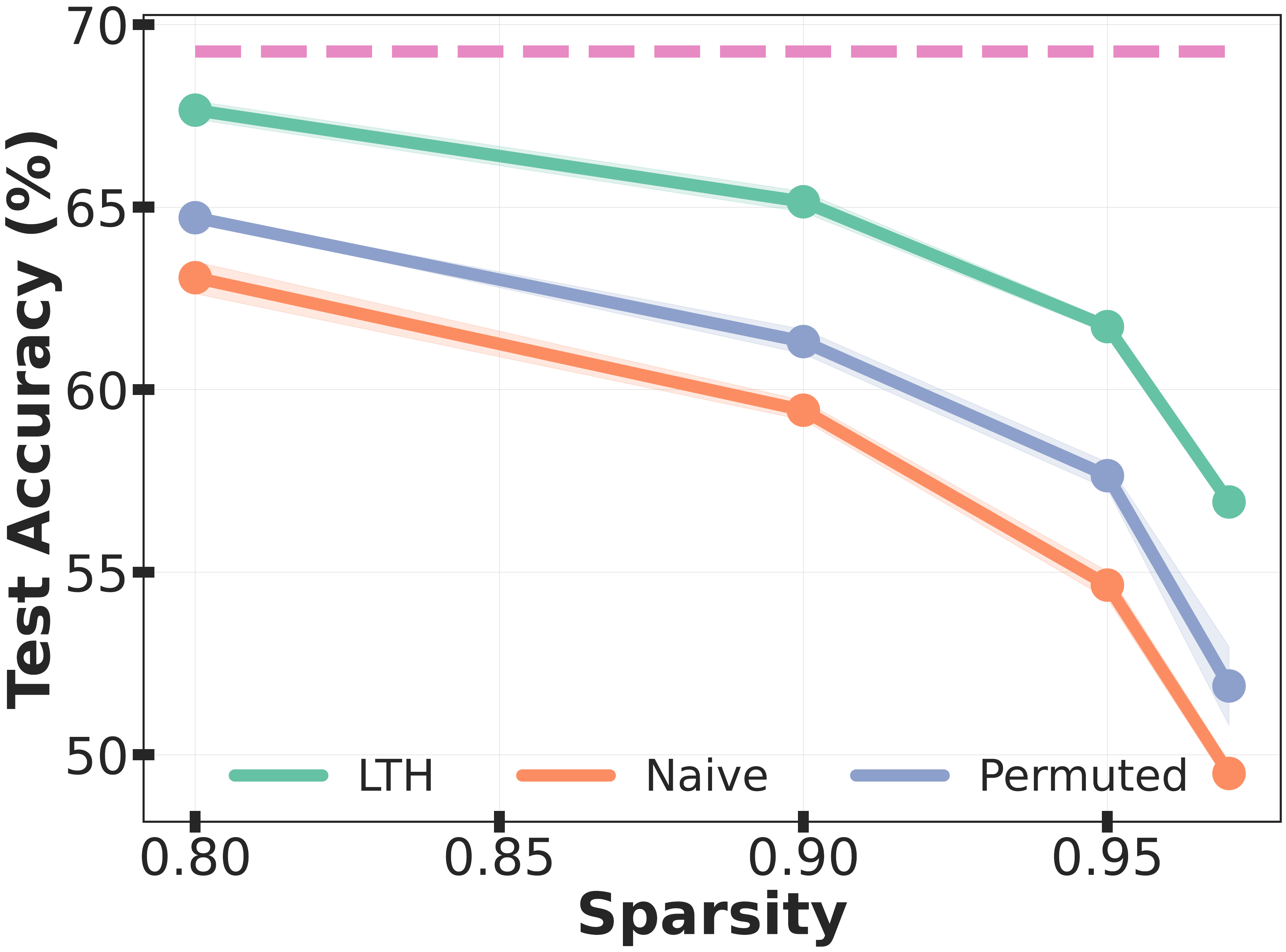}  % Replace with your image path
        \caption{rewind=25}
        \label{fig:accresnet_w1_sp_90_fig_rew25:2}
    \end{subfigure}
    \begin{subfigure}{0.235\textwidth}
        \centering
        \includegraphics[width=\linewidth]{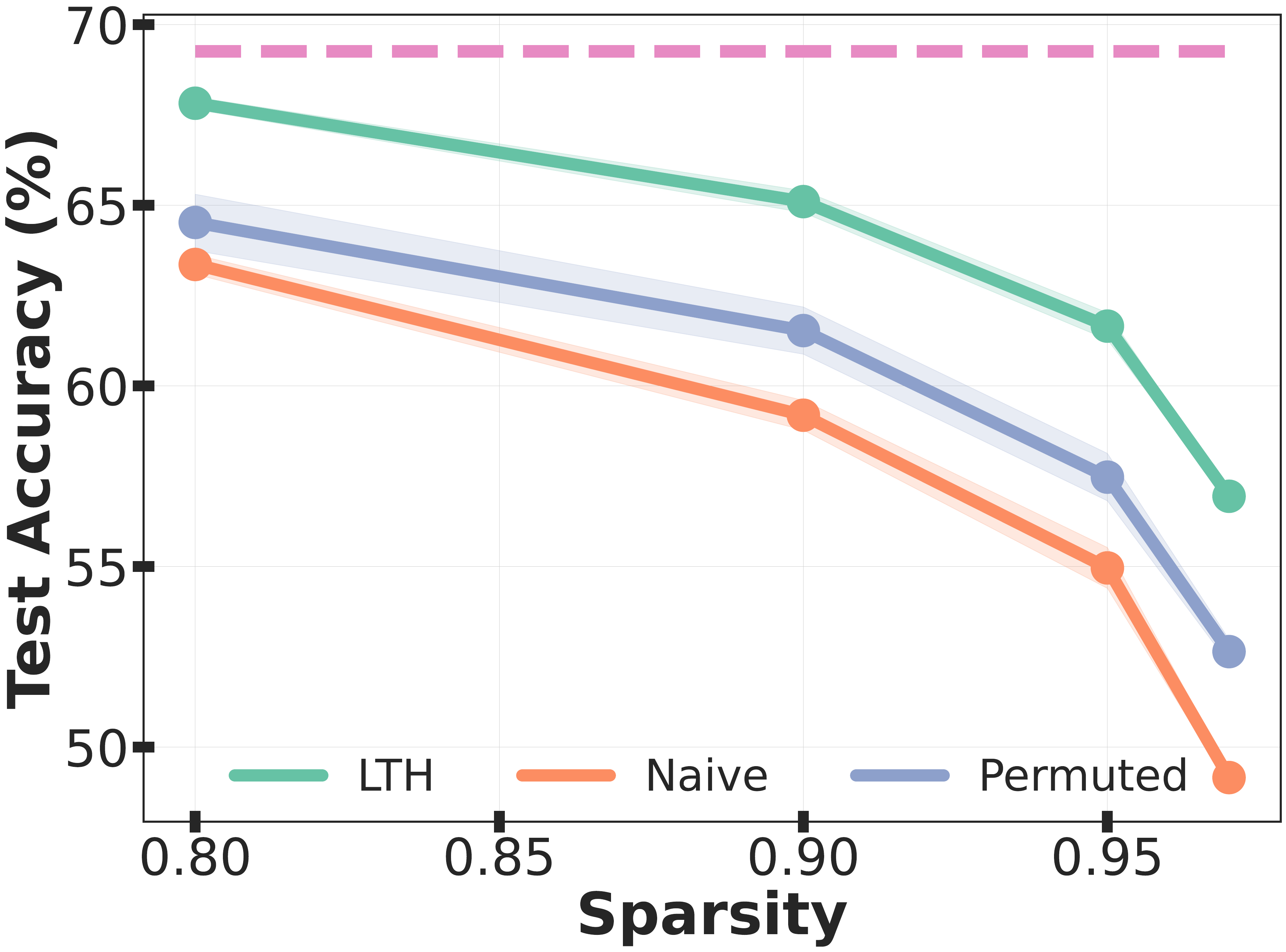}  % Replace with your image path
        \caption{rewind=50}
        \label{fig:accresnet_w1_sp_95_fig_rew50:2}
    \end{subfigure}
    % Second row, second figure
    \begin{subfigure}{0.235\textwidth}
        \centering
        \includegraphics[width=\linewidth]{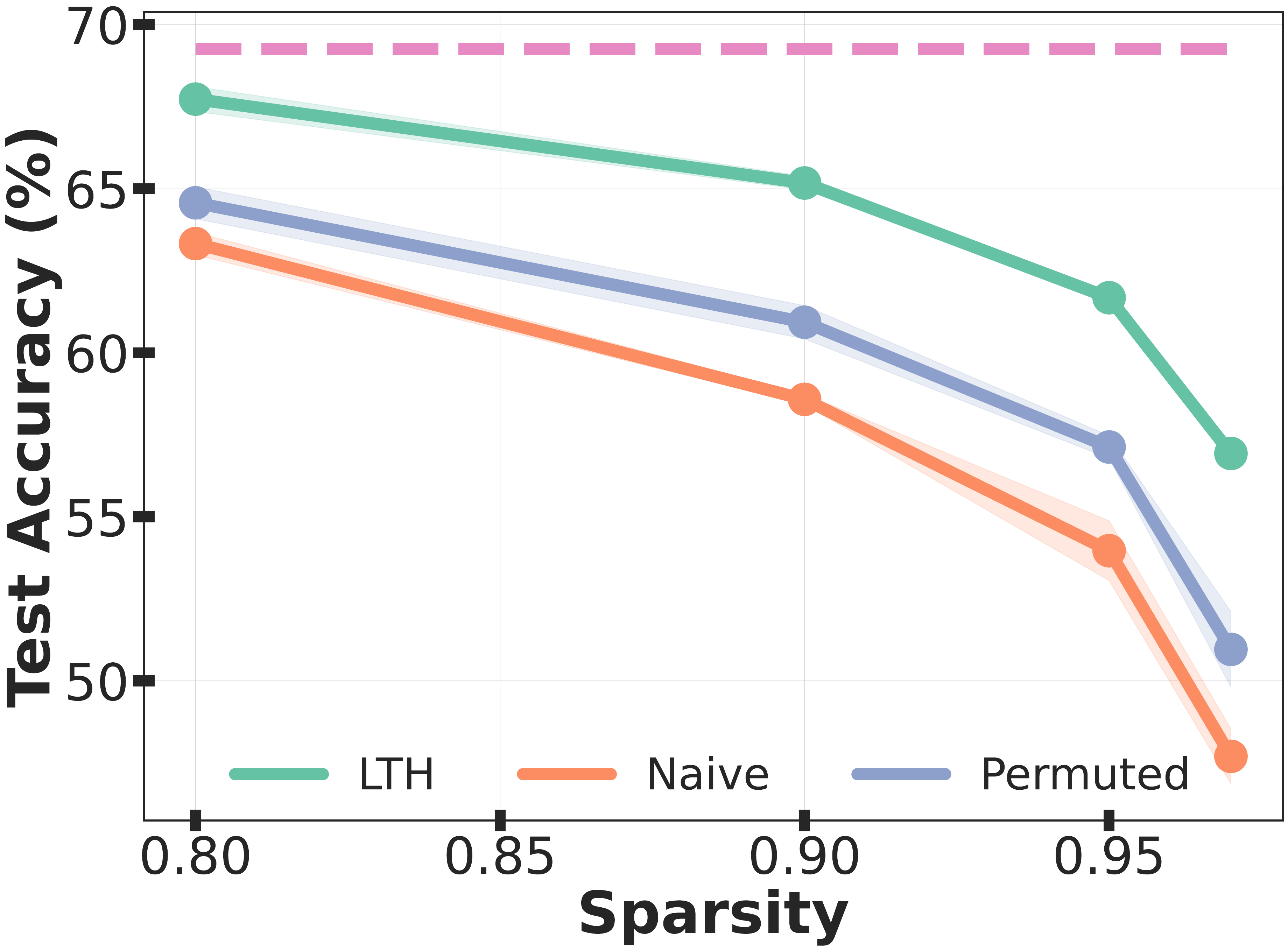}  % Replace with your image path
        \caption{rewind=100}
        \label{fig:accresnet_w1_sp_97_fig_rew100:4}
    \end{subfigure}\\%
    \begin{subfigure}{1.5em}
        \makebox[20pt]{\raisebox{50pt}{\rotatebox[origin=c]{90}{$w=4$}}}
    \end{subfigure}
    \begin{subfigure}{0.235\textwidth}
        \centering
        \includegraphics[width=\linewidth]{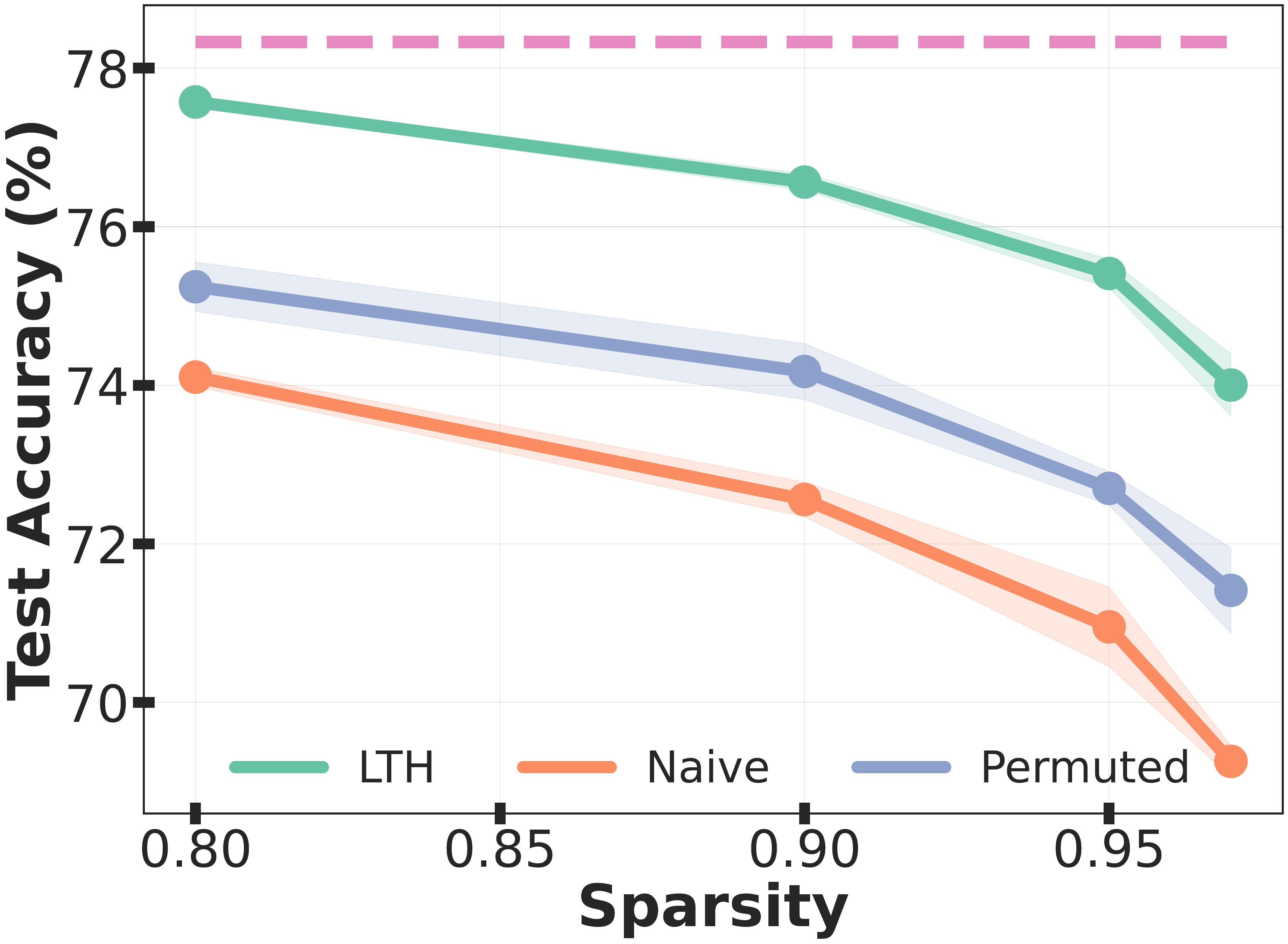}  
        \caption{rewind=10}
        \label{fig:resnet_rew10_w4:1}
    \end{subfigure}
    \begin{subfigure}{0.235\textwidth}
        \centering
        \includegraphics[width=\linewidth]{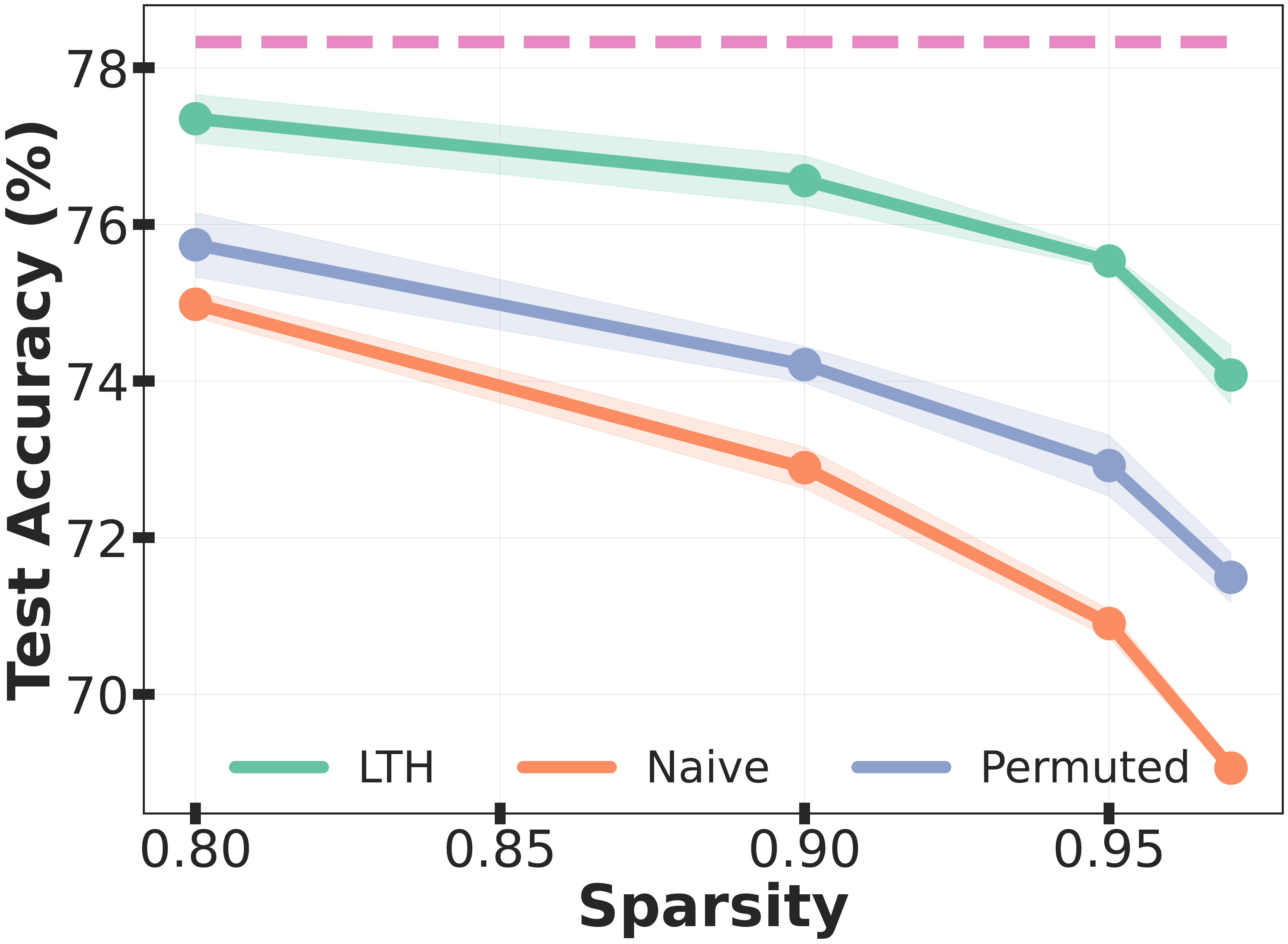}  
        \caption{rewind = 25}
        \label{fig:resnet_rew25_w4:2}
    \end{subfigure}
    \begin{subfigure}{0.235\textwidth}
        \centering
        \includegraphics[width=\linewidth]{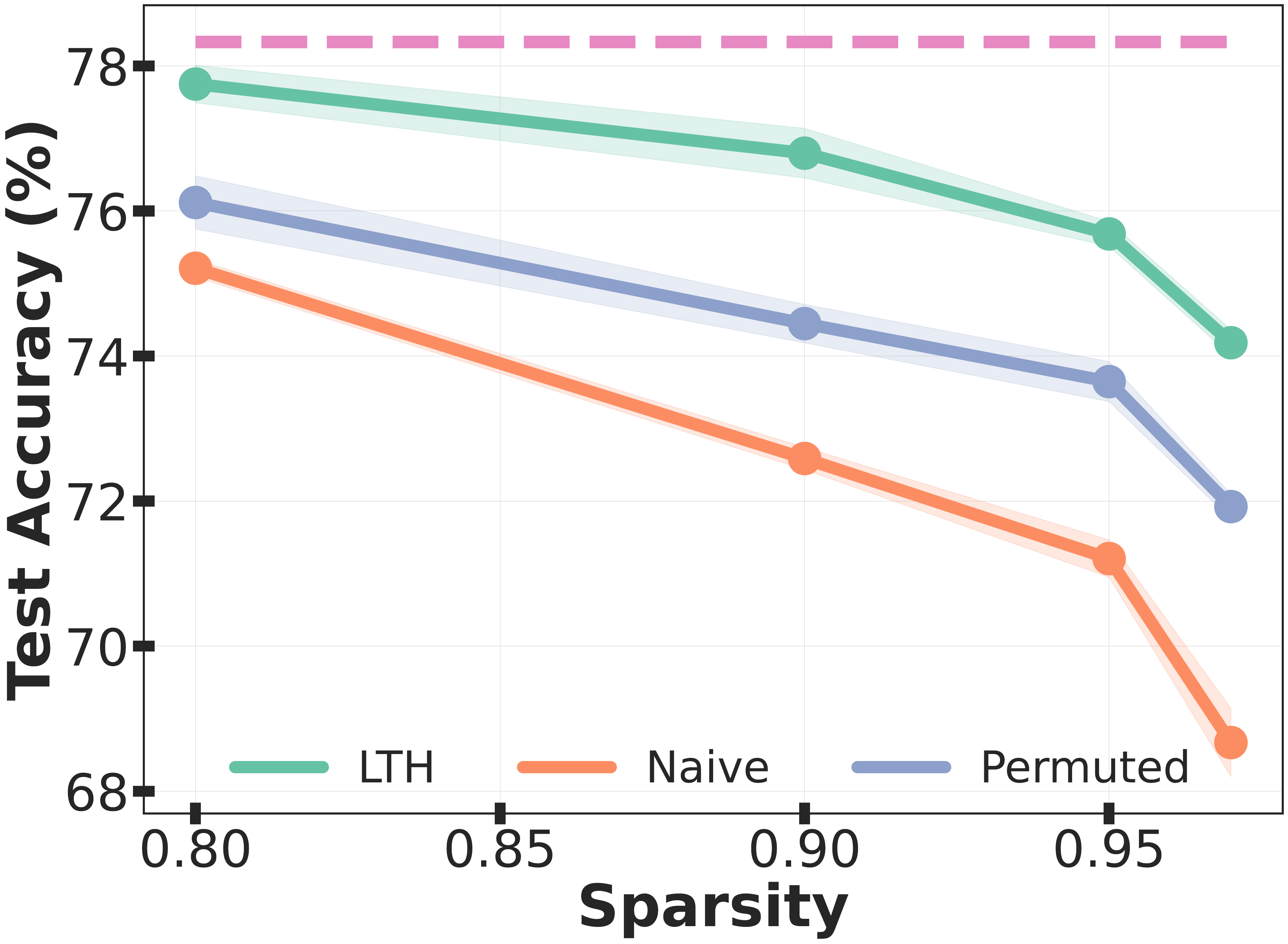}  
        \caption{rewind = 50}
        \label{fig:resnet_rew50_w4:3}
    \end{subfigure}
    \begin{subfigure}{0.235\textwidth}
        \centering
        \includegraphics[width=\linewidth]{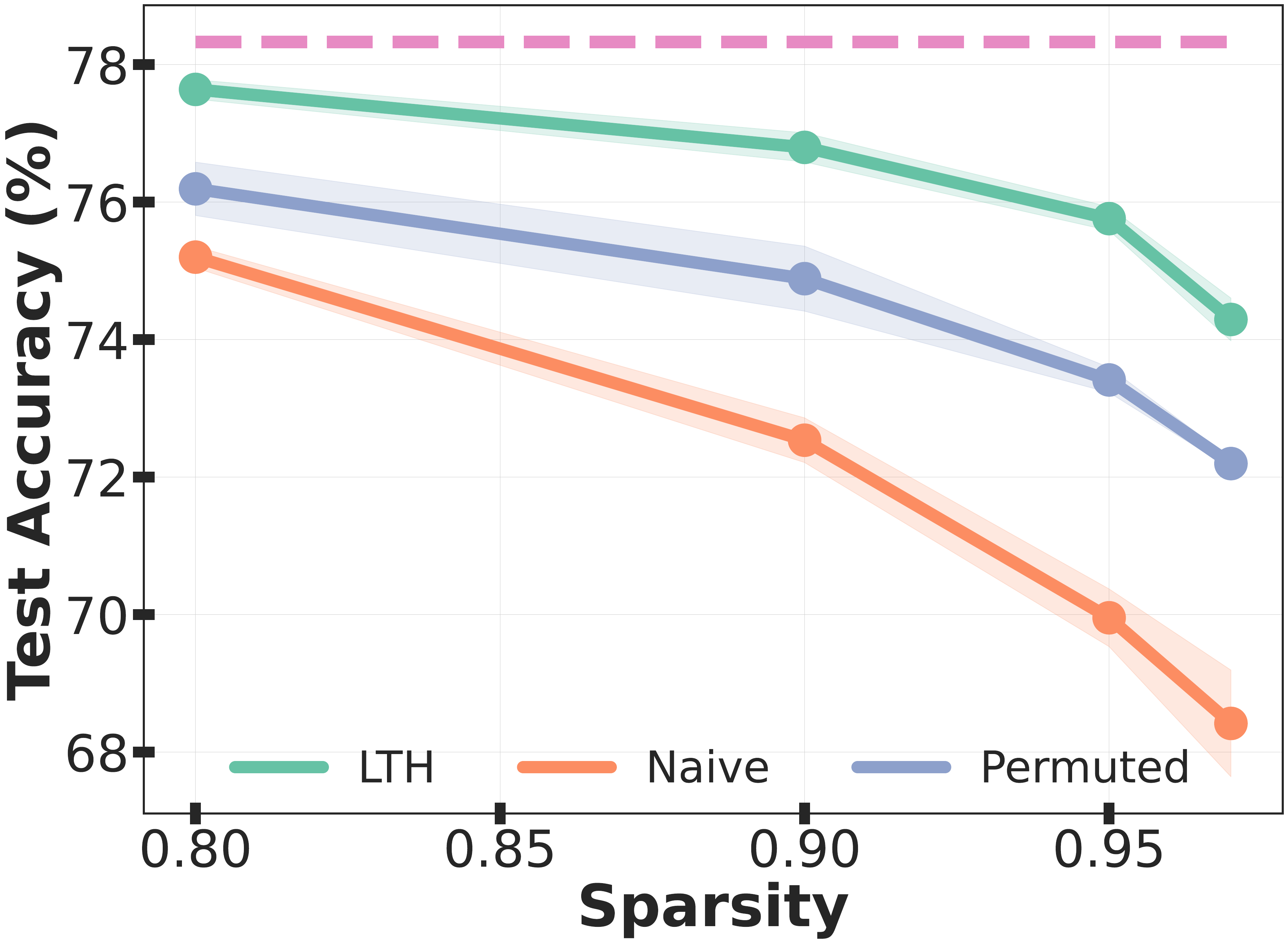} 
        \caption{rewind = 100}
        \label{fig:resnet_rew100_w4:4}
    \end{subfigure}\\
    \centering
    \begin{subfigure}{1.5em}
        \makebox[20pt]{\raisebox{50pt}{\rotatebox[origin=c]{90}{$w=8$}}}
    \end{subfigure}
    \begin{subfigure}{0.235\textwidth}
        \centering
        \includegraphics[width=\linewidth]{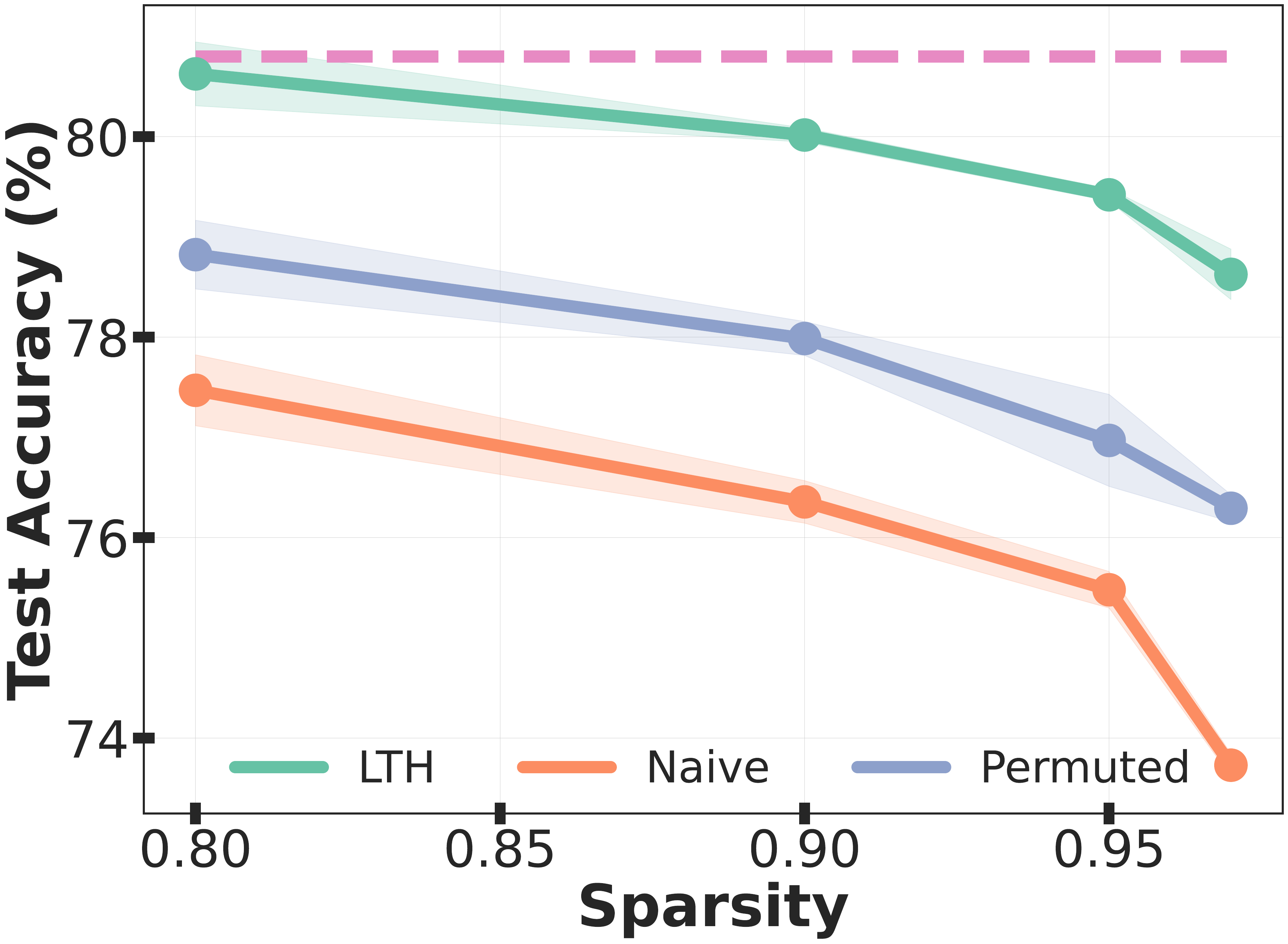}
        \caption{rewind=10}
        \label{fig:resnet_rew10_w8:1}
    \end{subfigure}
    \centering
    \begin{subfigure}{0.235\textwidth}
        \centering
        \includegraphics[width=\linewidth]{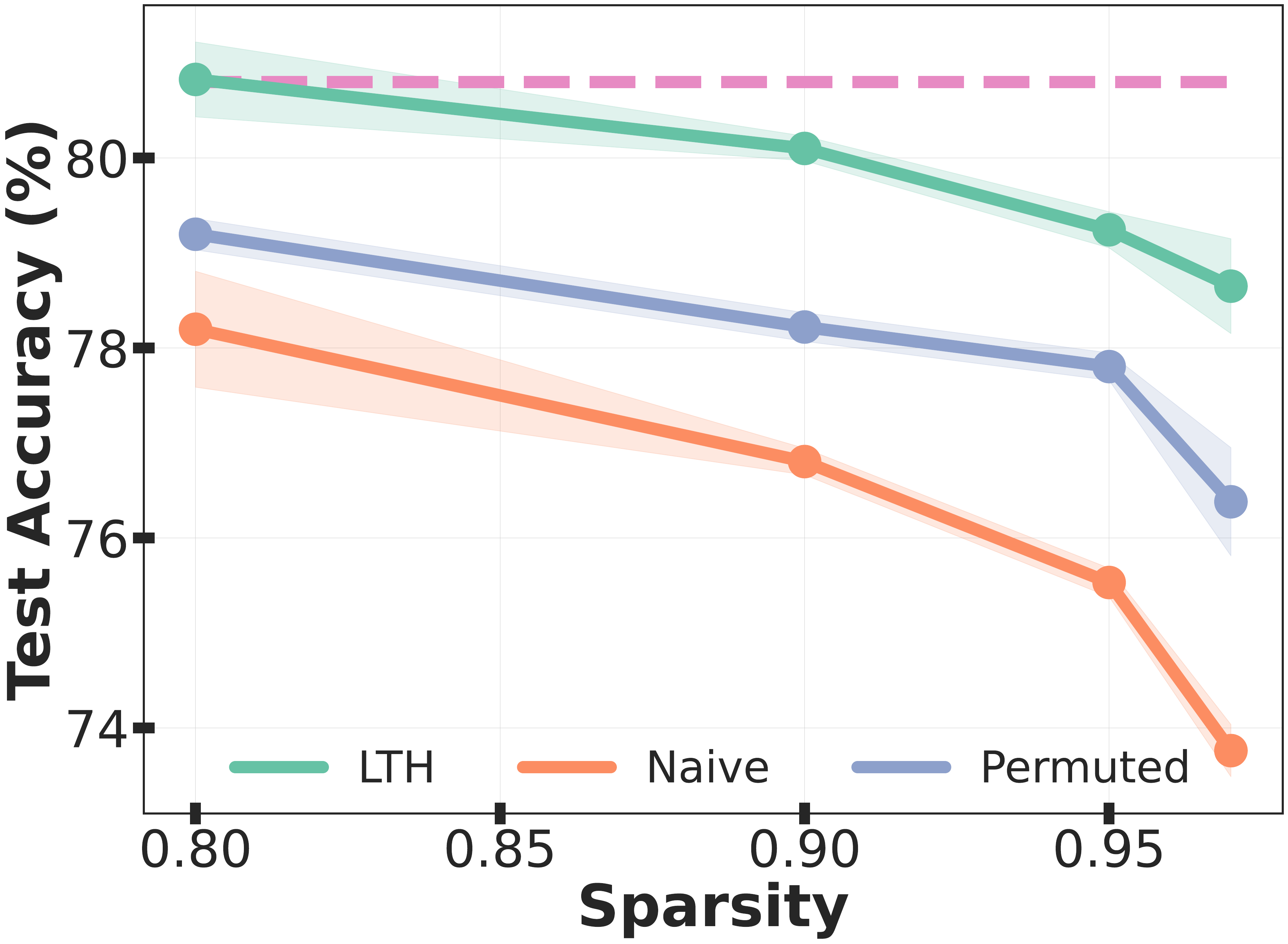}  
        \caption{rewind = 25}
        \label{fig:resnet_rew25_w8:2}
    \end{subfigure}
    \centering
    \begin{subfigure}{0.235\textwidth}
        \centering
        \includegraphics[width=\linewidth]{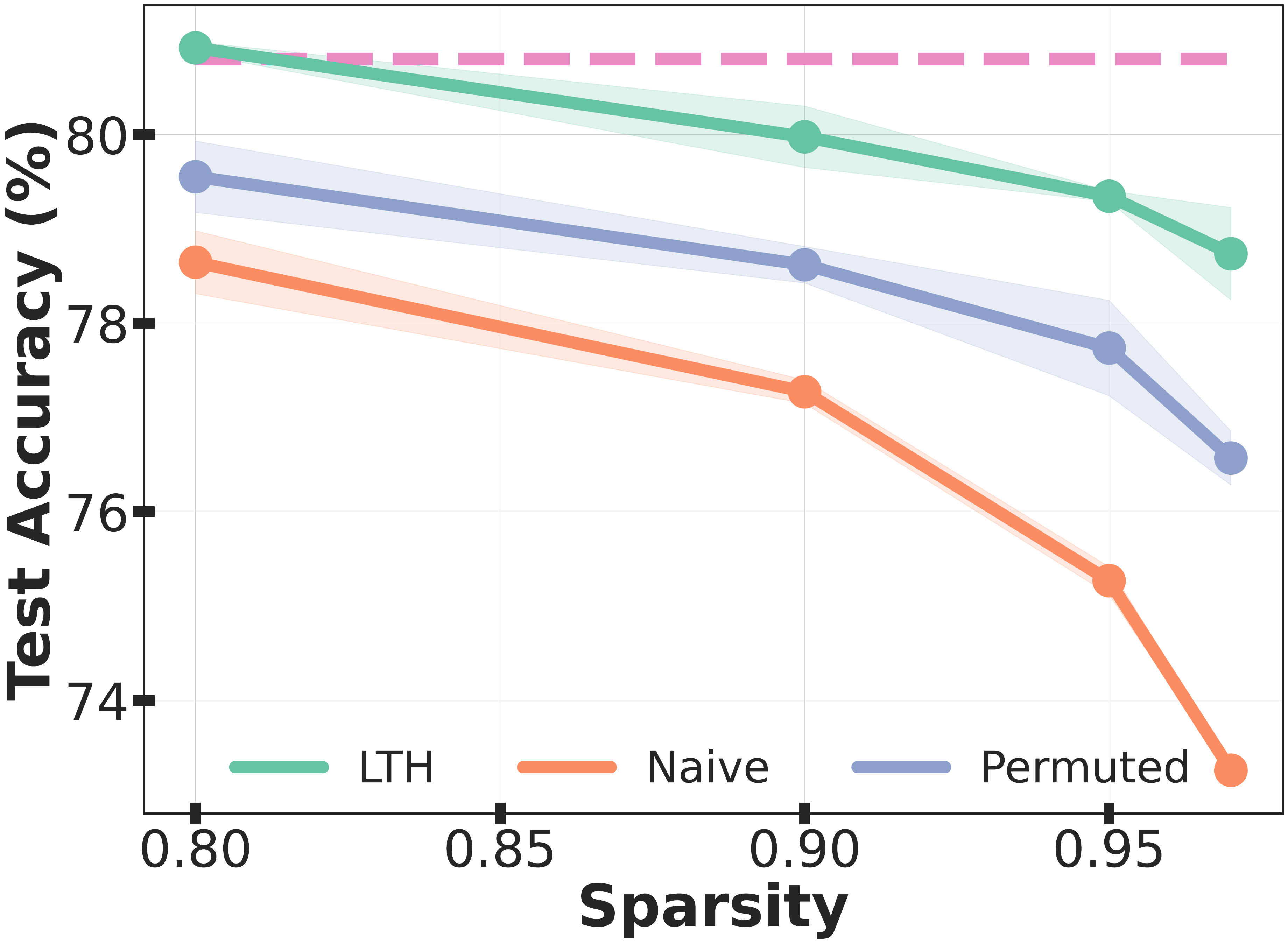}  
        \caption{rewind = 50}
        \label{fig:resnet_rew50_w8:3}
    \end{subfigure}
    \centering
    \begin{subfigure}{0.235\textwidth}
        \centering
        \includegraphics[width=\linewidth]{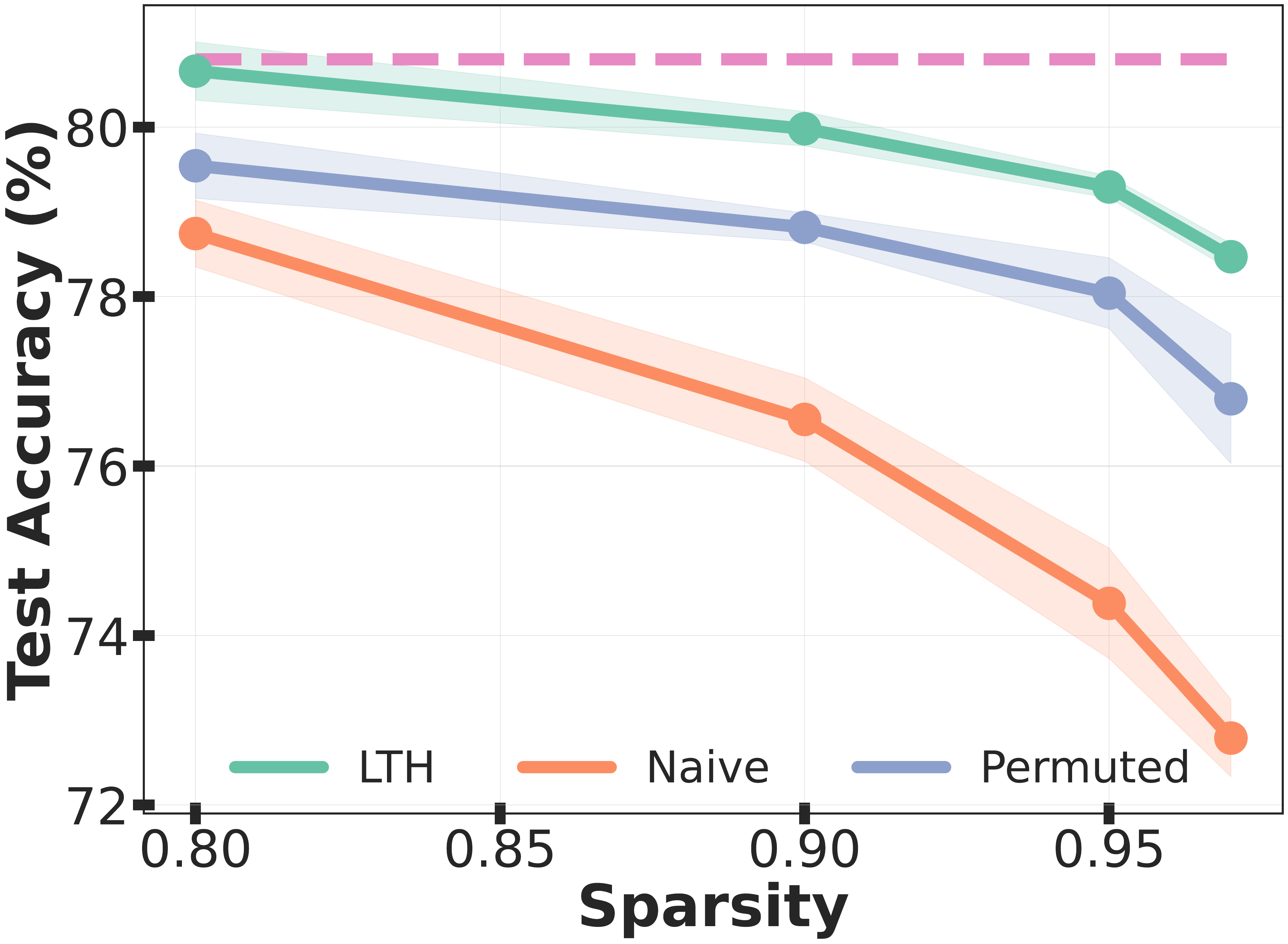} 
        \caption{rewind = 100}
        \label{fig:resnet_rew100_w8:4}
    \end{subfigure}\\
    \centering
    \begin{subfigure}{1.5em}
        \makebox[20pt]{\raisebox{50pt}{\rotatebox[origin=c]{90}{$w=16$}}} % 
    \end{subfigure}
    \begin{subfigure}{0.235\textwidth}
        \centering
        \includegraphics[width=\linewidth]{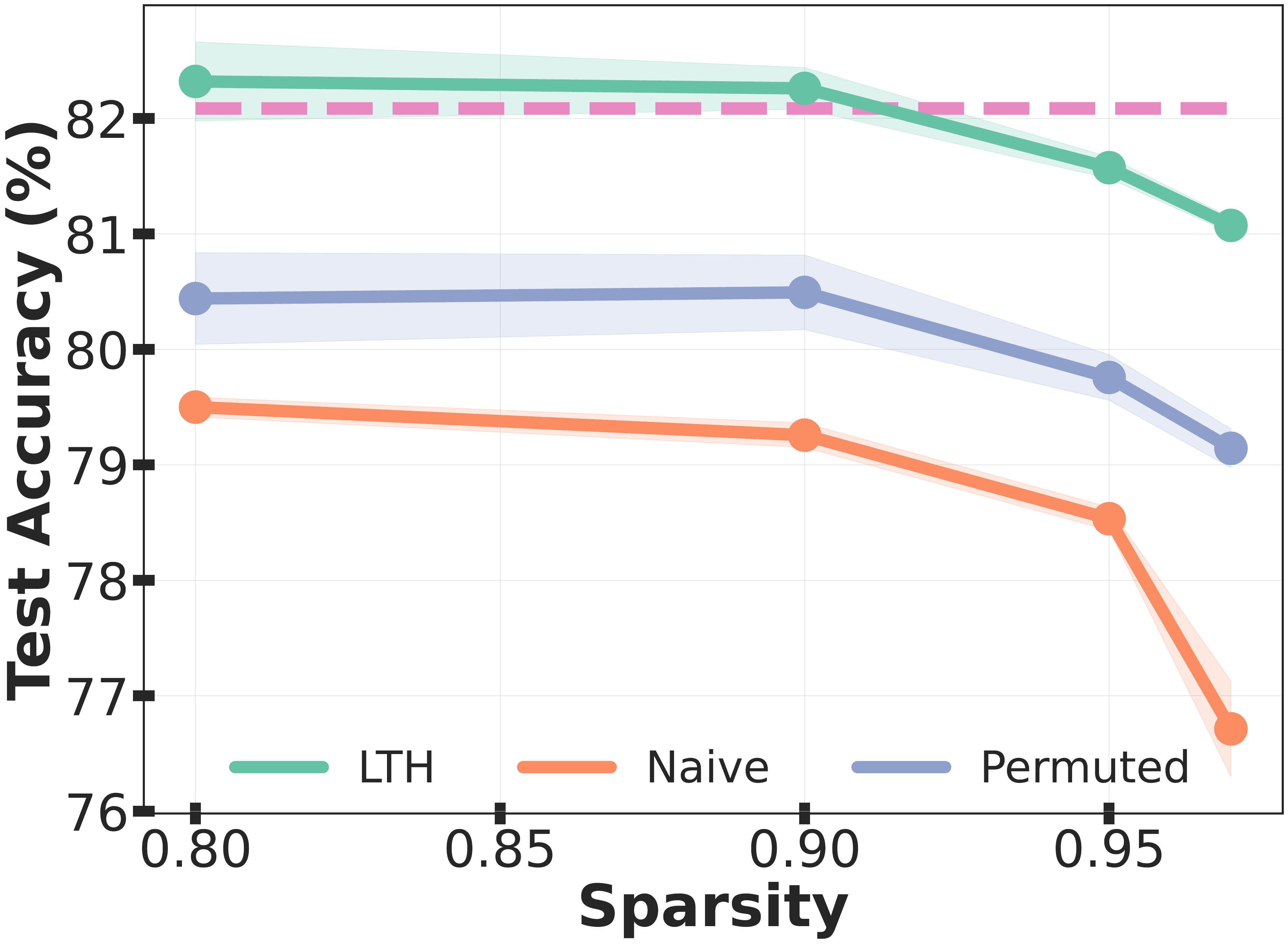}  
        \caption{rewind=10}
        \label{fig:rw_1_10_w16_c100:1}
    \end{subfigure}
    \centering
    \begin{subfigure}{0.235\textwidth}
        \centering
        \includegraphics[width=\linewidth]{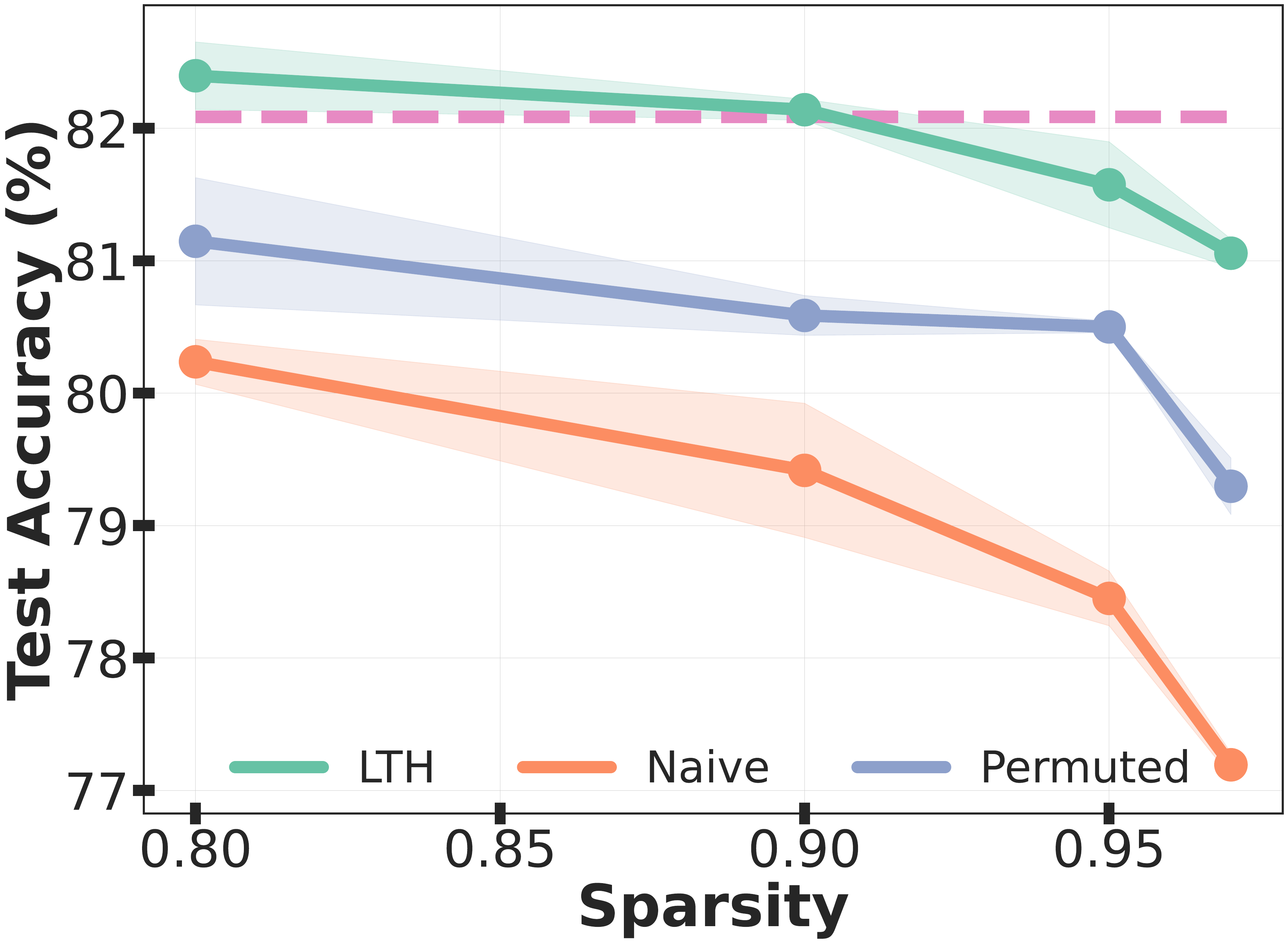}  
        \caption{rewind = 25}
        \label{fig:rw_1_25_w16_c100:2}
    \end{subfigure}
    \centering
    \begin{subfigure}{0.235\textwidth}
        \centering
        \includegraphics[width=\linewidth]{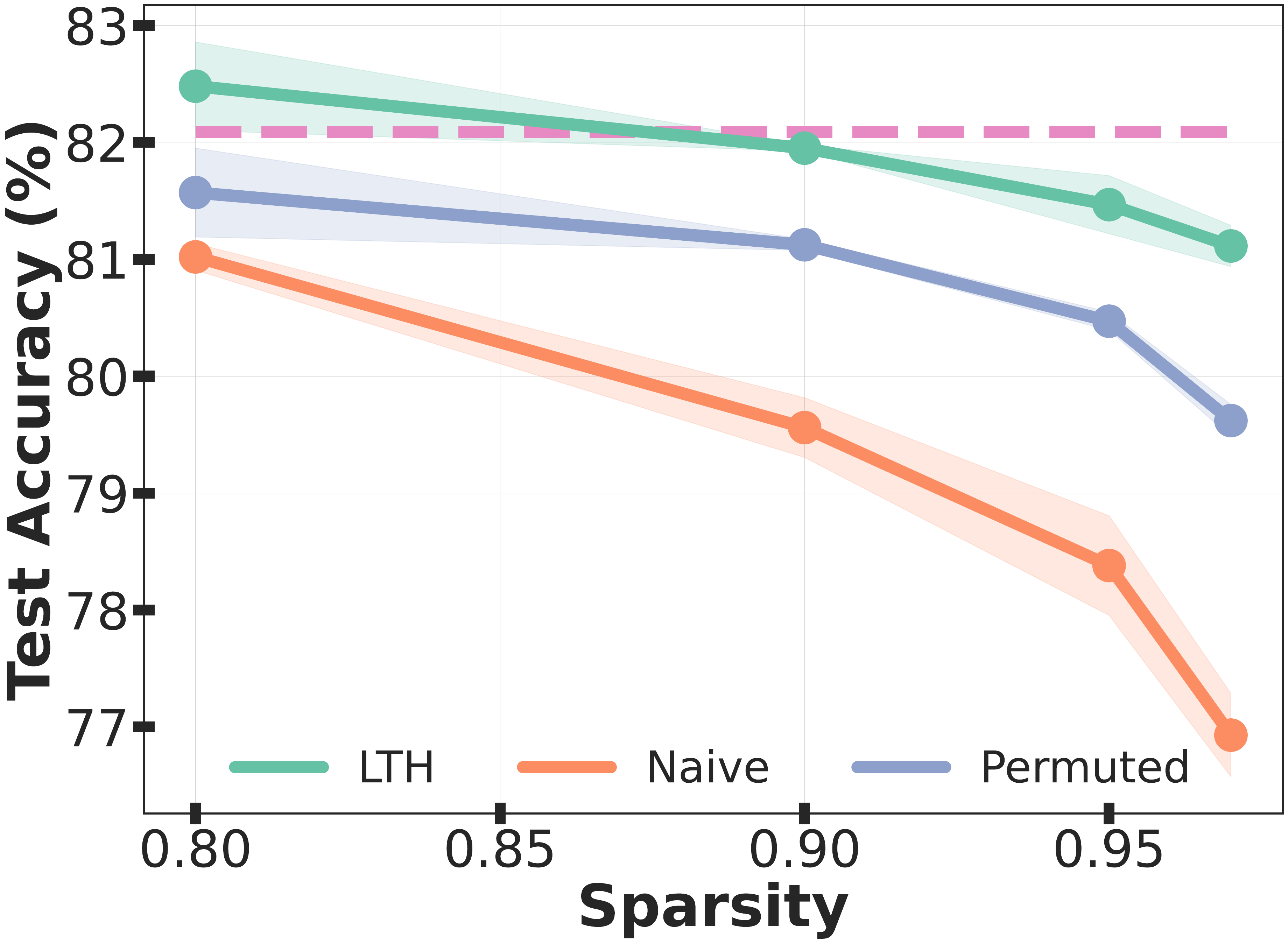}  
        \caption{rewind = 50}
        \label{fig:rw_1_50_w16_c100:3}
    \end{subfigure}
    \centering
    \begin{subfigure}{0.235\textwidth}
        \centering
        \includegraphics[width=\linewidth]{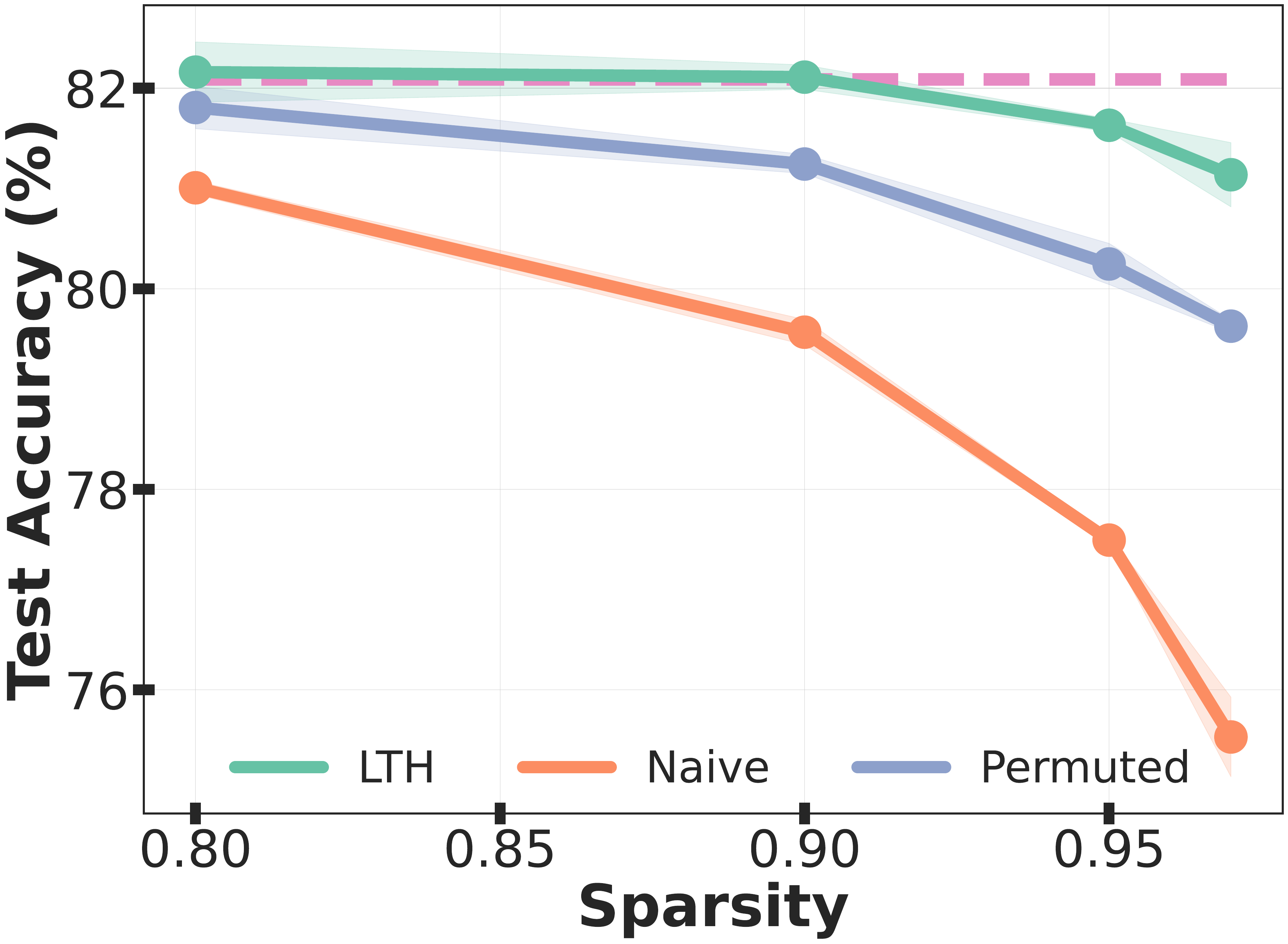} 
        \caption{rewind = 100}
        \label{fig:rw_1_100_w16_c100:4}
    \end{subfigure}
    \caption{\small{\textbf{Accuracy vs sparsity trend for} \textbf{ResNet20$\times\{w\}$/CIFAR-100}. Similar to the phenomenon seen in ~\cref{fig:resnet_w1_sp}, with higher width, the gap between permuted and naive solutions increases. As seen in ~\cref{fig:accresnet_w1_sp_97_fig_rew100:4,fig:resnet_rew100_w4:4,fig:resnet_rew100_w8:4,fig:rw_1_100_w16_c100:4} and the permuted solution approaches the \gls{lth} solution. The dashed ({\textbf{- -}}) line shows the dense model accuracy.}}
    \label{fig:resnet_c100_acc_vs_sparsity}
\end{figure}

\begin{figure}[tbp]
    \centering
    \begin{subfigure}{0.32\textwidth}
        \centering
        \includegraphics[width=\linewidth]{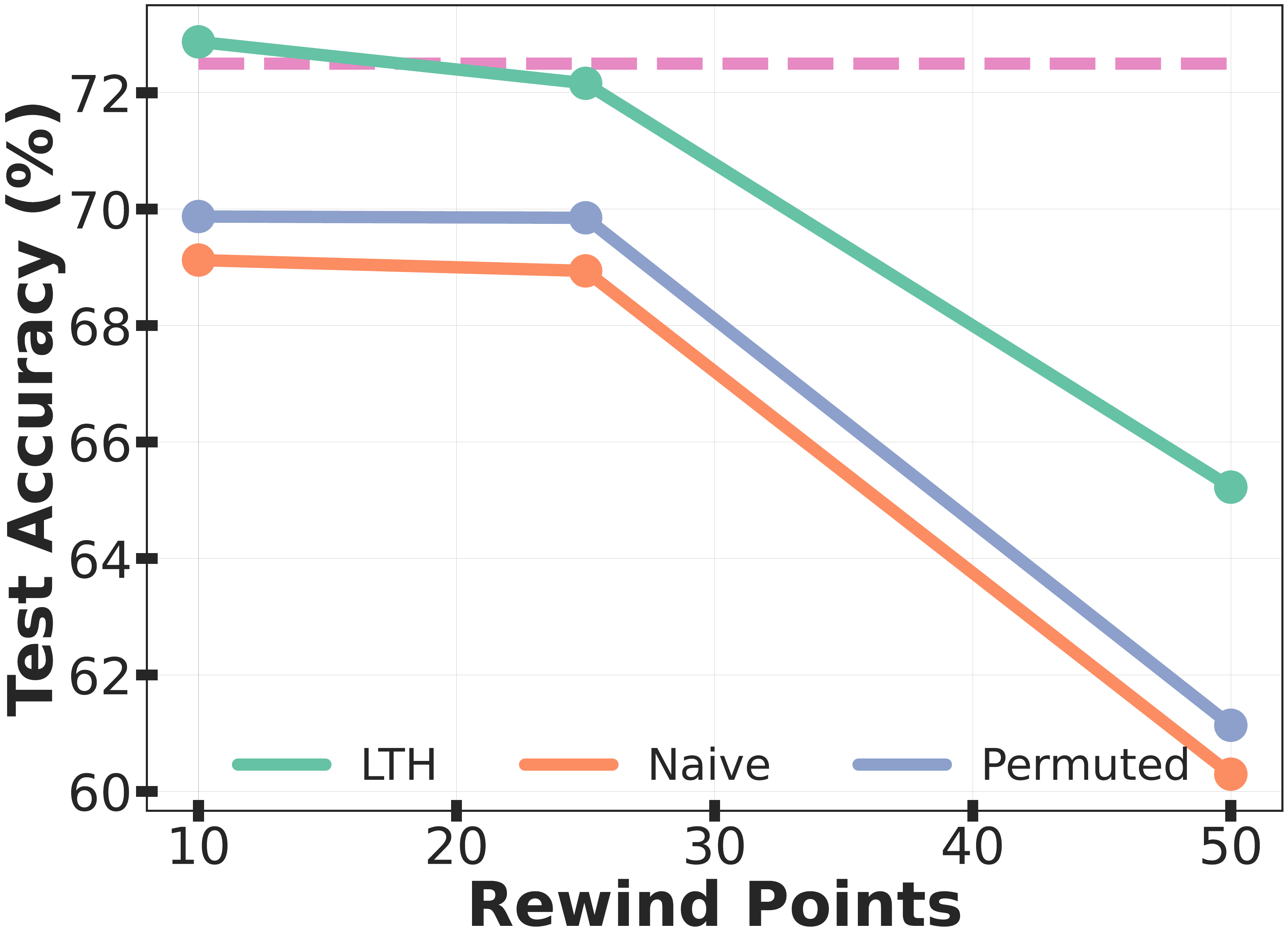}
        \caption{sparsity = 0.80}
        \label{fig:imagenet_80_plot}
    \end{subfigure}
    \begin{subfigure}{0.32\textwidth}
        \centering
        \includegraphics[width=\linewidth]{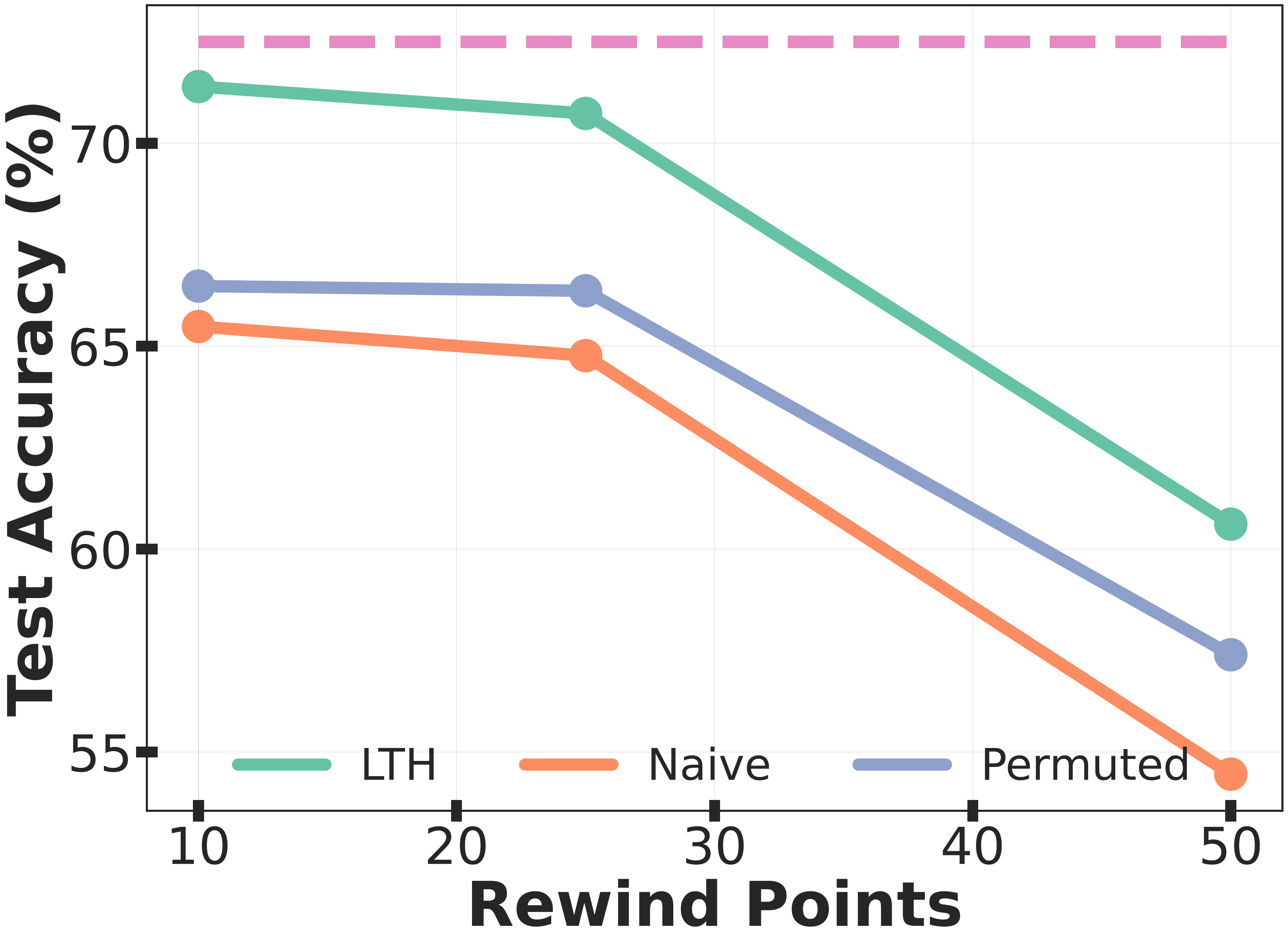}
        \caption{sparsity = 0.90}
        \label{fig:imagenet_90_plot}
    \end{subfigure}
    \begin{subfigure}{0.32\textwidth}
        \centering
        \includegraphics[width=\linewidth]{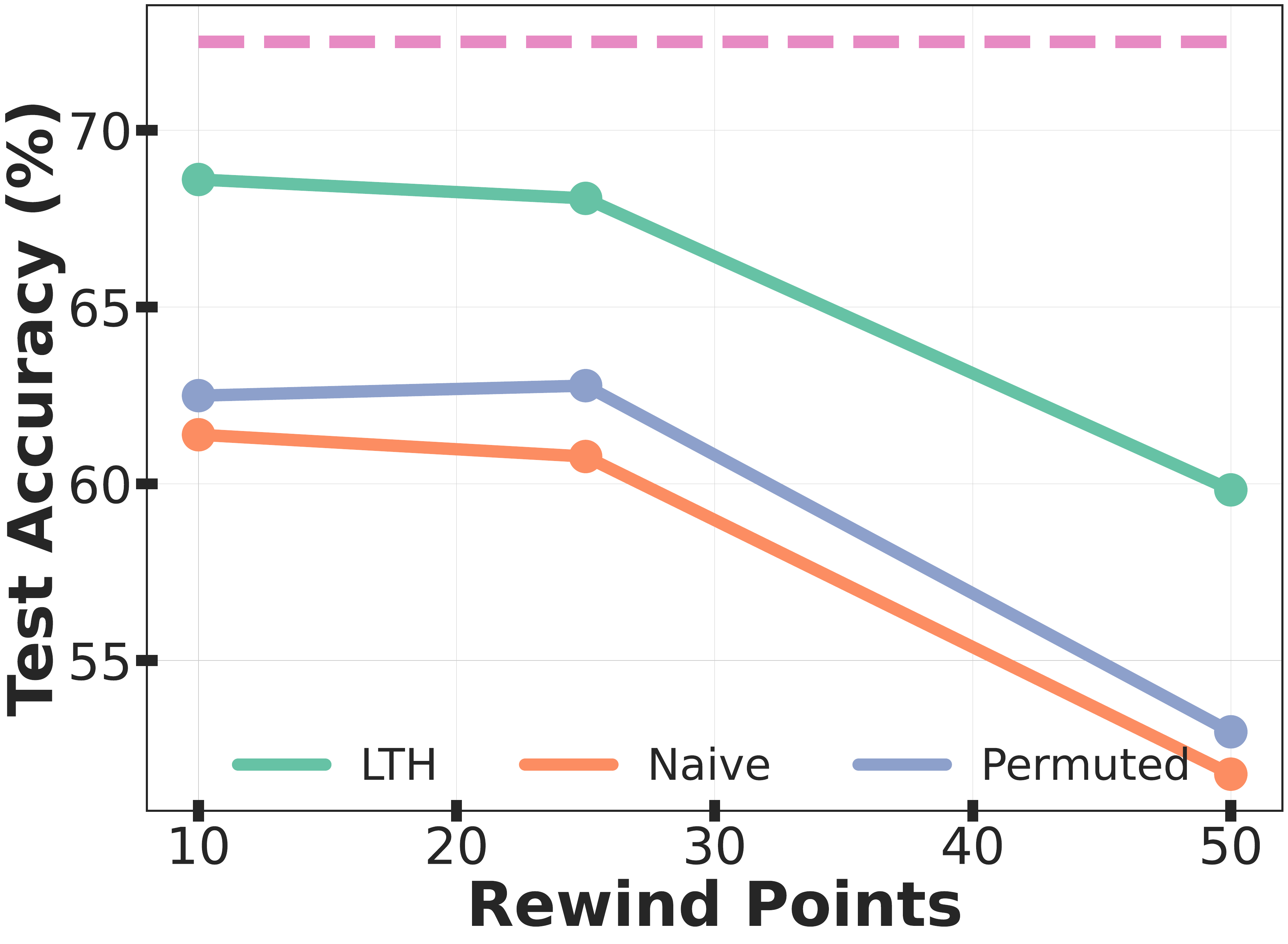}
        \caption{sparsity = 0.95}
        \label{fig:imagenet_95_plot}
    \end{subfigure}
    \caption{\small{{\textbf{ResNet50$\times\{1\}$/ImageNet}. Top-1 test accuracy vs rewinds points of sparse network solutions at various sparsity levels. We observe the permuted solution consistently peroforming better than the naive solution for all sparsities. The dashed ({\textbf{- -}}) line shows the dense model accuracy. }}}
    \label{fig:imagenet_plots_top1}
\end{figure}

Refer to~\cref{fig:resnet_w1_sp,fig:resnet_c100_acc_vs_sparsity} for additional accuracy-vs-sparsity plots for ResNet20 on CIFAR-10 and CIFAR-100. Refer to~\cref{fig:imagenet_plots_top1} for Top-1 accuracy vs.\ rewind points for ResNet50 on ImageNet.
\label{subsec:add_plots}

\section{Early Matching}
% In our work, we train two dense models to convergence in order to find a permutation mapping using activation matching. However, it has been observed that the permutation mapping becomes stable early in the training~\citep{sharma2024simultaneous}.
%% A more detailed version I wrote for my thesis. 
In our current methodology, we train both models $A$ and $B$ to convergence, resulting in the weight configurations $\textbf{w}_A^{t=T}$ and $\textbf{w}_B^{t=T}$, respectively. However, it has been observed in~\citep{sharma2024simultaneous} that it is possible to find permutation mapping earlier in training. In effort to reduce the computational cost associated with our approach, we aim to find a permutation $\pi$ that allows us to suitably align the weights of model $A$ at convergence, $\textbf{w}_A^{t=T}$, with the weights of model $B$ at an earlier training iteration $i\ll T$, $\textbf{w}_B^{t=i}$. In \cref{table:early_matching}, we provide additional results on early matching with the CIFAR-10 dataset, which shows that models can be matched earlier in the training, thus reducing the computational cost of our method.

\section{Additional Details for \cref{ensemble}}
\label{ensemble_details}
In \cref{ensemble}, the IMP solution is trained independently over $5$ different seeds with iterative pruning to obtain $5$ different sparse/pruned solutions with different sparse masks/topologies ($M_1$, $M_2$, $M_3$, $M_4$, $M_5$). The LTH ensemble is trained using the same mask ($M_1$) and initialization ($w_1$) over $5$ different runs (with different data order). Random initialization $w_1$ defines the winning ticket for mask $M_1$. The permuted ensemble is trained using $5$ different permutations ($\pi_1$, $\pi_2$, $\pi_3$, $\pi_4$, $\pi_5$) of the same mask ($M_1$) with five different random weight initializations ($w_1$, $w_2$, $w_3$, $w_4$, $w_5$).

\begin{table}[tbp]
\centering
\caption{\small{\textbf{ResNet20$\times\{1\}$/CIFAR-10}. Results using the ResNet20$\times\{1\}$ trained on CIFAR-10, from a rewind point $k = 20$, using various methods of sparse training with sparsity $S$. The LTH and naive methods remain fixed as they are independent of matching. For the permuted method, the permutation, $\pi$, is obtained by matching a fully trained dense model at $t = T$ ($T = 200$) with another model at an early point in training at $t = i$, where $i \in \{5, 20, 50, 100\}$.}}\label{table:early_matching}
% \resizebox{\textwidth}{!}{
\begin{tabular}{@{}p{1.3em}lccccc@{}}
    
    \toprule
    & & \multicolumn{5}{c}{Early Matching Point $t$} \\
         \cmidrule(l){3-7}
     $S$ & Method & $t=5$ & $20$ & $50$ & $100$ & $200$ \\
     \midrule
    \multirow{3}{2em}{{80\%}} 
    & {LTH}   &\multicolumn{5}{c}{92.25 $\pm$ 0.14} \\
    % \cmidrule(lr){2-7}
    & {naive} & \multicolumn{5}{c}{90.13 $\pm$ 0.11} \\
    \cmidrule(lr){2-7}
    & {perm.} &90.49 $\pm$ 0.37 & 90.34 $\pm$ 0.63 & 90.42 $\pm$ 0.29 & 90.42 $\pm$ 0.25 & 90.68 $\pm$ 0.18\\
    \midrule
    \multirow{3}{2em}{{90\%}} 
    & {LTH}   & \multicolumn{5}{c}{91.18 $\pm$ 0.27} \\
    % \cmidrule(lr){2-7}
    & {naive} & \multicolumn{5}{c}{88.83 $\pm$ 0.27} \\
    \cmidrule(lr){2-7}
    & {perm.} &89.16 $\pm$ 0.51 & 89.23 $\pm$ 0.59 & 89.39 $\pm$ 0.69 & 89.31 $\pm$ 0.60 & 89.50 $\pm$ 0.27 \\
    \midrule
    \multirow{3}{2em}{{95\%}} 
    & {LTH}   & \multicolumn{5}{c}{90.58 $\pm$ 0.26} \\
    % \cmidrule(lr){2-7}
    & {naive} & \multicolumn{5}{c}{87.31 $\pm$ 0.36} \\
    \cmidrule(lr){2-7}
    & {perm.} &87.37 $\pm$ 0.33 & 87.68 $\pm$ 0.77 & 87.43 $\pm$ 1.00 & 87.54 $\pm$ 0.43 &88.29 $\pm$ 0.52 \\
    \midrule
    \multirow{3}{2em}{{97\%}} 
    & {LTH}   & \multicolumn{5}{c}{89.21 $\pm$ 0.23}\\
    % \cmidrule(lr){2-7}
    & {naive} & \multicolumn{5}{c}{85.36 $\pm$ 0.14} \\
    \cmidrule(lr){2-7}
    & {perm.} &85.77 $\pm$ 0.44 & 85.93 $\pm$ 0.94 & 86.09 $\pm$ 0.51 & 85.88 $\pm$ 0.47 &86.12 $\pm$ 0.27 \\
    \bottomrule
\end{tabular}
% }
\end{table}

\section{Computational Overhead of the Permuted Solution}
The primary difference in computational complexity between the \gls{lth}, naive, and permuted solutions lies in the process of neuronal alignment, where weight/activation matching is used to locate permutations in order to bring the hidden units of two networks into alignment. To obtain the permuted solution, two distinct models must be trained independently to convergence, after which their weights or activations are aligned through a permutation-matching process. This alignment, though relatively efficient, adds a small computational overhead compared to LTH and naive solutions, which do not involve matching steps. However, it’s important to note that the primary goal of this study is not to improve training efficiency but rather to investigate why the LTH framework fails when applied to sparse training from new random initializations (not associated with the winning ticket's mask).

\section{Full Symmetry Figure including Lottery Ticket Hypothesis}
\label{full_lth_figure}
% Figure Source: https://docs.google.com/presentation/d/1sS2zjXqrXhsBVisSRm-sMA9ae9HM6NfxB29H7gzwSvU/edit?usp=sharing
\begin{figure}[!htbp]
    \centering
    \begin{subfigure}{0.49\textwidth}
      \centering
      \includegraphics[page=1, width=0.95\linewidth]{figures/sparse-rebasin-symmetry-illustration}
      \caption{Dense training and pruning model $A$.}%
      \label{fig:firstsymmetry-full}%
    \end{subfigure}
    \begin{subfigure}{0.49\textwidth}
      \centering
      \includegraphics[page=2, width=0.95\linewidth]{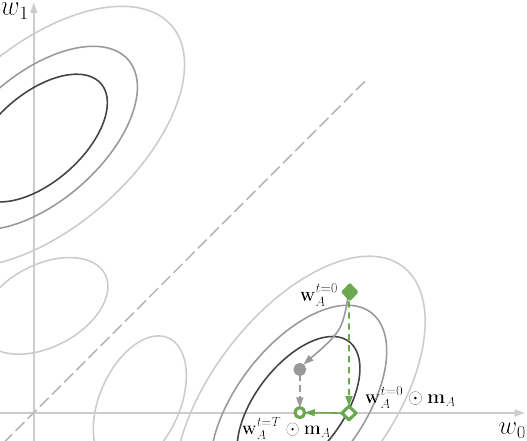}
      \caption{\gls{lth} training model $A$ with pruned mask.}%
      \label{fig:firstsymmetrylth}%
    \end{subfigure}
    % \qquad % space out the images a bit
    \begin{subfigure}{0.6\textwidth}
      \centering
      \includegraphics[page=3, width=0.95\linewidth]{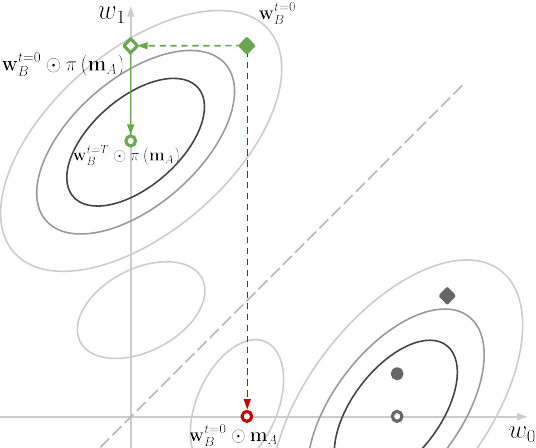}
      \caption{Sparse training model $B$ with $A$ mask.}
      \label{fig:secondsymmetry-full}%
    \end{subfigure}
  \caption{\small{\textbf{Weight Symmetry and the Sparse Training Problem (Full Figure)}. A model with a single layer and only two parameters, $\mathbf{w}=({w_0, w_1})$, operating on a single input $x_0$ has the weight symmetry in the 2D loss landscape as illustrated above. In (\subref{fig:firstsymmetry-full}) the original dense model, $\mathbf{w}_A$, is trained from a random dense initialization, $\mathbf{w}_A^{t=0}$ to a dense solution, $\mathbf{w}_A^{t=T}$, which is then pruned using weight magnitude resulting in the mask $\mathbf{m}_A = \left(1, 0\right)$. 
  In (\subref{fig:firstsymmetrylth}) we re-use the init.\ $\mathbf{w}_A^{t=0}$, to train model $A$ with the pruned mask from (\subref{fig:firstsymmetry-full}), $\mathbf{m}_A$, as in the \gls{lth}.
  In (\subref{fig:secondsymmetry-full}), naively using the same mask to train a model, B, from a different random initialization will likely result in the initialization being far from a good solution. Permuting the mask to match the (symmetric) basin in which the new initialization is in will enable sparse training.}}% caption for whole figure
  \label{fig:symmetry-illustration-full}% label for whole figure
\end{figure}
In \cref{fig:symmetry-illustration-full} we include the full version of \cref{fig:symmetry-illustration}, including an illustration of the \gls{lth} in \cref{fig:firstsymmetrylth}.

\end{document}